%% file: main.tex
\documentclass[11pt]{article}

\usepackage[utf8]{inputenc}
\usepackage[T1]{fontenc}
\usepackage{amsmath,amssymb,amsfonts}
\usepackage{mathtools}
\usepackage{bm}
\usepackage{graphicx}
\graphicspath{{figures_paper1/}}
\usepackage{subcaption}
\usepackage{booktabs}
\usepackage{multirow}
\PassOptionsToPackage{hyphens}{url}
\usepackage{hyperref}
\usepackage{xurl}%
\usepackage{cleveref}
\usepackage{xcolor}
\usepackage[most]{tcolorbox}%
\usepackage{algorithm}
\usepackage{algorithmic}
\usepackage[margin=1in]{geometry}
\usepackage{natbib}
\usepackage{enumitem}
\usepackage{wrapfig}
\usepackage[section]{placeins}%

\usepackage{amsthm}%

\theoremstyle{remark}

\newif\ifrevcolor \revcolorfalse
\definecolor{revred}{rgb}{0.72,0.05,0.05}
\ifrevcolor
  
  \newcommand{\revon}{\color{revred}}
\else
  
  \newcommand{\revon}{}
\fi

\title{A foundation model for energy and radiation systems built on
heterogeneous scientific interfaces}
\author{
  Samrendra Roy$^{1}$ \qquad
  Tapas Tripura$^{2,3}$ \\[4pt]
  Yoon Pyo Lee$^{1}$ \qquad
  Souvik Chakraborty$^{1,2,3}$ \\[4pt]
  Syed Bahauddin Alam$^{1,4}$\thanks{Corresponding author: \texttt{alams@illinois.edu}}
  \\[8pt]
  {\normalsize $^{1}$Department of Nuclear, Plasma, and Radiological Engineering,} \\
  {\normalsize University of Illinois Urbana-Champaign, Urbana, IL 61801, USA} \\[3pt]
  {\normalsize $^{2}$Department of Applied Mechanics,} \\
  {\normalsize Indian Institute of Technology Delhi, Hauz Khas, New Delhi 110016, India} \\[3pt]
  {\normalsize $^{3}$Yardi School of Artificial Intelligence (ScAI),} \\
  {\normalsize Indian Institute of Technology Delhi, Hauz Khas, New Delhi 110016, India} \\[3pt]
  {\normalsize $^{4}$National Center for Supercomputing Applications (NCSA),} \\
  {\normalsize University of Illinois Urbana-Champaign, Urbana, IL 61801, USA}
}

\date{}

\newcommand{\SecRef}[2]{\Cref{#1}}%
\begin{document}
\maketitle

\input{sections/abstract}

\begin{tcolorbox}[enhanced,colback=blue!4!white,colframe=blue!45!black,colbacktitle=blue!45!black,
  coltitle=white,fonttitle=\bfseries,title={Significance},
  boxrule=0.9pt,arc=2.5pt,left=6pt,right=6pt,top=4pt,bottom=4pt,before skip=8pt,after skip=10pt]

Foundation models entered science by analogy with language and vision, where a homogeneous substrate of tokens or patches already existed and breadth could be purchased with corpus scale. Energy and radiation engineering has no such substrate, so every scientific foundation model to date has manufactured one before training, placing the conversion step outside the evaluated system, and we show that this conversion is consequential rather than neutral: a coordinate convention inside one baseline's interface moves its reported error by more than an order of magnitude with the model unchanged. This paper therefore defines a foundation model for the domain operationally, by reuse and extension rather than corpus scale, and instantiates one that serves fluid, radiation and structural problems on Cartesian, spherical and unstructured domains within a single pretrained core, acquires a previously unseen reactor-relevant component by training 2.1\% of its parameters while every deployed prediction remains bit-identical, and, through a norm-matched randomized-library audit, converts the customary assumption that cross-physics pretraining helps into a measurement of when it does and when it does not.
\end{tcolorbox}

\input{sections/introduction2}
\input{sections/results2}
\input{sections/discussion}
\input{sections/Methods}

\section*{Acknowledgements}
This work used the Delta and DeltaAI systems at the National Center for Supercomputing Applications [awards OAC 2005572 and OAC 2320345] through allocation CIS240093 from the Advanced Cyberinfrastructure Coordination Ecosystem: Services \& Support (ACCESS) program, which is supported by National Science Foundation grants \#2138259, \#2138286, \#2138307, \#2137603, and \#2138296.

\section*{Data availability}
The lid-driven cavity dataset is from the conformalized neural-operator study of Kobayashi et al.~\citep{kobayashi2025conformalized}, the cosmic-ray dose dataset from the TRON study~\citep{kobayashi2025tron} and the elastoplastic deformation dataset from the S-DeepONet benchmark of He et al.~\citep{he2024sdeeponet}; the heat-exchanger dataset was generated by our group for a companion preprint~\citep{roy2026adversarial} and the subchannel dataset is from the virtual-sensing study of Hossain et al.~\citep{hossain2025virtualsensing}.%
All processed arrays, splits and normalization statistics needed to reproduce the reported numbers are available from the corresponding author on reasonable request.

\section*{Code availability}
The GEODE implementation, training and evaluation scripts, baseline adapters and pretrained checkpoints will be made available upon publication.

\section*{Use of large language models}
Large language models were used to assist with editing the text of the manuscript and with debugging analysis code. All experimental designs, results, interpretations and conclusions are the authors' own, and all text was reviewed and revised by the authors. No large language model is an author of this work.

\section*{Author contributions}
S.R., T.T., S.C. and S.B.A. contributed to the conception, execution and writing of this work.

\section*{Competing interests}
The authors declare no competing interests.

\bibliographystyle{unsrtnat}
\bibliography{Neutron_references}

\clearpage
\appendix
\setcounter{figure}{0}
\setcounter{table}{0}
\renewcommand{\thefigure}{S\arabic{figure}}
\renewcommand{\thetable}{S\arabic{table}}
\renewcommand{\theHfigure}{supp.\arabic{figure}}
\renewcommand{\theHtable}{supp.\arabic{table}}
\input{sections/supp_results}
\input{sections/supp_extra}

\end{document}

%% file: sections/abstract.tex
\begin{abstract} Scientific foundation models are commonly evaluated after heterogeneous physical problems have already been translated into a compatible gridded, tokenized or symbolic representation. This leaves the scientific interface outside both the pretrained model and the audit of what is actually reused. We study the complementary setting in which boundary histories, sparse monitor records and loading histories retain their native inference classes and their outputs remain on Cartesian, latitude--longitude and unstructured domains. GEODE couples task-specific scientific interfaces to a shared routed library of wavelet operators. A single jointly pretrained model represents cavity flow, radiation dose and elastoplastic stress, then acquires a heat exchanger and a reactor subchannel by training a private interface containing 2.1\% of its parameters. Earlier predictions remain unchanged by parameter isolation, whereas unrestricted fine-tuning degrades them by factors of 14--29. Crucially, preservation alone does not establish reuse: norm-matched randomized-library controls show that the contribution of pretrained computation is conditional on the task and data regime. A separate decomposition shows that full-field relative $L^2$ error can substantially understate error relative to spatial variation when field level dominates the norm. Task-specific operators remain more accurate on three of the five problems. These results distinguish multi-task coverage, preservation and pretrained reuse as separate properties that must be tested independently when scientific foundation models span heterogeneous interfaces. \end{abstract}

\noindent\textbf{Keywords:} scientific foundation models $\cdot$ neural operators $\cdot$ energy systems $\cdot$
radiation $\cdot$ geometry-adaptive decoding $\cdot$
cross-task transfer

%% file: sections/introduction2.tex
\section{Introduction}
\label{sec:intro}

Scientific foundation models face a representation problem before they face a scaling problem. Language and vision inherit broadly shared substrates of tokens or image patches. Scientific models often create an analogous substrate before pretraining by rasterizing fields, prescribing a common tensor layout, or encoding the governing equations symbolically. This strategy is powerful when the conversion is faithful, but the conversion itself then lies outside the model being evaluated. Consequently, a high downstream score does not by itself reveal how much performance comes from the pretrained computation and how much comes from the representation used to make the downstream problem compatible with it. This distinction becomes unavoidable when the scientific problems do not share an input language. In energy and radiation modeling, a prescribed boundary history, a sparse monitor record, a mechanical loading history and an equipment operating condition have different physical meanings and define different inference classes. Their outputs likewise occupy Cartesian grids, latitude--longitude coordinates and irregular component meshes. Reuse therefore cannot be identified with placing more datasets into one training run. A model can preserve earlier tasks simply by isolating their parameters, can fit several tasks without reusing learned computation, and can obtain a small full-field error while reproducing field level more accurately than spatial structure. These possibilities are usually conflated by aggregate downstream accuracy. We therefore ask a narrower and experimentally falsifiable question: when scientific tasks retain incompatible native interfaces, what evidence is required to establish that a pretrained computational core is actually reused? We study this question in energy and radiation systems because they provide a controlled stress test spanning parameter-to-field, sparse-observation-to-field and history-to-field mappings. We operationalize reuse through interventions that separately test multi-task coverage, preservation of earlier functions and contribution of pretrained computation to a new task, while reporting where specialists remain preferable. GEODE is the model used to perform this audit, rather than the premise from which the claim is inferred.

We use energy and radiation systems as the scope of the domain, and the inclusion of
radiation is deliberate rather than opportunistic. Radiation transport and radiation
protection are constitutive of nuclear energy rather than adjacent to it, and the
inference problem they present is structurally distinct from the others considered
here: a network of point detectors reports counts at scattered geographic locations,
and a distributed dose field must be recovered on a spherical coordinate chart with an
order of magnitude more output nodes than the other tasks. That combination, a
sparse-observation conditioning modality with a non-Cartesian native output domain,
is precisely what a common gridded substrate would have to eliminate, and it is the
reason the task is retained. The five tasks evaluated here therefore span fluid
transport, radiation-field reconstruction, path-dependent structural response,
compact heat-exchange and rod-bundle coolant transport. They do not constitute the
complete reactor modeling stack, and we make no claim that they do; they are chosen
to separate the requirements stated below rather than to survey the domain.

High-fidelity solvers remain indispensable within this ecosystem, but their repeated evaluation across operating conditions, inverse problems, uncertainty analyses and design loops is prohibitively expensive. Neural operators address this computational burden by learning mappings between function spaces and thereby amortizing simulation cost across families of inputs rather than approximating the solution of only one problem instance~\citep{lu2021learning,li2020fourier,li2021fourier,kovachki2023neural,tripura2023wavelet}. Although early neural operators were developed primarily for fields on regular grids, geometry-informed formulations, including Geo-FNO, the geometry-adaptive Waveformer, $\pi$G-Sp$^2$GNO, GINO, GNOT and Transolver, have extended operator learning to irregular meshes, point clouds and complex physical domains~\citep{li2022fourier_geometry,navaneeth2025geometry_waveformer,li2023geometry,sarkar2026pigsp2gno,hao2023gnot,wu2024transolver}. These developments make neural operators increasingly relevant to component-scale engineering models, but they generally retain a task-specific learning paradigm: one model is trained for one governing system, one prescribed input--output relationship, and one class of geometry. A surrogate for boundary-driven flow cannot ordinarily be reused to reconstruct a radiation field from monitors, predict stress from a loading history or model coolant transport in a fuel assembly. Consequently, the computational models needed across an energy or radiation workflow remain a collection of separately trained specialists, even when their repeated construction, deployment and maintenance impose the same practical burden.

Scientific foundation models have recently been proposed to replace this one-model-per-task paradigm with a shared representation pretrained across multiple partial differential equation (PDE) datasets and subsequently reused through fine-tuning, few-shot adaptation or in-context conditioning~\citep{yang2023context,subramanian2024towards}. Multiple Physics Pretraining (MPP) introduced autoregressive, task-agnostic pretraining of a transformer across heterogeneous spatio-temporal systems, using shared embedding and normalization layers to place their physical fields in a common latent space; it demonstrated simultaneous prediction on the pretraining systems and improved transfer to unseen systems, but its benchmark remained predominantly fluid mechanical and represented each problem as a regularly sampled field history~\citep{mccabe2023multiple}. DPOT enlarged this multi-equation paradigm through an autoregressive denoising objective, a Fourier-attention operator transformer and large-scale PDE pretraining, establishing favorable scaling and downstream transfer while continuing to formulate the supported problems as structured field-tensor evolution~\citep{hao2024dpot}. Poseidon combined a multiscale operator transformer with lead-time-conditioned normalization and a semigroup-based training strategy, achieving strong accuracy and sample efficiency across fifteen downstream tasks after pretraining on fluid dynamics; its demonstrated interface is nevertheless a field-to-field map on structured discretizations, with time entering as a conditioning variable, and acquiring a downstream task relies on fine-tuning rather than isolating the new task from those learned earlier~\citep{herde2024poseidon}. The Neural Combinatorial Wavelet Neural Operator (NCWNO) pursued a complementary objective by routing computations through gated local wavelet experts and using memory-based ensembling to learn several parametric PDE operators and acquire additional ones with resistance to catastrophic forgetting; this is the closest precedent for the continual-learning component of GEODE, but its demonstrations use compatible gridded function representations and do not combine distinct observational mappings or native output geometries within one model~\citep{chakraborty2024ncwno}. Walrus subsequently scaled cross-domain pretraining to a 1.3-billion-parameter transformer trained on nineteen two- and three-dimensional continuum-dynamics scenarios spanning astrophysics, geoscience, rheology, plasma physics, acoustics and classical fluids, with compute-adaptive tokenization and stabilization designed for heterogeneous resolutions and long-horizon forecasting~\citep{mccabe2025walrus}. Walrus provides the broadest physical and dimensional coverage in this trajectory and demonstrates powerful transfer, but it remains primarily a forecaster of fluid-like continuum fields from preceding field states. Taken together, these models establish that shared pretraining can transfer across governing systems, but they principally address collections of simulation trajectories that have already been expressed through a compatible field-based interface.

A second line of development broadens the language through which physical problems can be presented to a foundation model. PDEformer-2 represents the governing equation, coefficients, initial and boundary conditions and domain through a computational graph, and combines this representation with an implicit neural representation queried at arbitrary space--time coordinates; its 40-TB pretraining corpus spans multiple symbolic PDE forms, domain shapes, boundary conditions and numbers of variables, enabling zero-shot prediction within the pretrained distribution and few-shot adaptation beyond it~\citep{ye2025pdeformer2}. This equation-centered formulation offers substantially greater symbolic and geometric versatility than a field-history model, but presupposes that the task can be expressed through its governing PDE and associated conditions, which does not provide a native interface for observational reconstruction when the governing model is unavailable or incomplete. MORPH instead introduces a unified physics tensor and an attention-based backbone that operate across one-, two- and three-dimensional spatio-temporal datasets with different resolutions and mixtures of scalar and vector fields, followed by full-model or parameter-efficient adaptation~\citep{rautela2025morph}. Its notion of shape agnosticism encompasses dimensionality, resolution and field composition, but its unified physics-tensor format, convolutional preprocessing and patching presuppose regularly sampled fields, and its supported inputs remain histories of physical fields rather than different classes of scientific observation. PDE-FM similarly pretrains a shared spatial--spectral state-space backbone on twelve two- and three-dimensional datasets spanning hydrodynamic, radiative, elastic and astrophysical systems, with per-dataset adapters and physics-aware conditioning~\citep{soares2025pdefm}. Although it accommodates multiple coordinate systems and variable sets, the datasets are standardized as gridded trajectories and the reported study does not establish sequential acquisition with retention of earlier tasks. These models considerably expand the equation, dimensional, and physical scope of scientific pretraining, but they obtain this breadth by selecting a common representational substrate, either an explicit PDE specification or a spatio-temporal field tensor. That strategy is difficult to apply directly to energy and radiation systems, where a boundary actuation, a sparse radiation-monitor record, a mechanical loading history, and an equipment operating condition are not interchangeable descriptions of the same kind of field evolution. The outstanding challenge is therefore to share computation across an application domain without first eliminating the heterogeneity that characterizes its native models and observations.

The heterogeneity of energy and radiation modeling has three distinct components. First, the conditioning information differs not only in dimension but in scientific meaning and inference class. A flow solver may be conditioned on an imposed boundary actuation, a radiation-monitoring model on sparse sensor time series~\citep{kobayashi2024mimonet}, a structural model on a material loading history, and an equipment model on operating parameters or distributed heat input. These are respectively parameter-to-field, sparse-observation-to-field and history-to-field mappings, and their inputs cannot be aligned through coordinate-wise correspondence. Second, their outputs occupy different physical representations, including Cartesian flow domains with several coupled variables, global spherical coordinates carrying a scalar radiation field, irregular finite-element meshes resolving localized stress concentrations, and component-specific meshes containing internal solid boundaries. Third, energy systems evolve as new components, operating regimes, and sensing capabilities are introduced, so a useful shared model must acquire previously unseen tasks without corrupting those already deployed. Expressing every problem through one symbolic PDE language is not always possible for observational tasks, while mapping them onto a common field tensor can obscure native coordinates, introduce interpolation and masking artifacts, and conflate absent observations with physically meaningful zero values. Geometry generality alone is therefore insufficient, as is multi-equation pretraining that assumes a common inference interface. A foundation model for energy and radiation systems must combine cross-physics learning, heterogeneous conditioning modalities, native geometric decoding, and continual acquisition without catastrophic interference. The central question is whether a reusable computational core can be built across this application-level heterogeneity rather than only across different equations presented in a standardized numerical form.

\begin{tcolorbox}[colback=blue!4!white, colframe=blue!45!black, title={\textbf{Three quantities that standard foundation-model evaluation can conflate}}, fonttitle=\bfseries] \small \textbf{Interface performance is not backbone performance.} A downstream scientific problem must first be expressed in the representation accepted by the pretrained model. That conversion can materially alter the reported result. In our audit, correcting only a coordinate-normalization convention in the interface to one geometry-general baseline changes its LDC error from $59.21\%$ to $2.195\%$ and its PlasticDeform error from $19.27\%$ to $1.083\%$, with the operator weights unchanged (\Cref{tab:supp_gino_diagnostic}). A downstream score is therefore a property of the model together with the interface through which the scientific problem reaches it. \textbf{Preservation is not reuse.} Earlier predictions can be preserved exactly by parameter isolation, and independent specialists preserve them trivially by sharing nothing. Neither observation shows that computation learned before the new task contributes to learning it. We therefore test reuse against norm-matched randomized cores of identical architecture (\Cref{sec:conditional_transfer}), rather than inferring it from parameter-efficient adaptation alone. \textbf{Full-field error is not necessarily structural error.} Relative $L^2$ error is normalized by the norm of the complete target field. When field level dominates spatial variation, this normalization can make a model appear substantially closer to the target even when the zero-mean spatial profiles are much more similar. Across the tasks studied here the corresponding scale factor ranges from approximately $2$ to $121$, and in one Subchannel comparison a $37$-fold difference in full-field error corresponds to differences of at most $2.1$ in the zero-mean spatial profile (\Cref{sec:conditional_transfer}). \end{tcolorbox}

\begin{tcolorbox}[colback=orange!5!white, colframe=orange!55!black,
  title={\textbf{Two axes of scientific foundation modeling}},
  fonttitle=\bfseries]
\small
Existing scientific foundation models have primarily increased \emph{corpus breadth}: the number of equations, trajectories, dimensions and parameter regimes represented within a compatible input--output language. The present study examines a complementary quantity, which we call \emph{interface breadth}: the range of scientifically distinct conditioning modalities and output representations that can use a common pretrained computation without first being converted into one another. The distinction is descriptive rather than hierarchical. Models such as DPOT, Poseidon, PDEformer-2, MORPH, PDE-FM and Walrus provide substantially broader pretraining corpora and downstream suites than GEODE, whereas the experiments here ask what can be established when interface compatibility itself is part of the problem. We do not test zero-shot solution of unseen equations or scaling with large PDE corpora.
\end{tcolorbox}

\label{sec:definition}The term \emph{foundation model} describes a model pretrained
on broad data and subsequently reused across downstream tasks rather than reconstructed for each
application~\citep{bommasani2021opportunities}. In language and vision that
description is nearly self-executing, because the substrate is homogeneous and reuse
follows from it. In an engineering domain it is not, because the same phrase is
satisfied in appearance by any model trained on several datasets at once. What must
be added is a statement of what reuse means when the datasets are not mutually
convertible, and that statement has to be testable, since otherwise the term
certifies nothing. To make the foundation-model claim falsifiable in this heterogeneous-interface setting, we evaluate it using six operational requirements. These requirements are not proposed as a universal definition of scientific foundation models; they specify the evidence needed for the narrower claim examined here, namely that one pretrained computational core is shared, extended and measurably reused across incompatible scientific interfaces. Each requirement is paired with an experiment or control that could have falsified it. Under this test GEODE satisfies five requirements and partially satisfies one, because task-specific operators remain more accurate on three of the five problems (\Cref{tab:fm_criteria}). We report that outcome rather than adjusting the
requirement, and we note that a reader who scores the sixth requirement differently
changes one row of \Cref{tab:claims_evidence} and none of the six findings recorded
above it.  First, the model must possess a substantial shared computational core that is
pretrained and reused across all supported tasks, while permitting lightweight
task-specific interfaces when native representations are incompatible (F1).
Second, it must learn across structurally distinct physical systems rather than
across parameterizations of one governing equation or closely related family (F2).
Third, it must accommodate heterogeneous scientific interfaces in which the
conditioning information differs in physical meaning and in inference class,
including parameter-to-field, sparse-observation-to-field and history-to-field
mappings (F3). Fourth, it must produce solutions on different geometric domains,
coordinate systems and discretizations without requiring every output to be
rasterized onto a common computational grid (F4). Fifth, introducing a previously
unseen task must require adapting only a small fraction of the model, must leave
the functions associated with earlier tasks unchanged, and must demonstrate that
the pretrained core contributes to the new task relative to a norm-matched
randomized core of identical architecture (F5). Finally, this sharing must retain
practical predictive utility, so that the joint and adapted model is evaluated
against task-specific specialists and appropriate controls and any loss of accuracy
is reported rather than absorbed (F6).

The third clause of F5 is the one that does analytical work, and we state it
explicitly because without it the criterion is vacuous. Preservation of earlier
functions is obtained trivially by maintaining a separate model per task, which
shares nothing and therefore cannot interfere. What distinguishes a shared
foundation from a collection of independent surrogates is that the retained
computation is reused, and reuse is an empirical property that must be measured
against a control in which the same architecture holds no learned content. We
report that control throughout, and it is the reason several of our conclusions are
conditional rather than general. F1 to F4 define the representational scope
required to place the different physical and observational problems of an energy or
radiation workflow inside one model; F5 and F6 determine whether the result is a
reusable foundation or a simultaneous fit to several datasets.
\Cref{tab:fm_criteria} positions GEODE and the closest classes of existing neural
operators and scientific foundation models against these requirements, based strictly
on capabilities demonstrated in the cited studies rather than on extensions their
architectures might support.

\input{sections/fm_criteria_table}

The comparison in \Cref{tab:fm_criteria} shows that no single axis is sufficient to establish the complete claim. Walrus provides much broader physical, dimensional and data-scale coverage than GEODE; PDEformer-2 provides broader equation-level and zero-shot versatility; MORPH provides broader dimensional and spatio-temporal tensor coverage; Poseidon establishes stronger sample-efficient transfer across a larger downstream suite; and NCWNO supplies a direct precedent for continual multi-PDE operator learning. GEODE addresses a complementary application-level conjunction of requirements by separating the task-dependent representation of energy and radiation data from the computation intended to be shared across systems. Each task is assigned an input--output interface that maps its native conditioning information into a common latent representation and returns the resulting features to its native physical variables and query geometry. The central computational core is a task-conditioned mixture of wavelet experts that provides a shared library of multiscale transformations while allowing different systems to emphasize different expert combinations. A geometry-adaptive decoder uses physical query coordinates, positional information and distance-to-boundary features to evaluate this representation on Cartesian, spherical and irregular domains without requiring the physical outputs to be remeshed onto a universal Cartesian grid. This separation places sharing where it is physically plausible: GEODE does not assume that a sparse radiation-monitor record, a prescribed flow boundary condition and a mechanical loading history possess coordinate-wise semantic correspondence, but allows the latent computations extracted from them to draw on the same pretrained expert library. When a previously unseen component or physical task is introduced, the expert library and all earlier interfaces are frozen, while a new input--output interface and routing function are learned for the incoming system. GEODE therefore does not posit a universal representation into which every energy problem can be inserted without adaptation; it treats lightweight task-specific interfaces as the mechanism by which a common computational foundation can expand across heterogeneous systems.

We evaluate GEODE on five tasks drawn from energy and radiation engineering:
lid-driven cavity flow, global cosmic-ray dose reconstruction, elastoplastic
deformation, compact heat-exchanger flow and pressurized-water-reactor subchannel
transport. They span three inference classes, three classes of native output domain,
output meshes of $1{,}733$ to $65{,}341$ nodes and conditioning vectors of $84$ to
$102$ dimensions with no coordinate-wise correspondence. The resulting evaluation deliberately combines scientific interfaces that are not coordinate-wise interchangeable, and \Cref{tab:fm_criteria} records the comparison against cited alternatives only for capabilities demonstrated in those studies. Four findings follow.

First, one jointly pretrained model represents all three pretraining tasks in their
native representations, with median relative $L^2$ errors of $0.755\%$, $0.031\%$
and $0.399\%$ and skill scores above $0.99$ against per-node mean-field controls.
This establishes coverage rather than superiority: a task-specific operator remains
more accurate on the structured cavity problem, and a small shared DeepONet without
an expert library attains comparable joint accuracy. Joint prediction across fixed
geometries is therefore not, by itself, an argument for this architecture, and we do
not present it as one.

Second, the operative unit of extension is the scientific interface, not the router.
Adapting the encoder, output projection, skip paths, geometry decoder and gate, a
private interface of $3.9$ million parameters or $2.1\%$ of the model, reaches
$0.719\%$ on a previously unseen heat-exchanger component while every earlier
prediction remains unchanged, because no parameter on which it depends is
modified. The two alternatives fail in different ways: increasing router capacity
over several orders of magnitude leaves the new task near $10\%$ error while
degrading the pretrained tasks through the shared gate, and unrestricted fine-tuning
degrades them by factors of $14$ to $29$. This is a negative result about routing,
and it runs against the expectation, natural in the mixture-of-experts setting, that
recombining an existing library suffices to absorb a new problem.

Third, preservation and reuse are separate properties and only the first is
structural. Against norm-matched randomized libraries of identical architecture, the
pretrained core helps a new component most when its data are scarce, is
indistinguishable from random content once sufficient data are available, and in one
case determines whether the interface learns the absolute field level at all. This
is the measurement that distinguishes a shared foundation from a collection of
independent surrogates, since independent surrogates preserve earlier functions
perfectly and reuse nothing.

Fourth, the standard accuracy metric in this literature can obscure which of those
properties was obtained. Decomposing the subchannel case shows that a $37$-fold
difference in full-field relative error between two models corresponds to
differences of at most a factor of $2.1$ in the zero-mean spatial profile, the
remainder residing in absolute field level. Because relative $L^2$ error divides by
the norm of the complete field, it equals the error relative to spatial variation
divided by a scale factor $\kappa$ set by the ratio of field mean to field variation
(\Cref{eq:scale_factor}), and $\kappa$ ranges from approximately $2$ to $121$ across
these tasks.
We therefore report errors relative to both the complete field and its variation
throughout, and we regard this as a measurement standard for the field rather than a
property of this architecture.

Under the operational test used here, GEODE provides a shared pretrained core across heterogeneous scientific interfaces and can be extended through a private interface containing $2.1\%$ of its parameters while leaving earlier functions unchanged by construction. The randomized-library controls further determine when the frozen core contributes learned content rather than merely providing fixed computation. This evidence does not establish zero-shot solution of unseen governing equations, generalization to unseen geometries, or superiority to specialized models. The claim tested in this work is therefore one of auditable reuse across heterogeneous interfaces, not universal scientific prediction.

\begin{tcolorbox}[enhanced,colback=blue!4!white,colframe=blue!45!black,colbacktitle=blue!45!black,
  coltitle=white,fonttitle=\bfseries,title={What this paper establishes, and what it does not},
  boxrule=0.9pt,arc=2.5pt,left=6pt,right=6pt,top=4pt,bottom=4pt,before skip=8pt,after skip=10pt]
The Results therefore test three propositions separately: whether one model can represent the pretrained tasks without converting their native interfaces into one another, whether new tasks can be added without changing earlier functions, and whether the frozen computation learned before acquisition measurably improves the new task relative to an otherwise identical randomized core. The last proposition is essential, because preservation by itself can be obtained without sharing any learned computation.
\end{tcolorbox}

%% file: sections/fm_criteria_table.tex
\providecommand{\cmark}{\ensuremath{\checkmark}}
\providecommand{\xmark}{\ensuremath{\times}}
\providecommand{\pmark}{\ensuremath{\bigcirc}}

\begin{table}[htbp]
\centering
\caption{Operational requirements for a foundation model of energy and radiation
systems. These requirements concern \emph{interface breadth}, the number of mutually
inconvertible inference classes, conditioning modalities and native output geometries
a single pretrained core can serve. They are deliberately silent on \emph{corpus
breadth}, the number of trajectories, equations and dimensions observed during
pretraining, along which several of the systems listed here exceed the present work by
a wide margin. Neither axis subsumes the other and an engineering domain requires
both. Entries are assessed against what each class of system was demonstrated to do in
the cited work.  \cmark{} demonstrated; \pmark{} partial, restricted or
problem-specific; \xmark{} outside the reported setting. The table is not a claim
that these architectures could not be extended along the axes they leave open;
several plausibly could, and such extensions would constitute separate studies.
F1 shared pretrained core; F2 structurally distinct physics; F3 heterogeneous
conditioning modality and inference class; F4 native geometric and discretization
diversity; F5 small-fraction extension with unchanged earlier functions and
demonstrated reuse of the pretrained core against a randomized control; F6
evaluated against task-specific specialists with any accuracy cost reported.}
\label{tab:fm_criteria}
\small
\renewcommand{\arraystretch}{1.2}
\setlength{\tabcolsep}{4pt}
\begin{tabular}{@{}p{0.60\textwidth}cccccc@{}}
\toprule
\textbf{Class or system} & \textbf{F1} & \textbf{F2} & \textbf{F3} & \textbf{F4}
& \textbf{F5} & \textbf{F6} \\
\midrule
Independent task-specific operators\textsuperscript{a}
  & \xmark & \xmark & \xmark & \pmark & \pmark & \cmark \\
Geometry-general operators~\citep{li2022fourier_geometry,li2023geometry,hao2023gnot,wu2024transolver,navaneeth2025geometry_waveformer,sarkar2026pigsp2gno}
  & \xmark & \xmark & \pmark & \cmark & \xmark & \cmark \\
In-context operator learning~\citep{yang2023context}
  & \cmark & \pmark & \cmark & \xmark & \pmark & \pmark \\
Multiple Physics Pretraining~\citep{mccabe2023multiple}
  & \cmark & \cmark & \pmark & \xmark & \pmark & \cmark \\
DPOT, Poseidon~\citep{hao2024dpot,herde2024poseidon}
  & \cmark & \cmark & \pmark & \xmark & \pmark & \cmark \\
PDEformer-2~\citep{ye2025pdeformer2}
  & \cmark & \cmark & \pmark & \cmark & \pmark & \cmark \\
MORPH, PDE-FM~\citep{rautela2025morph,soares2025pdefm}
  & \cmark & \cmark & \pmark & \pmark & \pmark & \cmark \\
Walrus~\citep{mccabe2025walrus}
  & \cmark & \cmark & \pmark & \pmark & \pmark & \cmark \\
NCWNO~\citep{chakraborty2024ncwno}
  & \cmark & \cmark & \pmark & \xmark & \cmark & \pmark \\
\textbf{GEODE (this work)}
  & \cmark & \cmark & \cmark & \cmark & \cmark & \pmark \\
\bottomrule
\end{tabular}
\par\vspace{3pt}
{\footnotesize
\begin{flushleft}
\textsuperscript{a}\,A collection of separately trained surrogates, one per task.
It receives \pmark{} on F5 because earlier functions are preserved by construction
while no pretrained computation is reused, which is the distinction the third
clause of F5 is designed to expose. GEODE receives \pmark{} on F6 because
task-specific operators remain more accurate on three of the five evaluated
problems (\Cref{sec:joint_accuracy,sec:continual_acquisition}); the reported
advantage is broader native coverage and controlled extension, not uniform
accuracy.
\end{flushleft}}
\end{table}

%% file: sections/results2.tex
\section{Results}
\label{sec:results}

\paragraph{Evaluation protocol and provenance.}
All GEODE accuracies reported here come from a single clean checkpoint: the
$190.6$-million-parameter configuration with the signed-distance feature active,
retrained with the PlasticDeform validation split carved from its training pool,
selected at epoch $118$ on validation data only, and evaluated once on the untouched
test sets. Every downstream analysis is recomputed from that checkpoint using the
recipes recorded in its saved arguments. Private-interface acquisition results, the
data-fraction sweep, the randomized-library, dense-backbone and path-ablation arms
all follow the historical acquisition protocol, in which the private interface starts
from a fresh initialization. Baseline checkpoints are likewise selected on validation
data disjoint from the test set: the DeepONet and GINO runs reserve $1{,}490$ of the
$14{,}900$ PlasticDeform training examples for validation, fit on the remaining
$13{,}410$ and evaluate on the separate $100$-example test set, while LDC and
CosmicDose use the supplied validation and test arrays.

One correction made during this work is reported rather than set aside, because it
bears on how every cross-model comparison in this literature should be read. Our
initial GINO runs normalized the query coordinates inconsistently with the model's
forward-pass inputs. Repairing that convention, with the operator itself unchanged,
moved the reported LDC error from $59.21\%$ to $2.195\%$ and the PlasticDeform error
from $19.27\%$ to $1.083\%$. The superseded values are retained as
\Cref{tab:supp_gino_diagnostic} and are not used in any comparison. We record them
because they quantify something that cross-architecture comparisons of scientific
foundation models do not usually expose: when a model is evaluated on a task whose
native representation differs from its expected input, the reported score is a
property of the model and of the interface jointly, and an order of magnitude of that
score can reside in the interface. This observation motivates our treatment of the
interface as a trained component of the system rather than as preprocessing, and it
is the reason we identify every adapted evaluation explicitly and never assign a
proxy score to a model we could not run natively.

The experiments are organized as tests of the six requirements of
\Cref{sec:definition} rather than as a survey of tasks. Requirements F1 to F4
concern representational scope and are examined in
\Cref{sec:heterogeneous_representation,sec:joint_accuracy,sec:geometry_efficiency};
F5 concerns extension and reuse and is examined in
\Cref{sec:continual_acquisition,sec:conditional_transfer}, which are the sections
that carry the foundation-model claim; F6 concerns the accuracy cost of sharing and
is examined against task-specific operators, mean-field and reduced-order controls,
and matched architectural ablations throughout.

Complete task specifications, dataset partitions, training protocols and metric
definitions are provided in Methods and the Supplementary Information. In the first stage, one model was trained jointly on three systems representing complementary elements of this application domain: lid-driven cavity flow (LDC), global cosmic-ray dose reconstruction (CosmicDose) and elastoplastic stress prediction (PlasticDeform). LDC provides a canonical boundary-driven recirculation problem that captures fluid-mechanical structures relevant to reactor plena, coolant cavities and other confined-flow components, without being presented as a reactor simulation. CosmicDose represents radiation monitoring by reconstructing a global dose field from sparse neutron-monitor measurements, the reconstruction that underlies radiation protection for aviation crews, high-altitude and space operations and ground-level monitoring networks, while PlasticDeform represents the path-dependent structural response that enters fatigue and integrity assessment of load-bearing components subjected to a sign-reversing displacement history. These tasks differ in their governing physics, input modality, predicted variables, coordinate systems and numerical discretizations. Specifically, LDC maps a prescribed lid-velocity history to velocity and pressure on a Cartesian grid, CosmicDose maps geographically distributed sensor measurements to a scalar radiation field on the sphere, and PlasticDeform maps a sign-reversing displacement history to a stress field on an irregular finite-element mesh. In the second stage, compact heat-exchanger flow and pressurized-water-reactor subchannel transport were withheld from pretraining and introduced sequentially, providing direct energy-system tests of whether the pretrained model could acquire new component-level physics, retain its earlier capabilities and reuse its frozen expert library.
Across the five tasks, the input vectors range from 84 to 102 dimensions but have no shared coordinate-wise physical interpretation. The outputs contain between one and three physical variables and are evaluated on domains containing between $1{,}733$ and $65{,}341$ nodes. This experimental design addresses four questions central to an expandable foundation model for energy and radiation systems: whether one shared model can accurately represent multiple physical and observational mappings; whether its routing mechanism organizes the expert library into meaningful shared and task-dependent computation; whether previously unseen energy-system components can be acquired without altering capabilities that may already be deployed; and whether the pretrained expert content provides useful transfer rather than functioning only as a frozen regularizer. Mean-field and reduced-order controls, task-specific neural operators, matched architectural ablations, unrestricted fine-tuning, rehearsal and norm-matched randomization provide the corresponding reference points. Complete task specifications, dataset partitions, training protocols and metric definitions are provided in Methods and the Supplementary Information.
The scope of this evaluation is deliberately bounded. Although the selected tasks connect fluid transport, heat transfer, structural response and radiation-field reconstruction within a common application domain, they do not represent the complete energy or reactor modeling stack. All output domains are two-dimensional, the geometry is fixed within each pretrained task, and the reported predictions correspond to individual states rather than temporal rollouts. Moreover, training does not impose governing-equation residuals, conservation laws or constitutive constraints. The experiments therefore test whether computation can be shared and reused across heterogeneous representations arising in energy and radiation engineering; they do not establish zero-shot solution of arbitrary governing equations, unrestricted generalization to unseen geometries or universal superiority over specialized models.

\begin{table}[htbp]
\centering
\caption{Claims made in this work, the evidence establishing each, the control
that could have falsified it, and its scope. The final column records whether the
claim depends on accepting the foundation-model assessment of
\Cref{sec:definition}. Six of the seven do not: they are results about interfaces,
extension mechanisms and measurement that stand independently of how the term is
adjudicated.}
\label{tab:claims_evidence}
\small
\renewcommand{\arraystretch}{1.25}
\setlength{\tabcolsep}{4pt}
\resizebox{\textwidth}{!}{%
\begin{tabular}{@{}p{4.3cm}p{2.4cm}p{4.2cm}p{4.0cm}c@{}}
\toprule
\textbf{Claim} & \textbf{Established in} & \textbf{Falsifying control applied} &
\textbf{Scope} & \textbf{Independent} \\
\midrule
A reported cross-model score is a property of the model and its interface jointly,
and a large fraction can reside in the interface
& \Cref{sec:results} protocol; \Cref{tab:supp_gino_diagnostic}
& Coordinate-convention repair in the task interface with the operator unchanged
& One baseline, two tasks, single seeds; demonstrates magnitude, not a general bound
& \cmark \\
\addlinespace
One pretrained core serves three inference classes and three native output domains
without rasterizing any onto a common grid
& \Cref{sec:heterogeneous_representation,sec:joint_accuracy,sec:geometry_efficiency}
& Task specialists, shared network without expert library, per-node mean field,
reduced-order regression
& Two-dimensional outputs, single states, geometry fixed within each task
& \cmark \\
\addlinespace
The task interface, not the router, is the operative unit of extension
& \Cref{sec:continual_acquisition}
& Router capacity varied over several orders of magnitude; unrestricted
fine-tuning; rehearsal
& One new component at full data; not a proof of necessity for other
architectures
& \cmark \\
\addlinespace
Deployed predictions are unchanged under extension, by construction
& \Cref{sec:continual_acquisition,sec:methods_acquisition}
& Re-evaluation of every earlier task after each acquisition step
& Structural consequence of parameter disjointness, not a learned property
& \cmark \\
\addlinespace
Reuse of the pretrained core is conditional and must be measured, not assumed
& \Cref{sec:conditional_transfer}
& Norm-matched randomized library; dense backbone without experts; three seeds; the
same frozen-core control applied to adapted MORPH
& One component swept across data fractions, one audited at full data
& \cmark \\
\addlinespace
Full-field relative $L^2$ error can be dominated by field level rather than
spatial structure
& \Cref{sec:conditional_transfer}
& Spatial-mean decomposition; profile-aware loss audit on a matched specialist
& Scale ratio of $2$ to $121$ across the tasks evaluated here
& \cmark \\
\midrule
GEODE satisfies F1 to F5 and partially satisfies F6, and is therefore a foundation
model on the interface axis
& \Cref{sec:definition}; \Cref{tab:fm_criteria}
& The six requirements themselves, each applied as an experiment
& Interface axis only; no corpus-axis capability is claimed or tested
& \xmark \\
\bottomrule
\end{tabular}}
\end{table}

\subsection{One model spans the heterogeneous representations of energy and radiation problems}
\label{sec:heterogeneous_representation}

\begin{figure}[htbp]
  \centering
  \begin{minipage}{\textwidth}\raggedright\textbf{a}\\[-1pt]
  \centering\includegraphics[width=0.88\textwidth]{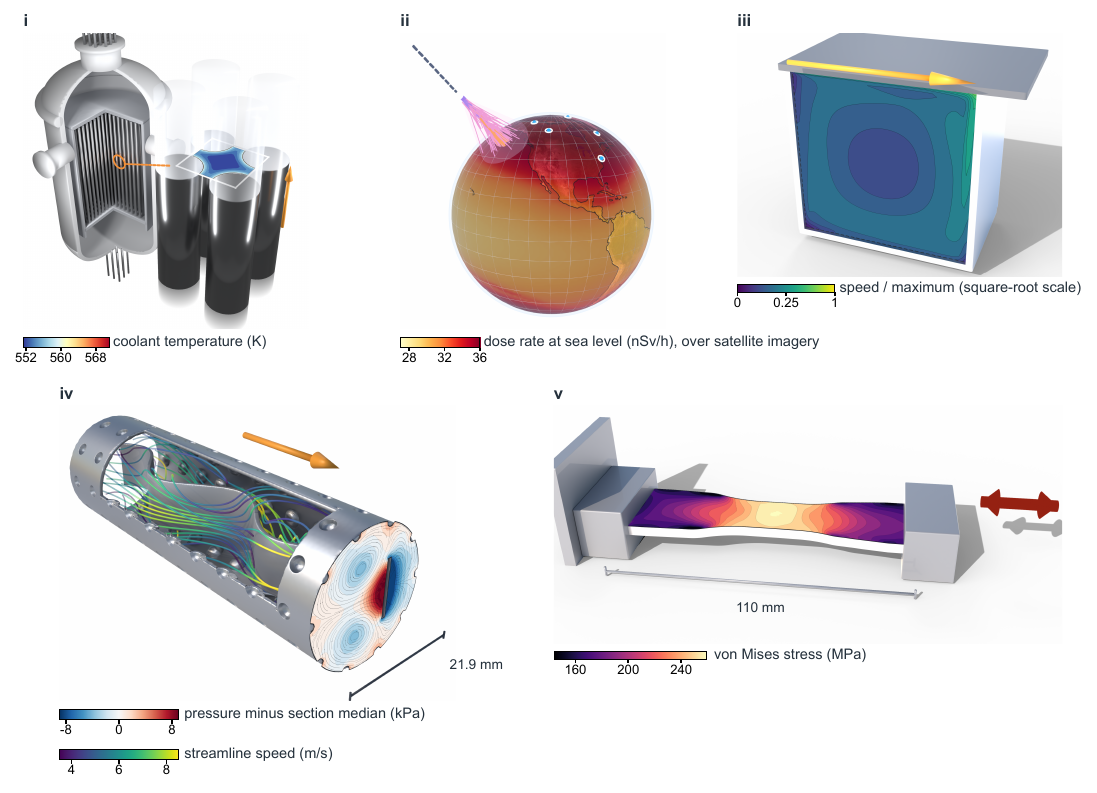}\end{minipage}\\[3pt]
  \begin{minipage}{\textwidth}\raggedright\textbf{b}\\[-1pt]
  \centering\includegraphics[width=0.95\textwidth,clip,viewport=0 325 961 709]{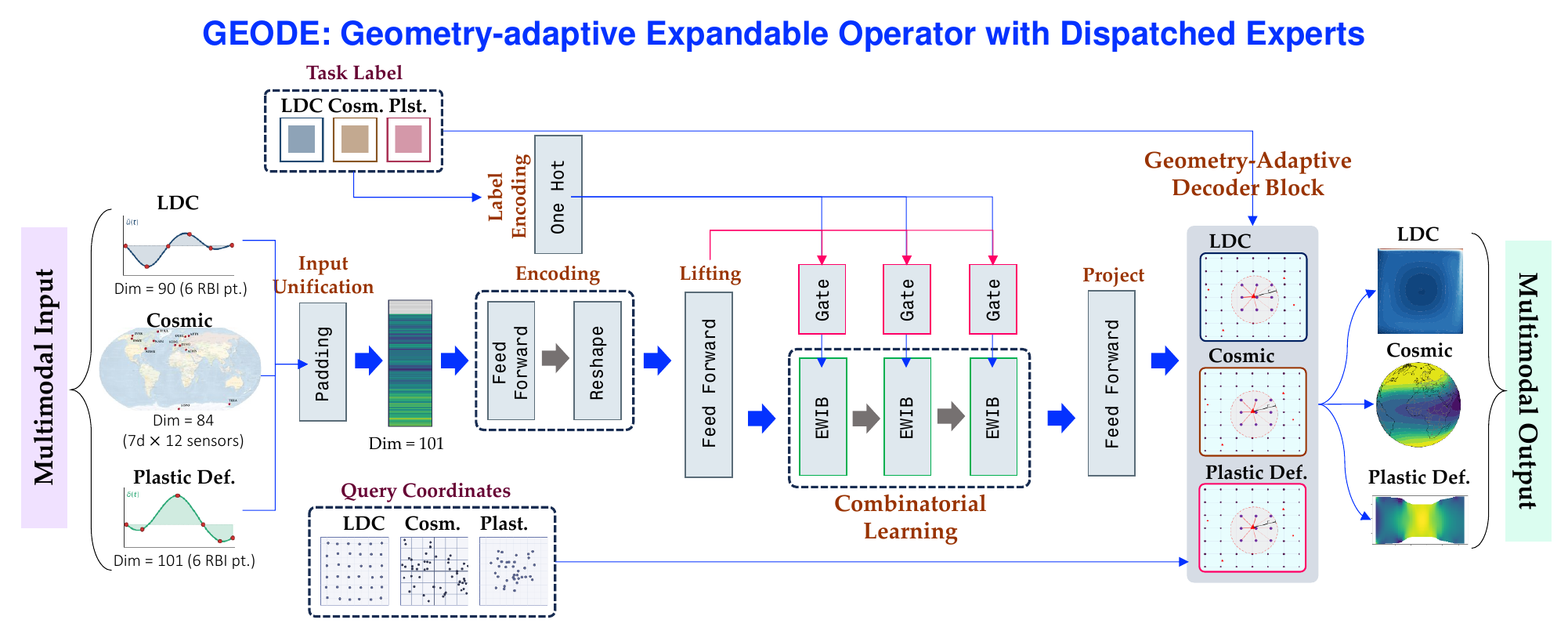}\end{minipage}\vspace{-4pt}
  \caption{\textbf{One operator library across five heterogeneous systems.} \textbf{a}, The five systems as the components their data describe, with the solver's ground-truth field on the plane the data occupy: (i) reactor subchannel between four fuel rods, coolant temperature on the transverse plane, with a schematic vessel cutaway marking where the subchannel sits; (ii) sea-level cosmic-ray dose rate, with the input's twelve neutron-monitor stations marked (those on the far side hidden) and one schematic air shower; (iii) lid-driven cavity, speed; (iv) heat-exchanger tube, pressure on the section and solver streamlines through the cutaway; (v) dog-bone specimen, von Mises stress. Details are given in the text. \textbf{b}, Model structure: task interfaces project heterogeneous inputs onto one $48\times48$ latent grid, three routed wavelet-expert layers compute on it, and the common geometry-adaptive decoder formulation writes each field onto its native mesh; a new system trains only a private interface (input encoder, output projection, skip path, gate and geometry decoder, $3.9$ million parameters, $2.1\%$ of the pretrained model) while the $190.6$ million-parameter library and earlier interfaces stay frozen.}
  \label{fig:results_overview}
\end{figure}

GEODE separates task-specific data representation from computation in a shared latent space (\Cref{fig:results_overview}). Inputs are normalized within each task, padded to a common width and projected onto an abstract $48\times48$ latent grid that is not the physical discretization of any dataset. Three shared wavelet-integral layers, each containing eight Daubechies experts, operate on this latent field. A task-conditioned gate assigns expert weights independently at every layer and feature channel, providing $8^3=512$ possible dominant paths. This number describes routing capacity and is not a claim about the number of tasks supported. Because routing is dense, all experts are evaluated and their weighted responses are combined with a local skip transformation.

The latent representation is returned to physical space through a geometry-adaptive kernel decoder using source and query coordinates and harmonic positional features; the configuration with a signed distance function (SDF) feature additionally includes the bounding-rectangle distance described in Methods. This common decoding formulation supports Cartesian, spherical and irregular output domains without rasterizing them onto one physical grid. When a new system is introduced, the $190.6$ million-parameter pretrained model and all earlier interfaces remain fixed, while a new input encoder, output projection, skip path, gate and geometry decoder are trained. This private interface contains $3.9$ million parameters, or $2.1\%$ of the pretrained model. GEODE therefore shares its principal computational library while retaining the transformations required by different scientific modalities.

\subsection{Joint learning remains accurate across flow, radiation and structural problems}
\label{sec:joint_accuracy}
\label{sec:specialists}

\begin{figure}[htbp]
  \centering
  \includegraphics[width=\textwidth]{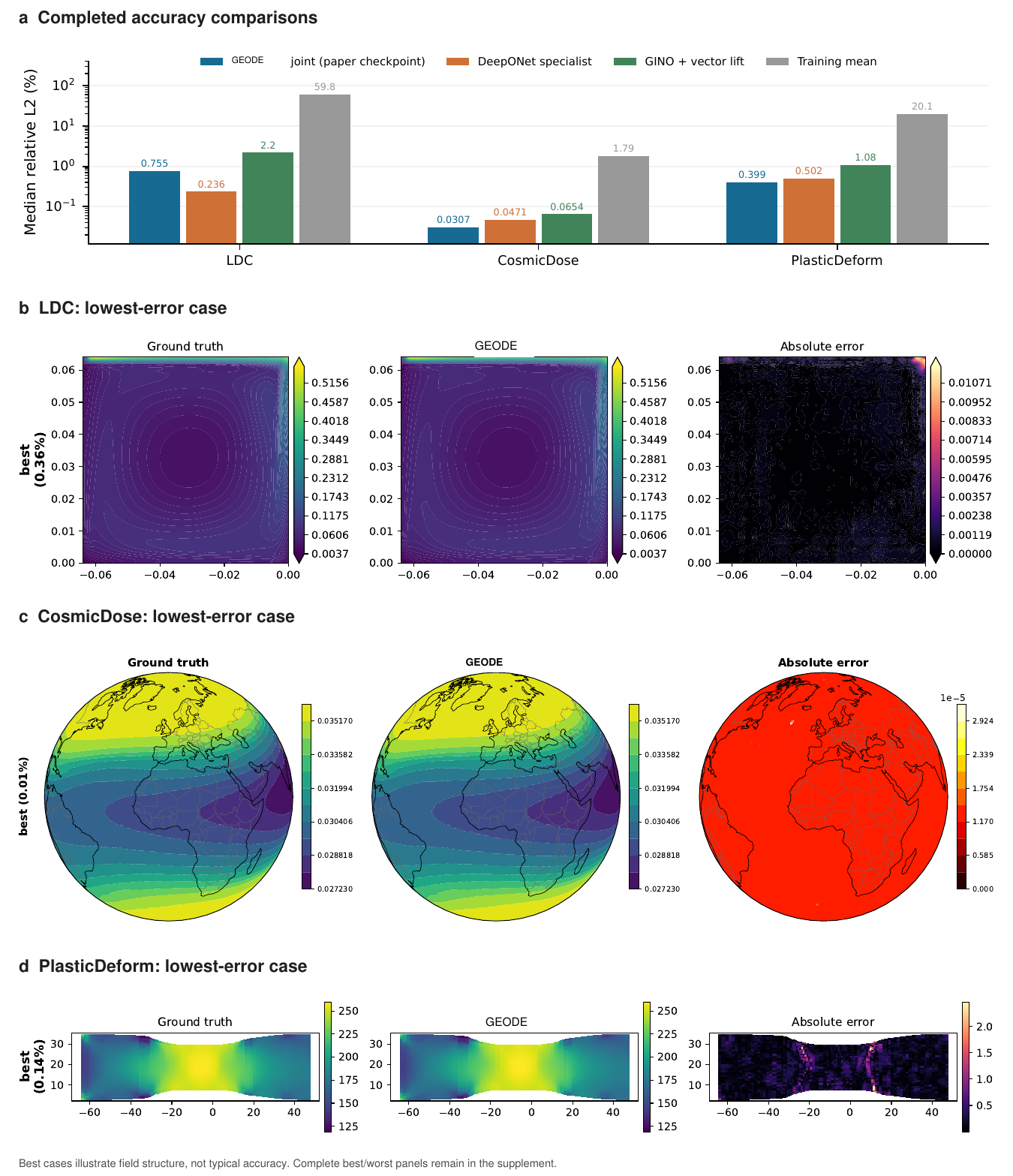}
  \caption{\textbf{Selected comparisons and physical field structure.} a, Clean DeepONet and GINO specialists, the clean joint GEODE checkpoint and training-mean controls; every checkpoint is selected on validation data disjoint from the test set. b--d, Lowest-error test cases of the paper checkpoint, preserving color bars, contours and spherical projection. These illustrate structure, not typical accuracy; complete best/worst panels remain in the supplement. GINO uses a vector lift and 30 epochs, without a matched-compute or convergence claim. Table~\ref{tab:operator_comparison} provides the broader comparison. This figure bears on requirements F1 to F4 and F6 of \Cref{sec:definition}; the DeepONet specialist in panel a remains more accurate on LDC.}
  \label{fig:joint_accuracy}
\end{figure}

We assessed predictive accuracy on $495$ LDC, $359$ CosmicDose and $100$
PlasticDeform test instances. The jointly trained model reaches median relative
$L^2$ errors of $0.755\%$, $0.031\%$ and $0.399\%$ (\Cref{fig:joint_accuracy}),
with linearly interpolated interquartile ranges of $0.54$--$1.34\%$,
$0.019$--$0.053\%$ and $0.25$--$2.52\%$. PlasticDeform is heavy tailed, with a mean
of $4.80\%$ and a 95th percentile of $34.7\%$ despite its median; the largest
relative errors occur for small stress fields, where the target norm in the
denominator becomes small. Normalizing root-mean-square error by the standard
deviation of the complete target field gives $0.020\,\sigma$, $0.0054\,\sigma$ and
$0.074\,\sigma$.

Optimization variability is material and constrains what may be concluded from
single runs. Two further seeds of the same recipe and budget give $2.87\%$ and
$1.97\%$ on LDC, $0.100\%$ and $0.134\%$ on CosmicDose, and $0.42\%$ and $0.39\%$ on
PlasticDeform (\Cref{tab:seeds}). The LDC and CosmicDose errors vary by up to
roughly fourfold across seeds, PlasticDeform does not, and the checkpoint reported
here, fixed before the replicates were trained, is the best of the three on LDC and
CosmicDose. We therefore treat the seed envelope, not the point estimate, as the
resolution of any accuracy comparison involving those two tasks, and we state below
where a reported difference falls inside it. Every subsequent experiment uses this
one checkpoint, so the routing, acquisition and transfer comparisons are
within-checkpoint contrasts in which the shared core is held fixed by construction.

The low errors are not explained by reproducing an average field. A per-node
training-mean predictor gives median errors of $59.80\%$, $1.79\%$ and $20.13\%$,
and the squared-error skill score of \Cref{eq:skill_methods} gives $0.9993$,
$0.9995$ and $0.9935$. This control is necessary but, for CosmicDose, not
sufficient, and we state the limitation directly. The CosmicDose reference fields
are produced by a forward radiation-transport model evaluated at fixed altitude,
atmospheric depth and water fraction, so that within this configuration the
time-varying fields form an essentially one-parameter family indexed by the solar
modulation state, and the neutron-monitor stations supplied as input are among those
from which that state is conventionally estimated
(\Cref{sec:methods_datasets}). A low error and a high skill score on this task are
therefore expected of any model that recovers one scalar and decodes it correctly,
and we do not present CosmicDose as evidence of difficulty, of sparse-to-dense
reconstruction, or of inference under underdetermination. Its role in this study is
as an interface: scattered station readings on a spherical chart, with an output
discretization an order of magnitude larger than the other tasks, exercising exactly
the conditioning modality and output geometry that a common field-tensor
representation would have to eliminate.

The predictions recover the defining structures of all three systems: the
latitudinal bands in CosmicDose, the primary recirculation in LDC and the
gauge-section stress concentration in PlasticDeform. Residuals concentrate near
steep gradients rather than reproducing the large-scale form of the targets.
Spectral analysis shows agreement with the PlasticDeform stress spectrum across the
resolved range and with the LDC kinetic-energy spectrum at large and intermediate
scales, with modest attenuation at the highest wavenumbers. Additional diagnostics
give a predicted-to-reference divergence ratio of $0.995$ for LDC, no PlasticDeform
node with negative predicted von Mises stress, and CosmicDose north--south asymmetry
of $0.0159$, matching the reference to three significant figures. These are
empirical consistency checks on quantities the training objective does not
constrain, not enforced guarantees. Complete error distributions, best- and
worst-case fields, spectra and diagnostic definitions are provided in the
Supplementary Information.

Comparison with existing operators requires separating predictive error, adaptation protocol and scientific input contract. DeepONet uses its parameter/history branch and coordinate trunk on these datasets. Our GINO implementation adds a learned vector-to-latent-field lift to the authors' operator, and must be identified as an adapted interface rather than a pretrained scientific foundation model. MORPH, PDEformer-2 and PDE-FM have broader published capabilities than a single fixed-grid formulation. We have now evaluated MORPH-Ti with explicit learned vector and query adapters; the other models remain at the source-level interface-assessment stage. We therefore record unexecuted comparisons as not yet evaluated; absence of a score does not establish incompatibility or poor accuracy. \Cref{tab:supp_compatibility} records the available implementation and input-contract evidence.

\input{sections/comparison_table}

The clean DeepONet specialists reach $0.236\%$, $0.0471\%$ and $0.502\%$ on LDC,
CosmicDose and PlasticDeform. Corrected GINO with a learned vector lift reaches
$2.195\%$, $0.0654\%$ and $1.083\%$. The clean joint GEODE checkpoint reaches
$0.755\%$, $0.0307\%$ and $0.399\%$. DeepONet outperforms it on LDC by a factor of
approximately $3.2$, as do Transolver, FNO and adapted DPOT. On PlasticDeform, all
three GEODE seeds ($0.39$ to $0.42\%$) are below every other model evaluated, the
nearest being DeepONet at $0.502\%$. On CosmicDose the reported checkpoint
($0.0307\%$) is below every other model, the nearest being GNOT at $0.0432\%$, but the
two replicate seeds ($0.100\%$ and $0.134\%$) are not; under the seed-envelope rule
stated above, the CosmicDose ranking is unresolved. These are individual runs with different
capacities and schedules, $200$ epochs for DeepONet and $30$ for GINO, rather than
matched-compute or independent-seed comparisons, and GINO validation loss still
improves near its budget endpoint on LDC and PlasticDeform, so its attainable
accuracy is not established. The LDC gap to DeepONet exceeds the GEODE seed envelope
of $0.755$ to $2.87\%$ and should be read as real.

This comparison defines the scope of the accuracy claim, and it also defines the
alternative against which this architecture must justify itself. The most economical
competitor is not another foundation model but a collection of small independent
operators. DeepONet specialists for all five tasks total $1.970$ million parameters
against GEODE's $198.5$ million and preserve earlier predictions trivially because
they share nothing, and task-specific operators, a DeepONet specialist on LDC and
same-architecture specialists on HeatExchanger and Subchannel, are more accurate than
GEODE on three of the five problems. A
single shared DeepONet with $0.422$ million parameters, using padded inputs, explicit
task indicators and a common branch and coordinate trunk, attains $0.7425\%$,
$0.0529\%$ and $0.7253\%$ across the three pretraining tasks, so joint prediction
over these fixed geometries is likewise not an argument for the expert architecture.
We state this without qualification: if the requirement is the lowest error on a
known, fixed set of tasks, a collection of small specialists is the appropriate
choice, and nothing in this paper contradicts that.

What separate models cannot provide is reuse. They preserve earlier functions by
having no shared state to disturb, which is why the third clause of F5 requires
demonstrated benefit from the pretrained core against a randomized control of
identical architecture. Sections~\ref{sec:continual_acquisition}
and~\ref{sec:conditional_transfer} test exactly that, and report where it holds and
where it does not: the frozen library is worth a factor of $1.3$ over a norm-matched
random library when a new component's data are scarce, is indistinguishable from
random content at $10\%$ of that component's data, and in the subchannel case
determines whether the interface departs from a constant-level solution at all. The
supported motivation for sharing is therefore controlled extension with measured
reuse, not universal specialist superiority, and we do not claim the latter anywhere
in this work.

The clean POD-plus-ridge rerun fits its basis and preprocessing on training data only, selects rank and ridge penalty on independent validation, and evaluates test data after selection (\Cref{tab:cx_clean_pod}). Its median errors are $10.9836\%$, $0.3378\%$ and $20.1056\%$. Historical values of $8.58\%$, $0.35\%$ and $20.32\%$ came from a different internal-holdout procedure whose basis already included the held-out rows; they are retained as historical evidence rather than mixed into the clean comparison.

The shared DeepONet described above is the direct test of joint learning without
wavelet experts. Its LDC median is marginally lower than that of the reported GEODE
checkpoint ($0.7425\%$ against $0.755\%$) and lower than both GEODE replicate seeds,
whereas its PlasticDeform median is higher than all three. This comparison does not
test sequential acquisition or retention.

All six additional operator families have completed the bounded evaluation (\Cref{tab:cx_wave_details}). FNO, WNO, GNOT and Transolver use separate task-specific models; shared DeepONet and adapted NCWNO are trained jointly. The adapters and author implementations are documented in the supplement. Every score passed an independent CPU checkpoint replay and checks of the full saved error vectors. The wall-time cap limits interpretation: adapted NCWNO completed only four joint epochs, and all three GNOT runs also reached the cap. These rows measure performance under the stated budgets, not converged architecture rankings. The adapted MORPH and DPOT comparisons are reported below; MPP, Poseidon and PDEformer-2 remain unevaluated in the supplementary comparison.

The first pretrained foundation-model control uses the public MORPH-Ti core with a learned vector-to-latent lift and bilinear coordinate readout (\Cref{tab:cx_morph}). On LDC, fully fine-tuned pretrained and identically configured random-initialization models attain $1.1531\%$ and $1.0889\%$, respectively. With the cores frozen, the corresponding errors are $4.7442\%$ and $35.7849\%$. Thus pretraining helps the frozen-feature adapter in this experiment, but provides no full-training accuracy advantage under the stated budget. This is a learned-modality adaptation of MORPH, not native zero-shot prediction from a physical field history. Different stopping epochs and a single seed preclude a statistical or matched-compute superiority claim.

For CosmicDose, full-training pretrained and random-initialization errors are $0.0787\%$ and $0.0632\%$, respectively. For PlasticDeform, full-training pretrained and random-initialization errors are $1.3073\%$ and $0.8914\%$, respectively.

The second pretrained foundation-model control uses the public DPOT-Ti core with a learned vector-to-latent lift and bilinear coordinate readout (\Cref{tab:cx_dpot}). On LDC, fully fine-tuned pretrained and identically configured random-initialization models attain $0.1548\%$ and $0.1991\%$; both have lower median error than the other tabulated operators on this task, including the DeepONet specialist, and both stopped at the wall-clock budget rather than at convergence. For CosmicDose the corresponding errors are $0.0467\%$ and $0.0521\%$, and for PlasticDeform $0.8485\%$ and $0.8354\%$. Pretraining gives a lower median on LDC and CosmicDose in these runs, but not on PlasticDeform. The random LDC arm has lower mean and P95 error, so the median advantage is not uniform across the distribution. On PlasticDeform both DPOT arms trail the clean DeepONet specialist ($0.502\%$) and the clean GEODE checkpoint ($0.399\%$). These three tasks do not isolate mesh structure as the cause of the differences. As with MORPH, this is a learned-modality adaptation of a field-history model rather than native prediction, and single seeds with unequal stopping epochs preclude a statistical superiority claim in either direction.

\subsection{Expert routing separates shared and system-specific computation}
\label{sec:routing}

\begin{figure}[htbp]
  \centering
  \includegraphics[width=\textwidth]{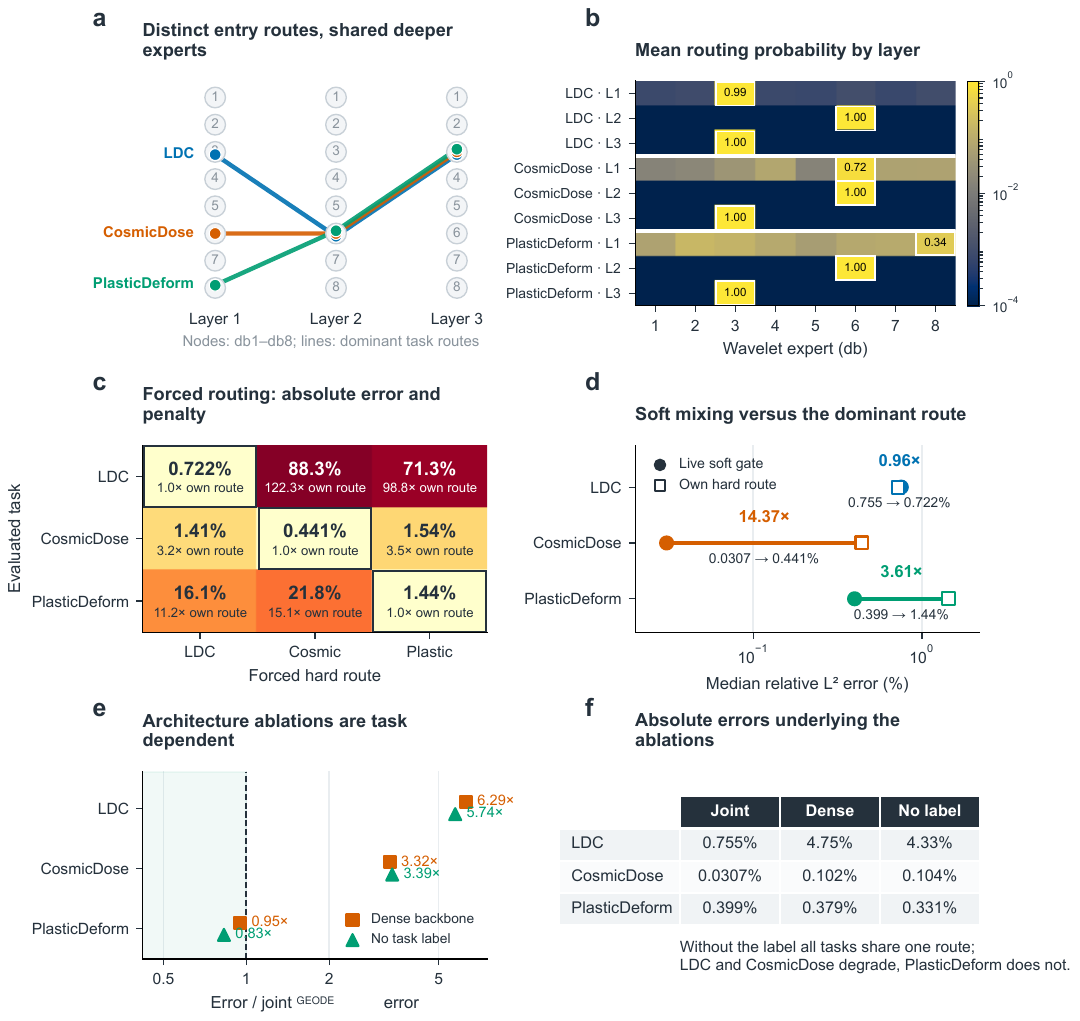}
  \caption{\textbf{Task-dependent early routing and shared deeper computation.} a, Dominant pathways derived from saved gates. b, Mean gate probabilities on a logarithmic color scale; outlined cells identify dominant experts. c, Forced-route errors and penalties relative to each task's own hard route. d, Live soft mixing versus that hard route. e,f, Dense-backbone and no-label ablations, shown as ratios to joint GEODE and as absolute errors. Ratios below one favor the ablation. The CosmicDose ablation errors lie within the range spanned by independent training seeds of the joint model (\Cref{tab:seeds}) and support no conclusion on that task; only the LDC effects exceed the seed envelope. All panels use the paper checkpoint, including the completed no-label rerun; no additional trials or uncertainty estimates are implied. This figure bears on requirement F1 of \Cref{sec:definition}; gate probabilities alone do not establish necessity, which is why the forced-route and randomized-library interventions are reported alongside them.}
  \label{fig:routing_mechanism}
\end{figure}

We next examined how the shared library distributes computation across tasks (\Cref{fig:routing_mechanism}). In the first layer, LDC, CosmicDose and PlasticDeform select db3, db6 and db8 as dominant experts, with effective expert counts of $1.1$, $3.0$ and $6.5$: the LDC gate is nearly one-hot, whereas the PlasticDeform gate spreads its weight over most of the library. In the second and third layers, all three converge on db6 followed by db3, with effective counts of $1.0$. Task differentiation is therefore concentrated near the heterogeneous interfaces, whereas deeper processing follows a common route. This organization is consistent with early task-specific transformation followed by computation in an aligned latent representation, although gate probabilities alone do not establish necessity.
We tested necessity by overriding the gates at inference. LDC has an error of $0.722\%$ on its own hard route and $88.3\%$ and $71.3\%$ on the CosmicDose and PlasticDeform routes. CosmicDose increases from $0.441\%$ to $1.41\%$ and $1.54\%$, while PlasticDeform increases from $1.44\%$ to $16.1\%$ and $21.8\%$. Because the tasks share their dominant deeper experts, these interventions isolate the first-layer assignment. Relative to the live gate, cross-task substitution increases error by factors of approximately $13$ to $117$, showing that routing is load-bearing. The live soft gate is better than the task's own one-hot route by factors of $14$ and $3.6$ for CosmicDose and PlasticDeform, whose first-layer gates are diffuse, indicating a measurable contribution from lower-weight experts; for LDC, whose first-layer gate places $0.994$ of its weight on db3, the hard route is comparable ($0.96$). Applied to a norm-matched random library, all forced routes remain uniformly inaccurate and differ by at most $25\%$, with no task-aligned route pattern. Thus routing carries information only when the experts do.

Matched ablations locate the contribution of the expert mixture more narrowly than
joint accuracy alone would suggest, and we read them against the seed envelope of
\Cref{tab:seeds}. Replacing the mixture with a dense backbone without experts
($22$ million parameters) changes the errors from $0.755\%$, $0.031\%$ and $0.399\%$
to $4.75\%$, $0.102\%$ and $0.379\%$. Removing the explicit task label gives
$4.33\%$, $0.104\%$ and $0.331\%$, and the three tasks then select the same dominant
route (db7, db7, db1). Only the LDC effects lie outside the seed envelope, and even
there they exceed the weakest seed by factors of approximately $1.7$ and $1.5$ from
single runs, so they are indicative rather than established. The CosmicDose ablation
values of $0.102\%$ and $0.104\%$ fall within the range spanned by the GEODE seeds
themselves ($0.031$ to $0.134\%$) and therefore support no conclusion about the
expert mixture or the task label on that task. PlasticDeform is unaffected by the
dense substitution and improves by $17\%$ without the task label. The defensible
statement is consequently that the expert mixture and explicit task conditioning
benefit the structured cavity problem, are not resolvable on CosmicDose at the
present number of training runs, and are not required for PlasticDeform. We do not
claim that the expert library is necessary for joint accuracy; the evidence that the
library carries content is the forced-route intervention above and the
randomized-library audit of \Cref{sec:conditional_transfer}, both of which compare
learned against randomized expert weights while holding the architecture fixed. When
LDC is re-presented through a new interface and gate, the model converges on a
different route (db8, db5, db5 against the original db3, db6, db3) while reaching a
test error within a factor of $1.2$ to $1.4$ of the original across four runs.
Expert indices should therefore be interpreted as learned allocations, not as a
taxonomy connecting wavelet order to physics.
Together, these analyses show that the pretrained tasks use causally important but non-unique pathways. A new gate can recombine existing computations, but routing alone cannot reconcile a new input modality or output representation with the shared latent space.

\subsection{New energy-system components are acquired without altering deployed surrogates}
\label{sec:continual_acquisition}
\label{sec:transfer}

\begin{figure}[htbp]
  \centering
  \includegraphics[width=\textwidth]{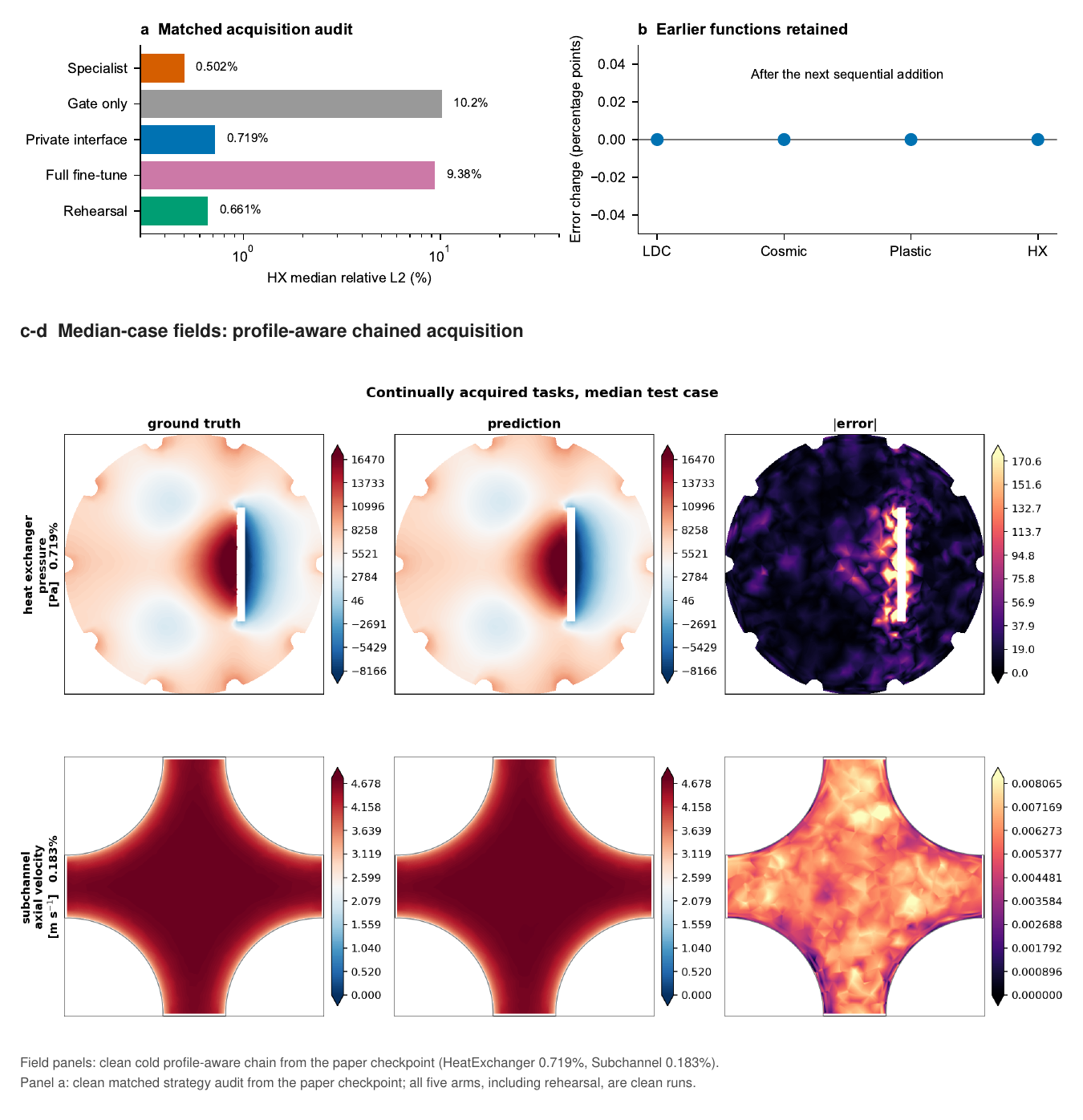}
  \caption{\textbf{Sequential acquisition with preserved earlier functions.} Panel a reports the matched HeatExchanger strategy audit; panel b shows that earlier-task errors are unchanged after each addition. Panels c--d show the median-case HeatExchanger pressure and Subchannel velocity fields from the chained protocol on their native meshes (clean cold profile-aware chain, HeatExchanger $0.719\%$ and Subchannel $0.183\%$). This figure tests requirement F5 of \Cref{sec:definition}. Preservation of earlier outputs is a structural consequence of parameter disjointness and not a learned property.}
  \label{fig:continual_acquisition}
\end{figure}

We withheld HeatExchanger and Subchannel from pretraining and introduced them sequentially (\Cref{fig:continual_acquisition}). HeatExchanger maps a $102$-dimensional vector to three fields on an unstructured domain containing $3{,}977$ nodes and an internal plate. Subchannel predicts three fields on a cross-shaped $1{,}733$-node coolant domain bounded by four quarter-circle fuel rods. For each task, the library and earlier interfaces are frozen. Only a new interface and gate are trained without replay; together they contain $3.9$ million parameters.

Routing alone does not acquire HeatExchanger, and the failure is informative in two
directions at once. Updating only the task embedding and the $2.3$-million-parameter
gate network leaves the new task at $10.15\%$ and, because that gate network is
shared, simultaneously degrades the pretrained tasks to $22.6\%$, $0.306\%$ and
$6.0\%$. Router capacity is not the binding constraint: the audit varies it from the
task embedding alone through rank-$16$ adapters to a private full gate, and the new
task remains between approximately $9.5\%$ and $10.3\%$ throughout. Under the same
full-data protocol, updating the complete private interface with the experts frozen
reduces the error to $0.719\%$, against $0.502\%$ for a specialist of the same
architecture and budget. Acquisition therefore requires the encoder, output
projection, skip paths and geometry decoder in addition to routing, and it must not
write to shared state. This result runs against the expectation, natural in the
mixture-of-experts setting and implicit in routing-based approaches to continual
operator learning, that a new problem can be absorbed by recombining an existing
library. A gate can reweight computations; it cannot reconcile a conditioning vector
of different physical meaning, a different number of output variables and a new query
geometry with a latent representation built for other tasks.

The two less restrictive strategies are worse on both the new and the old tasks.
Training every parameter on HeatExchanger alone yields $9.38\%$ and moves the LDC,
CosmicDose and PlasticDeform errors from $0.755\%$, $0.031\%$ and $0.399\%$ to
$16.4\%$, $0.905\%$ and $5.65\%$, degradations of approximately $22$, $29$ and $14$
times. Rehearsal reaches $0.661\%$ on HeatExchanger within the same budget, slightly
better than the frozen interface, but requires retention of all earlier data and
moves LDC to $2.69\%$. That shift lies within the seed spread of retraining
(\Cref{tab:seeds}) and we do not characterize it as large. The relevant property is
categorical rather than quantitative: the earlier predictions change, and a surrogate
that has already been verified in service cannot accommodate a silent change of any
magnitude without revisiting that verification.

Interface adaptation does not change earlier outputs. The reason is structural rather
than learned. Acquisition updates a parameter vector disjoint from every parameter,
preprocessing statistic and geometry state on which an earlier task depends, so the
earlier predictor is a function of quantities that were not modified, so its outputs
are unchanged in exact arithmetic, and re-evaluation after each step reproduces the
earlier test errors exactly at reported precision (\Cref{sec:methods_acquisition}). We therefore describe this as parameter isolation
rather than as resistance to catastrophic forgetting, which would imply a learned
property. The same construction guarantees nothing about whether the frozen library
is useful to the new task, which is a separate question tested in
\Cref{sec:conditional_transfer}.

The sequential chain starts from the full-data interface run above: HeatExchanger reaches $0.719\%$, compared with $0.507\%$ for a specialist trained for twice as many epochs, and the pretrained tasks retain $0.755\%$, $0.031\%$ and $0.399\%$ exactly. Because the HeatExchanger interface is frozen in turn, its error is unchanged when Subchannel is added. The Subchannel step reaches $0.083\%$ with the standard loss and $0.183\%$ with the profile-aware loss, against $0.034\%$ for its specialist, which reaches the same value with and without the profile term. On the earlier checkpoint the two variants gave $0.161\%$ and $0.063\%$, so the profile term helped there and hurts here; the specialist is indifferent to it in both cases.
These results establish a bounded form of continual acquisition. GEODE adds a heterogeneous system by training $2.1\%$ of its parameters without replay or modification of earlier functions. The mechanism is not embedding-only or routing-only adaptation, and each new modality incurs a private interface cost that accumulates with task number.

\subsection{Reuse across energy components is conditional and component-specific}
\label{sec:conditional_transfer}
\label{sec:library_audit}

\begin{figure}[htbp]
  \centering
  \includegraphics[width=\textwidth]{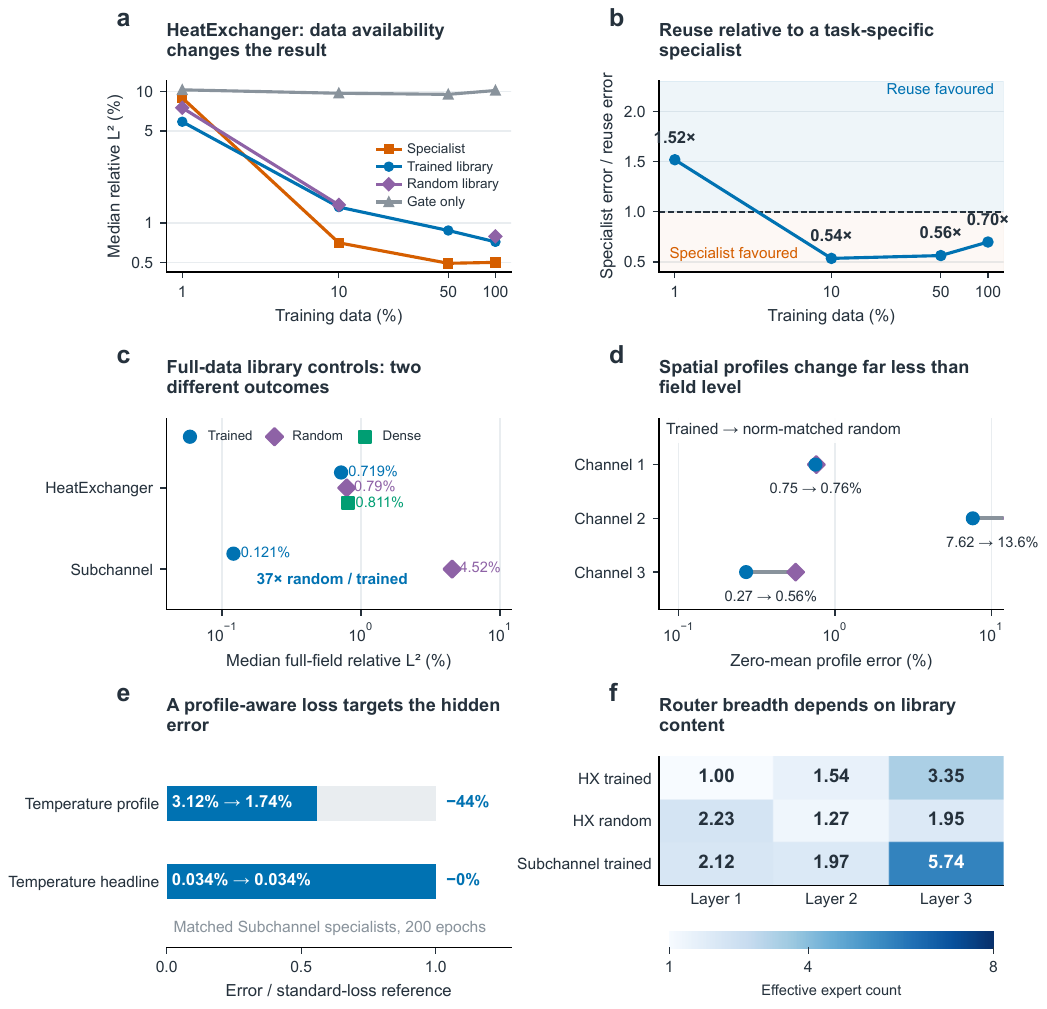}
  \caption{\textbf{Transfer depends on data availability, target and field component.} a, HeatExchanger data-fraction sweep. b, Specialist-to-trained-library error ratios derived from a; values above one favor reuse. c, The separate full-data audit of trained, norm-matched random and dense libraries. d, Subchannel zero-mean-profile errors before and after library randomization. e, The temperature-channel profile-loss audit on matched Subchannel specialists, normalized to the standard-loss reference. f, Effective expert counts in the adapted routers. The HeatExchanger sweep, the full-data controls and the Subchannel library audit share one protocol on the paper checkpoint. Missing runs are omitted, and the single-run comparisons do not establish statistical significance. This figure tests the reuse clause of requirement F5 of \Cref{sec:definition}, the clause that distinguishes a shared core from a collection of independent surrogates.}
  \label{fig:conditional_transfer}
\end{figure}

Exact preservation does not establish that a frozen library provides useful information. We therefore varied the HeatExchanger data fraction and compared trained experts with norm-matched random experts (\Cref{fig:conditional_transfer}). At $1\%$ of the data, the trained library reaches $5.88\%$, compared with $8.92\%$ for a specialist and $7.50\%$ for the randomized library. At $10\%$, the trained and randomized libraries reach $1.32\%$ and $1.37\%$, while the specialist reaches $0.707\%$. At $50\%$ and $100\%$, the specialist obtains $0.494\%$ and $0.502\%$, whereas the frozen-library interface obtains $0.877\%$ and $0.719\%$. Transfer therefore helps in the lowest-data regime, where the learned content is worth a factor of $1.3$ over a random library of the same architecture, and the specialist becomes more accurate once sufficient task data are available. Each entry is a single run. Gate-only adaptation remains between approximately $9.5\%$ and $10.3\%$ throughout.
The value of the learned library differs between targets. At full data and under the same protocol, HeatExchanger reaches $0.719\%$ with trained experts, $0.790\%$ with random experts and $0.811\%$ with a dense backbone that has no expert library; the learned expert content is worth about $10\%$ for this task. Subchannel differs in kind. Predicting the training-mean field for every test sample gives $5.12\%$, because the temperature level varies by $34$~K between samples around a $574$~K mean, so a model that learns the spatial profile but not the sample-specific level scores about $5\%$. Under the same protocol on the paper checkpoint, the trained library reaches $0.121\%$ and the randomized library $4.52\%$, a factor of $37$: the interface over the trained library learns the sample-specific level (its temperature-level error is $0.5$~K) and the interface over the randomized library does not ($26$~K), so the latter sits at the constant-level solution. Two further seeds of each arm (\Cref{tab:seeds}) show that this is an escape rate rather than a fixed ratio: the trained library learns the level in all three seeds ($0.068\%$ and $0.073\%$), whereas the randomized library stays at the constant-level solution in a second seed ($4.653\%$) and escapes it in the third ($0.160\%$); the HeatExchanger interface is seed-stable ($0.657\%$ and $0.669\%$). The difference is a matter of escape rather than degree: the trained-library run stayed on the $5\%$ plateau for $55$ epochs before leaving it, whereas the same run from the earlier checkpoint left it within ten epochs and reached $0.066\%$ against $4.92\%$.

Decomposing Subchannel predictions into spatial mean and zero-mean profile identifies the transferred component. Profile errors over the trained library are $0.75\%$, $7.6\%$ and $0.27\%$, compared with $0.76\%$, $13.6\%$ and $0.56\%$ over the randomized library; the temperature-level errors are $0.5$~K and $26$~K. The profile errors differ by factors of $1.0$, $1.8$ and $2.1$ across the three channels while the complete-field errors differ by $37$, so the complete-field difference is first the difference between learning and not learning the field level, and second a better temperature profile. This is important for temperature, whose mean is approximately $574$~K while its spatial variation spans about $21$~K.
HeatExchanger uses effective expert counts of $1.0$, $1.5$ and $3.4$ with the trained library and $2.2$, $1.3$ and $2.0$ after randomization. Subchannel uses $2.1$, $2.0$ and $5.7$ with the trained library and $1.3$, $2.4$ and $2.5$ after randomization. The router is sensitive to library content, but dense recombination is not uniformly required.

Two conclusions follow, and they are of different kinds. The first concerns this
architecture: reuse of the pretrained library is conditional rather than general. It
is worth a factor of $1.3$ over a norm-matched random library when a new component's
data are scarce, is indistinguishable from random content at $10\%$ of that
component's data, is worth approximately $10\%$ at full data, and in the subchannel
case governs whether the interface escapes a constant-level solution at all, in three
of three seeds against one of three. This is the property that separates a shared
foundation from a collection of independent surrogates, since independent surrogates
preserve earlier functions perfectly while offering no reuse whatsoever, and it is
the reason we require a randomized-library control rather than treating the benefit
of pretraining as given.

The second conclusion is a measurement point that does not depend on this
architecture. The subchannel comparison produces a $37$-fold difference in headline
relative error between two models that differ only in whether their frozen experts
carry learned content. Decomposing the prediction into spatial mean and zero-mean
profile shows that the difference is concentrated principally in the field level:
the profile errors differ by factors of $1.0$, $1.8$ and $2.1$ across the three
channels, while the temperature-level errors are $0.5$~K and $26$~K against a mean of
approximately $574$~K with a spatial variation of about $21$~K. A model that learns
the spatial profile but not the sample-specific level scores about $5\%$, as the
mean-field control at $5.12\%$ shows. Reported alone, the $37$-fold figure would have
suggested a transfer of spatial structure an order of magnitude larger than the
decomposition supports. The reason is an identity rather than a property of this
model: full-field relative $L^2$ error equals the error relative to spatial variation
divided by a scale factor $\kappa$ that grows with the ratio of field mean to field
variation (\Cref{eq:scale_factor}), and $\kappa$ ranges from approximately $2$ to
$121$ across the tasks evaluated here. For temperature, pressure and dose fields,
which carry large offsets, a headline relative error can therefore be insensitive to
precisely the lower-amplitude variation that determines gradients, hot spots and
margins. We report errors relative to both the complete field and its variation
throughout, and we note that adding a loss term after subtracting the spatial mean
reduces the temperature profile error of a subchannel specialist from $3.12\%$ to
$1.74\%$ while leaving its headline error unchanged at $0.034\%$, which is a direct demonstration that the headline metric was not measuring the quantity improved.

\subsection{Geometry-adaptive decoding evaluates fields on component meshes at deployment cost}
\label{sec:geometry_efficiency}

\begin{figure}[htbp]
  \centering
  \includegraphics[width=\textwidth]{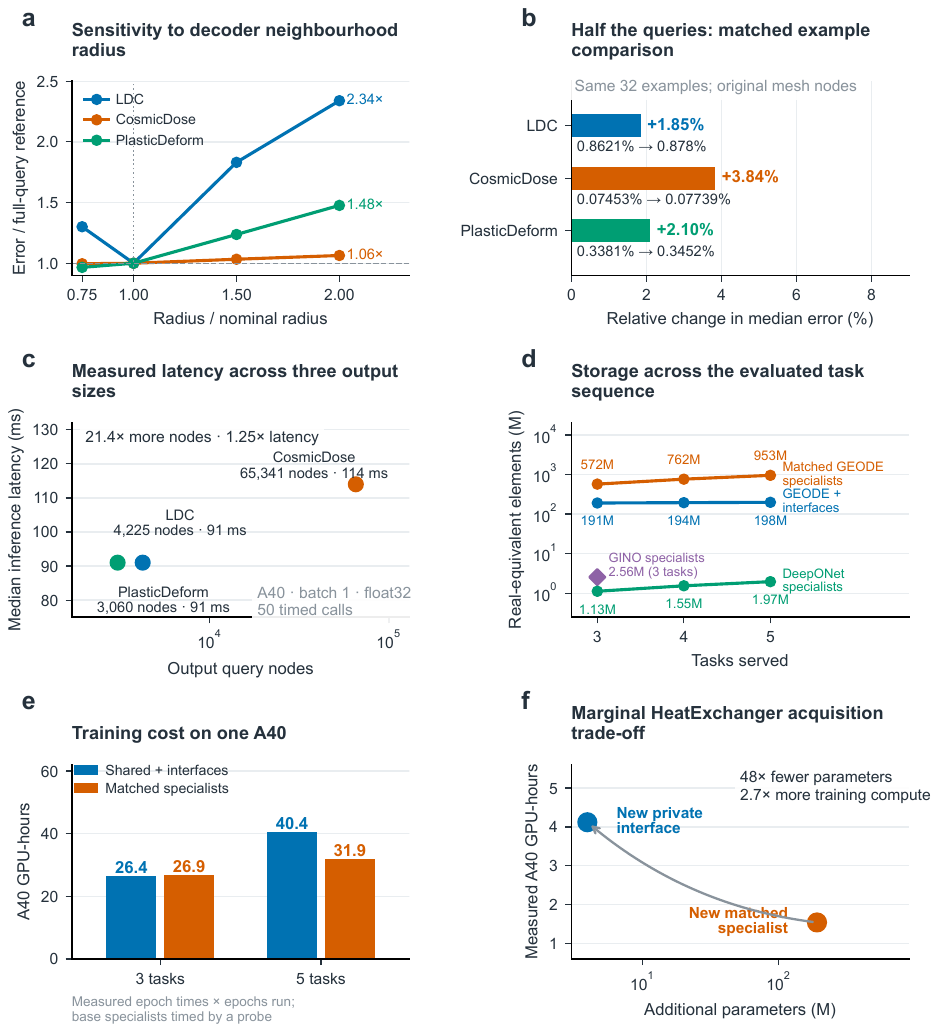}
  \caption{\textbf{Query sensitivity and deployment costs.} a,b, Radius sensitivity and half-query errors on the same 32 examples per task. Subsets use original nodes, not unseen geometries. c, Batch-one A40 latency: 50 synchronised calls after 15 warm-ups. Different tasks are compared, not a controlled scaling sweep. d, Log-scale storage, including measured DeepONet and GINO specialists; GINO covers three tasks only. e,f, Cumulative and marginal HeatExchanger training costs. GPU-hours are measured per-epoch times on one A40 multiplied by the epochs run; the base-task specialists are priced from a timing probe on the same hardware. Matched-GEODE storage savings do not extend to the smaller baselines in d; marginal costs are not extrapolated across tasks.}
  \label{fig:geometry_efficiency}
\end{figure}

We finally examined sensitivity to query discretization, neighborhood construction and output size (\Cref{fig:geometry_efficiency}). For each pretrained task, we randomly selected $50\%$ of the original nodes, rebuilt the graph and evaluated the unchanged model at those coordinates. The same 32 examples per task are used for the full-query and reduced-query audit. At its calibrated radius, LDC changes from $0.862\%$ to $0.878\%$, CosmicDose from $0.0745\%$ to $0.0774\%$, and PlasticDeform from $0.338\%$ to $0.345\%$. These full-query reference values are distinct from the full-test-set headline medians. These differences also reflect the changed set of locations contributing to the error norm. This experiment shows stability to an altered query set, not generalization to coordinates or geometries absent from training.

The decoder radius is calibrated to provide approximately $17$--$21$ source neighbors per query point, with means of $17.5$, $21.2$ and $20.9$. Rescaling it by $0.75$, $1.5$ and $2.0$ changes LDC from its nominal $0.862\%$ to $1.12\%$, $1.58\%$ and $2.02\%$; CosmicDose from $0.0745\%$ to $0.0742\%$, $0.0770\%$ and $0.0793\%$; and PlasticDeform from $0.338\%$ to $0.327\%$, $0.418\%$ and $0.499\%$. CosmicDose and PlasticDeform are stable to a moderate reduction in support, LDC is not, and all three degrade gradually as the neighborhood expands. The radius remains a task-calibrated hyperparameter.
Inference grows weakly with output size over the evaluated range. With batch size one in float32 on one NVIDIA A40, PlasticDeform requires $91$~ms for $3{,}060$ nodes, LDC requires $91$~ms for $4{,}225$ nodes, and CosmicDose requires $114$~ms for $65{,}341$ nodes. A $21$-fold increase in output points therefore increases latency by $25\%$. These medians use $50$ calls after $15$ warm-up calls with device synchronization.

System-level cost presents a different trade-off. The GPU-hour values below are measured per-epoch times of the clean runs on one NVIDIA A40 multiplied by the epochs each run used; the three base-task specialists, which were trained on other hardware, are priced from a timing probe of the same configurations on the same A40. Joint pretraining takes $26.4$ GPU-hours for $120$ epochs, compared with $26.9$ GPU-hours summed across three matched GEODE specialists at the same budget. The joint model stores $190.6$ million parameters in one artifact, compared with $571.8$ million across three specialists. In the HeatExchanger marginal-cost example, an interface adds $3.9$ million parameters and $4.1$ GPU-hours for $100$ epochs, whereas its matched specialist adds $190.6$ million parameters and $1.5$ GPU-hours for $200$ epochs, because every interface epoch still traverses the frozen library. The Subchannel step costs $9.9$ GPU-hours against $3.5$ for its specialist. At five tasks, GEODE plus two interfaces contains $198.5$ million parameters and requires $40.4$ cumulative GPU-hours, compared with $953.0$ million and $31.9$ GPU-hours across five specialists. GEODE exchanges additional adaptation compute for lower marginal storage and one deployed model. The measured DeepONet specialists require 1.126, 1.548 and 1.970 million parameters for three, four and five tasks; the three GINO specialists require 2.562 million. Thus the shared model saves storage relative to matched GEODE specialists, but uses substantially more storage than these smaller operators.
Against the simulations used to generate the data, inference is faster by approximately $535\times$ to $16{,}500\times$, but this is a general surrogate advantage rather than evidence specific to GEODE. The solver timings use different software and hardware from the A40 measurements and are therefore indicative rather than controlled. Overall, the decoder supports several native geometries, tolerates moderate query changes and adds comparatively little latency as output size increases. It does not establish zero-shot generalization to new geometries. The principal system-level advantage is marginal: additional systems can be served by adding small interfaces to one deployed model, a trade-off relevant to many-task settings but potentially unattractive for a single system.

%% file: sections/comparison_table.tex
In the comparisons below, ``adapter'' denotes an input--output compatibility module added by us: a learned map from a task's parameter or history vector to the model's latent representation, followed by a readout at the supplied output coordinates. These compatibility adapters are distinct from GEODE's private continual-acquisition interface, which also includes task-specific skip paths, a geometry decoder and a fresh gate.

\begin{table}[htbp]
  \centering
  \caption{Median physical relative $L^2$ error (\%) from completed runs. ST denotes separate single-task training; pretr. FT denotes full fine-tuning of pretrained weights; POD denotes proper orthogonal decomposition. $^{\dagger}$ the clean single-task PlasticDeform run did not converge under the shared recipe (its validation loss sat at $0.90$ for the first sixty epochs) and is reported as obtained; the historical specialist reached $0.34\%$. The joint GEODE row is the clean signed-distance-enabled checkpoint, selected on validation data only. DeepONet and GINO use a training-only PlasticDeform validation split; GINO uses corrected coordinates and a learned vector-to-field lift. Separate DeepONet specialists used 200 epochs and GINO 30. The additional clean baselines use up to 100 epochs, validation-based stopping and a 20-minute training/validation cap; their exact budgets and adapters are detailed in the supplement. Differing schedules and capacities prevent a matched-compute or converged-model ranking.}
  \label{tab:operator_comparison}
  \scriptsize
  \setlength{\tabcolsep}{3pt}
  \renewcommand{\arraystretch}{1.15}
  \begin{tabular}{@{}llccc@{}}
    \toprule
    \textbf{Method} & \textbf{Protocol} & \textbf{LDC} & \textbf{CosmicDose} & \textbf{PlasticDeform} \\
    \midrule
    Per-node training mean & Control & 59.80 & 1.79 & 20.13 \\
    POD plus ridge & ST, clean & 10.9836 & 0.3378 & 20.1056 \\
    DeepONet & ST, clean & 0.236 & 0.0471 & 0.502 \\
    GINO + vector lift & ST, clean & 2.195 & 0.0654 & 1.083 \\
    FNO + adapters & ST, clean & 0.2708 & 0.0952 & 0.5034 \\
    WNO + adapters & ST, clean & 1.1889 & 0.0752 & 2.2866 \\
    GNOT + vector input & ST, clean & 0.4016 & 0.0432 & 1.5610 \\
    Transolver + vector input & ST, clean & 0.2189 & 0.0791 & 1.6689 \\
    Shared DeepONet & Joint, clean & 0.7425 & 0.0529 & 0.7253 \\
    MORPH-Ti + adapters & ST, pretr. FT & 1.1531 & 0.0787 & 1.3073 \\
    MORPH-Ti + adapters & ST, random init. & 1.0889 & 0.0632 & 0.8914 \\
    DPOT-Ti + adapters & ST, pretr. FT & 0.1548 & 0.0467 & 0.8485 \\
    DPOT-Ti + adapters & ST, random init. & 0.1991 & 0.0521 & 0.8354 \\
    NCWNO + adapters & Joint, clean & 11.3864 & 0.9450 & 5.9175 \\
    GEODE architecture & ST, clean & 0.385 & 0.038 & $11.8^{\dagger}$ \\
    GEODE & Joint, clean & \textbf{0.755} & \textbf{0.0307} & \textbf{0.399} \\
    \bottomrule
  \end{tabular}
\end{table}

%% file: sections/discussion.tex
\section{Discussion}
\label{sec:discussion}

The principal result of this study is not that several physical tasks can be fit by one network. A substantially smaller shared DeepONet already shows that joint prediction alone is insufficient to establish the value of the proposed architecture, and independent specialists show that preservation of earlier tasks does not establish sharing. The informative question is what remains after these simpler explanations are removed. Three results survive that test. First, the interface through which a scientific problem reaches a pretrained model can change the reported error by more than an order of magnitude with the backbone unchanged. Second, preservation and pretrained reuse are experimentally distinct: parameter isolation guarantees the former, whereas randomized-library controls are required to establish the latter. Third, full-field relative $L^2$ error can substantially compress differences in spatial-structure error when field level dominates the target norm. Together, these observations define the evidentiary contribution of GEODE more strongly than its aggregate multi-task accuracy. The evidence here indicates that
the cost is not small: an interface convention moved one baseline's reported error by
more than an order of magnitude with the operator unchanged. Corpus scale alone does
not remove the need for such an interface, because a model that consumes gridded field
histories does not acquire the ability to consume a scattered monitor record by
observing more gridded field histories. Our adapted comparisons illustrate both sides
of this point. Supplied with a learned interface, the pretrained DPOT core became the
most accurate model evaluated on LDC, which shows that corpus-axis models can be highly
effective once an interface is provided; the capability to read the lid-velocity
history nonetheless came from the interface we added, and the adapted model is a
per-task specialist that neither shares computation with the other tasks nor extends
with retention. The same adaptation of MORPH reproduces, in an independently
pretrained model, the conditionality of reuse that we report for GEODE: with its core
frozen, the pretrained model reached $4.74\%$ on LDC against $35.78\%$ for an
identically adapted random core, whereas with every parameter trained it held no
advantage ($1.15\%$ against $1.09\%$).
We therefore pursued breadth along the interface axis and treated the scientific
interface as an internal, trained and audited component. A shared expert library
coupled to task-specific interfaces supports predictive mappings for confined flow,
global radiation reconstruction and path-dependent structural response, and
subsequently expands to compact heat-exchanger flow and pressurized-water-reactor
subchannel transport. These tasks are connected by an application domain rather than
by a common equation or data representation: they involve fluid transport, radiation
monitoring and solid mechanics; parameter-to-field, sparse-observation-to-field and
history-to-field inference; and Cartesian, spherical and irregular finite-element
outputs. Evaluated against the six requirements of \Cref{sec:definition}, the model
satisfies the four representational requirements and the extension requirement,
including its reuse clause, and partially satisfies the competitiveness requirement,
since task-specific operators remain more accurate on three of the five problems. We
regard that outcome, rather than any single accuracy figure, as the substance of the
claim. It establishes that a reusable computational foundation for this domain can be
built without first eliminating the heterogeneity that characterizes it, and it
identifies precisely where such a foundation is currently inferior to the collection
of separate models it would replace.

Joint training quantifies both the feasibility and the cost of this consolidation. The reported checkpoint maintains median relative $L^2$ errors below $0.8\%$ across the three pretrained tasks and reproduces their defining structures, and two replicate seeds reproduce its PlasticDeform error while giving larger LDC and CosmicDose errors; including cavity recirculation, global latitudinal variation in radiation dose and localized stress concentration. Mean-field controls, scale-normalized errors, spectral comparisons and task-specific diagnostics show that this performance is not obtained by reproducing only average fields. Nevertheless, several single-task operators remain more accurate on LDC, by a factor of about five for the best of them, On PlasticDeform the shared model is the most accurate model evaluated in all three of its training seeds, whereas on CosmicDose that ranking holds for the reported checkpoint but not for the replicates. Every checkpoint is selected on validation data disjoint from the test set. The supported claim is consequently broader native coverage with bounded, task-dependent degradation on the structured cavity problem, not universal superiority over specialists. This distinction also clarifies the comparison with scientific foundation models such as Multiple Physics Pretraining, DPOT and Poseidon~\citep{mccabe2023multiple,hao2024dpot,herde2024poseidon}. Those models achieve substantially greater corpus scale, temporal coverage and downstream adaptation by expressing their systems through comparatively standardized field-based interfaces. GEODE explores a complementary axis: fewer tasks, but greater heterogeneity in the scientific meaning of the inputs and in the geometry and discretization of the outputs. Progress in scientific foundation modeling should therefore be assessed along both axes, since increasing the number of trajectories on a common tensor representation does not by itself establish compatibility with the distinct observations and component models encountered in an energy or radiation workflow.

The routing experiments provide evidence that the shared library is organized rather than merely overparameterized. The pretrained tasks select different dominant experts in the first layer and converge on the same dominant experts in the two deeper layers, suggesting that task-dependent processing is concentrated near the heterogeneous interfaces before computation becomes more strongly shared in the aligned latent space. This interpretation is supported by intervention rather than gate visualization alone: forcing a task through another task's dominant route increases its error by factors of approximately 13 to 117, the same intervention on a randomized library changes nothing, and the live soft routes outperform their one-hot approximations wherever the first-layer gate is diffuse. At the same time, the expert assignments are not a unique physical taxonomy. Re-presenting LDC through a new interface produces a different route with similar accuracy, and matched ablations show that the value of explicit task labels and expert mixing varies by task. Expert indices and wavelet orders should therefore be understood as learned computational allocations shaped by joint optimization and competition among tasks, rather than as fixed labels for particular governing equations. The more general conclusion is that heterogeneous tasks can develop functionally load-bearing early pathways while retaining shared deeper computation.

The sequential experiments identify the task-specific input-output interface, rather than routing alone, as the operative unit of expansion. A new gate cannot compensate for a conditioning vector with different physical meaning, a different number of output variables and a new query geometry. Increasing router capacity over several orders of magnitude accordingly fails to acquire HeatExchanger, whereas adapting the encoder, output projection, skip paths, gate and geometry decoder while freezing the expert library produces accurate predictions. This private interface contains $3.9$ million parameters, or $2.1\%$ of the pretrained model. Parameter isolation leaves earlier outputs unchanged to numerical precision when HeatExchanger and then Subchannel are introduced, while unrestricted fine-tuning causes substantial interference and rehearsal requires retention of the earlier datasets and changes their predictions, by an amount within the seed spread of retraining. This form of stability is useful whenever an existing surrogate has already been verified for a specified operating envelope, because adapting the system to a new task need not alter the function that was previously evaluated. The result is a property of parameter isolation rather than evidence that the shared model has learned resistance to forgetting. The result should not, however, be described as learned immunity to catastrophic forgetting. Retention follows from freezing the shared library and isolating new parameters, and every new task incurs an interface cost that does not amortize. GEODE is therefore best viewed as a shared computational library with a controlled extension mechanism, rather than as a monolithic parameter set into which arbitrary new physics can be absorbed through routing alone.

The transfer controls further show that preservation and reuse are separate properties. For HeatExchanger, the pretrained library provides a clear benefit in the lowest-data regime, is indistinguishable from a norm-matched randomized library at $10\%$ of the data, and a specialist becomes more accurate when sufficient target data are available. Under the complete-data audit, the trained experts outperform randomized experts and a dense backbone by about $10\%$. Subchannel gives the opposite result: over the trained experts the interface learns the temperature level in all three seeds, whereas over norm-matched random experts it stays at the constant-level solution, which the training-mean field itself reaches at about five percent, in two of three seeds and escapes it in one. Much of this difference, however, concerns field level and scale rather than a comparable improvement in the zero-mean spatial profile. This decomposition is important for energy and radiation fields, where large offsets in temperature, pressure or dose can make relative full-field error insensitive to lower-amplitude spatial variations that determine gradients, hot spots and safety margins. The improvement obtained with a profile-aware Subchannel loss illustrates why downstream transfer should be separated into calibration and structural components. The evidence supports conditional cross-physics reuse: pretrained computation can be valuable when data are scarce or when a target benefits from scale information already represented by the library, but it does not constitute a universal representation of physical dynamics. A natural next step is to train across a larger collection of tasks using objectives that reward subsequent reuse explicitly rather than allowing experts to optimize only for the systems that created them.

Geometry-adaptive decoding provides a complementary form of reuse, but its demonstrated scope is narrower than geometry generalization in the unrestricted sense. The same coordinate-based decoder serves Cartesian, spherical and irregular finite-element outputs with node counts ranging from $1{,}733$ to $65{,}341$, and the pretrained predictions remain stable when half of the original query nodes are removed and the neighborhood graph is rebuilt. These results establish flexibility with respect to output discretization and query density. They do not establish generalization to unseen shapes, because the reduced query points belong to the original domains and the decoder neighborhood radius remains task calibrated. Geometry is fixed within each pretrained task, and on a separate benchmark in which the boundary changes between samples, a from-scratch specialist based on the present architecture reaches approximately $18\%$ relative $L^2$ error, with no material improvement after doubling the discretization resolution. The limitation originates primarily in the input path: a parameter vector is lifted onto a fixed latent grid whose spatial semantics are assumed to remain constant. Supplying geometry as an input field, point cloud or mesh representation is a direct route toward geometry-varying systems and remains compatible with coordinate-based decoding. Relatedly, the present inputs are finite-dimensional vectors, even when they encode sensor measurements or loading histories. The claims should therefore be understood in the parametric-operator setting rather than the fully resolution-invariant function-to-function setting of classical operator learning. Extending the encoder to accept sampled functions and varying geometries would connect these two regimes.

The system-level trade-off also places limits on where the architecture is advantageous. At five tasks, GEODE and its two additional interfaces store $198.5$ million parameters, compared with $953.0$ million across five matched specialists, but require $40.4$ cumulative GPU-hours rather than $31.9$ under the reported training protocols. The architecture reduces model replication and marginal storage at the price of additional adaptation compute and a constant private memory cost for every new interface. This comparison is against matched GEODE specialists. The measured DeepONet specialists total only 1.970 million parameters across five tasks, and one shared DeepONet uses 0.422 million across the three pretraining tasks; GEODE does not offer lower storage than these smaller alternatives. Total deployed storage counts the shared backbone once plus each private interface, while trainable acquisition parameters count only the new interface. Parameter payload at a specified precision, serialized checkpoints, optimizer state and runtime activation memory are different quantities. Five tasks are insufficient to determine the sequence length at which this balance ceases to be favorable, to measure saturation of the expert library or to make meaningful use of the nominal $512$ dominant routing paths. Scaling studies should therefore vary the number and diversity of tasks while holding expert capacity fixed, and should report storage, training compute, inference cost and accuracy jointly. Dense gating is another constraint because it evaluates all $E$ experts and scales as $\mathcal{O}(E)$; sparse top-$k$ routing could reduce this cost to $\mathcal{O}(k)$ as the library grows. The current model also requires an explicit task label. Inferring the task context from observations, geometry or accompanying metadata would be necessary for settings in which the operating mode is unknown or changes during deployment.

Several measurement and validation boundaries prevent the present results from being interpreted as evidence of deployment readiness in high-consequence systems. Relative $L^2$ error divides by the norm of the complete target field and can appear small when a large spatial mean dominates the variation; across the evaluated tasks, the resulting scale effect ranges from approximately $2$ to $121$. Reporting errors relative to both the complete field and its variation is therefore necessary, particularly for temperature and radiation fields. The available datasets also do not provide per-sample estimates of numerical or experimental uncertainty, so the study cannot establish whether surrogate error lies below the uncertainty of the reference solution. All evaluated outputs are two-dimensional and correspond to individual states rather than temporal rollouts, and training does not impose conservation laws, governing-equation residuals or constitutive constraints. The reported physical diagnostics are consequently empirical consistency checks rather than guarantees. Although the supplementary perturbation experiments characterize local sensitivity to normalized additive noise and show stable routing over the tested range, they do not cover sensor failure, rare transients, distribution shift or adversarial observations. Deployment in reactor, radiation-monitoring or other high-consequence settings will require three-dimensional and time-dependent evaluation, calibrated predictive uncertainty, explicit validity monitoring, broader robustness testing and integration of physical constraints where appropriate.

Taken together, the results support a narrower but more testable conclusion. Scientific foundation-model evaluation across heterogeneous interfaces should separate at least three questions that aggregate downstream accuracy does not resolve: whether earlier functions are preserved, whether computation learned before a new task is actually reused, and whether the reported field metric reflects field level, spatial structure or both. GEODE provides one implementation through which these questions can be measured without first converting every task into the same scientific representation. The findings that give this content are four. The scientific
interface, not the router, is the unit of extension, and increasing router capacity
over several orders of magnitude does not substitute for it. Preservation of deployed
behavior follows from parameter isolation and is structural rather than learned,
which is why we report it as a property of the computation and not as resistance to
catastrophic forgetting. Reuse of the pretrained core is neither structural nor
guaranteed, and must be measured against a randomized control of identical
architecture; measured that way it is real when a new component's data are scarce and
absent when they are not. Full-field relative error can be dominated by field level
rather than spatial structure, by a factor reaching $121$ on these tasks, so a large
apparent difference between models can correspond to a small difference in the
physics they represent.

The alternative this architecture must justify itself against is not another
foundation model but a collection of small independent operators, and the comparison
should be made in the terms a practitioner would use. DeepONet specialists for all five tasks total $1.970$ million parameters against
$198.5$ million here and preserve earlier predictions by construction because they
share nothing, and task-specific operators are more accurate than GEODE on three of
the five problems. What they cannot offer is reuse, and the results here give a
quantitative account of what that absence costs: a factor of $1.3$ over a
norm-matched random library when a new component's data are scarce, approximately
$10\%$ once they are not, in one case the difference between learning the absolute
field level and remaining at a constant-level solution, and nothing at all at
intermediate data volumes. A practitioner should therefore ask how many systems will
eventually be served, how often a new one will arrive with little data, and whether a
silent change to an already-verified surrogate is acceptable. For a small, fixed and
well-instrumented set of systems, independent specialists remain the appropriate
choice, and nothing here argues otherwise.

The boundaries are equally definite. All evaluated outputs are two-dimensional single
states on fixed within-task geometries; training imposes no conservation or
constitutive constraints and the physical diagnostics are empirical checks rather
than guarantees; no calibrated predictive uncertainty is provided; the model requires
an explicit task label and does not infer its operating context; and five tasks are
too few to determine where the storage and compute balance ceases to favor sharing or
to exercise the routing capacity the library nominally possesses. Corpus-axis
capabilities, including zero-shot prediction for unseen governing equations and
scaling behavior across large PDE collections, are outside what this study
demonstrates. Establishing an energy and radiation foundation model as infrastructure
for component screening, virtual sensing, field reconstruction and decision support
will require three-dimensional and time-dependent evaluation, calibrated uncertainty,
explicit validity monitoring, geometry supplied as an input rather than assumed fixed,
and a substantially larger and more diverse collection of tasks. The contribution is therefore not a claim that heterogeneous scientific modeling has been solved by one universal architecture. It is an experimentally controlled account of what sharing across heterogeneous scientific interfaces does and does not mean: interfaces contribute to measured performance, preservation can be guaranteed without reuse, pretrained reuse is conditional, and commonly reported field errors can overstate differences in spatial structure. These distinctions provide falsifiable tests for future scientific foundation models whose apparent breadth depends on how different physical problems are represented.

%% file: sections/Methods.tex
\section{Methods}
\label{sec:methods}

\subsection{Task formulation}
\label{sec:task_formulation}

We consider a collection of scientific prediction tasks indexed by $\ell\in\{1,\ldots,M\}$. Each task supplies a finite-dimensional conditioning vector $\mathbf{a}^{(\ell)}\in\mathbb{R}^{d_\ell}$ and asks for one or more physical quantities at a set of spatial query points $Q^{(\ell)}=\{\mathbf{q}^{(\ell)}_i\}_{i=1}^{N_\ell}$. The target is a field $\mathbf{u}^{(\ell)}:\Omega_\ell\rightarrow\mathbb{R}^{c_\ell}$, where $\Omega_\ell$ is a two-dimensional spatial domain or coordinate chart, $N_\ell$ is the number of evaluated nodes and $c_\ell$ is the number of output variables. This formulation includes parameter-to-field, sparse-observation-to-field and loading-history-to-field mappings. It therefore does not require every task to expose the coefficients of a common partial differential equation or to share the coordinate-wise meaning of its inputs. The model approximates
\begin{equation}
  \widehat{\mathbf{u}}^{(\ell)}(\mathbf{q})
  = \mathcal{G}\!\left(\mathbf{a}^{(\ell)},\ell,\mathbf{q}\right),
  \qquad \mathbf{q}\in\Omega_\ell,
  \label{eq:heterogeneous_operator}
\end{equation}
while allowing $d_\ell$, $c_\ell$, $N_\ell$, the physical variables and the geometry to differ among tasks. The task identifier is supplied explicitly. The present formulation is a finite-dimensional conditioning-to-field, or parametric/observational operator, setting: the conditioning input is vectorized, whereas the output is coordinate-conditioned and field-valued. It should not be interpreted as the fully resolution-invariant function-to-function setting in which arbitrary sampled input functions or unseen geometries can be supplied without an interface. GEODE is consequently a supervised multi-task surrogate rather than a model that infers an unseen governing equation from its symbolic form.

The five tasks used in this study are summarized in \Cref{tab:methods_tasks}. Lid-driven cavity flow (LDC), CosmicDose and PlasticDeform are used for joint pretraining. HeatExchanger and Subchannel are excluded from pretraining and introduced sequentially to test acquisition of new energy-system mappings. All reported outputs are single states on fixed within-task geometries. The query discretization can change at evaluation time, but variation of the physical boundary from one sample to another is outside the present formulation.

\begin{table}[htbp]
  \centering
  \caption{Tasks used to train and evaluate GEODE. The input dimension refers to the vector presented to the task interface before any padding. Output nodes refer to the native discretization used for the principal evaluation.}
  \label{tab:methods_tasks}
  \small
  \setlength{\tabcolsep}{4pt}
  \renewcommand{\arraystretch}{1.15}
  \begin{tabular}{@{}lcccll@{}}
    \toprule
    Task & $d_\ell$ & $N_\ell$ & $c_\ell$ & Output domain & Mapping \\
    \midrule
    LDC & 90 & $4{,}225$ & 3 & Cartesian grid & boundary history $\rightarrow$ field \\
    CosmicDose & 84 & $65{,}341$ & 1 & latitude--longitude grid & sparse observations $\rightarrow$ field \\
    PlasticDeform & 101 & $3{,}060$ & 1 & finite-element mesh & loading history $\rightarrow$ field \\
    \addlinespace
    HeatExchanger & 102 & $3{,}977$ & 3 & unstructured flow slice & parameters $\rightarrow$ field \\
    Subchannel & 102 & $1{,}733$ & 3 & rod-bundle cross-section & parameters $\rightarrow$ field \\
    \bottomrule
  \end{tabular}
\end{table}

\subsection{GEODE architecture}
\label{sec:methods_architecture}

GEODE separates task-facing representation from reusable computation. In compact form, the prediction for task $\ell$ is
\begin{equation}
  \widehat{\mathbf{u}}^{(\ell)}(\mathbf{q})
  = D_{\psi_\ell}\!\left(
      B_{\Theta}\!\left(E_{\phi_\ell}(\mathbf{a}^{(\ell)}),
      R_{\rho_\ell}\right),\mathbf{q}
    \right),
  \label{eq:geode_factorization}
\end{equation}
where $E_{\phi_\ell}$ maps the native conditioning data to a common latent grid, $B_\Theta$ is the expert wavelet library, $R_{\rho_\ell}$ produces task- and sample-conditioned routing weights, and $D_{\psi_\ell}$ maps latent features to the native query set and output variables. In the jointly pretrained model, the task-facing operations are implemented by common modules conditioned on the task identifier and jointly optimized with the library. For sequential acquisition, a new, disjoint copy of the complete task interface is introduced while $B_\Theta$ and every earlier interface are frozen. This distinction is important: the experiments do not claim that routing alone constitutes a sufficient interface for an unseen modality.

\subsubsection{Task interfaces and input encoding}
\label{sec:methods_encoder}

For joint pretraining, each conditioning vector is standardized using statistics computed from that task's training split, flattened only after standardization and zero-padded to $d_{\max}=101$. Padding is performed after normalization so that artificial zeros do not enter the estimated means or standard deviations. A linear projection $P_0:\mathbb{R}^{101}\rightarrow\mathbb{R}^{2{,}304}$ maps the padded vector to an abstract $48\times48$ latent array. After reshaping, a pointwise convolution $P_1:\mathbb{R}\rightarrow\mathbb{R}^{W}$ lifts the array to width $W=48$, giving $\mathbf{v}_0^{(\ell)}\in\mathbb{R}^{48\times48\times48}$. The latent grid $\Gamma=[0,1]^2$ is an indexing space and is not interpreted as the physical domain. A newly acquired task receives its own input projection with its native width; the $102$-dimensional HeatExchanger and Subchannel vectors are therefore not truncated or inserted into the $101$-dimensional pretrained projection.

The LDC input contains a prescribed lid-velocity history represented at six radial-basis-function interpolation points over 15 snapshots, giving $d_{\mathrm{LDC}}=90$. CosmicDose contains seven daily measurements from each of 12 neutron-monitoring stations, giving $d_{\mathrm{CosmicDose}}=84$. PlasticDeform contains a 101-dimensional representation of the sign-reversing displacement history and associated parameters. Subchannel contains inlet temperature, inlet velocity and a 100-point axial heat-flux profile. HeatExchanger contains inlet velocity, inlet temperature and a 100-point discretization of the imposed heat-flux profile. Because these coordinates have different physical meanings, no coordinate-wise correspondence among tasks is assumed.

\subsubsection{Shared wavelet expert library}
\label{sec:methods_experts}

The shared core comprises $L=3$ expert wavelet integral blocks (EWIBs). Each block contains $E=8$ wavelet integral experts instantiated with the Daubechies db1 through db8 bases. For expert $e$, let $\mathcal{W}_{\psi_e}$ and $\mathcal{W}^{-1}_{\psi_e}$ denote the forward and inverse two-dimensional discrete wavelet transforms. The expert operation is
\begin{equation}
  \mathcal{K}_{j,e}(\mathbf{v})
  = \mathcal{W}^{-1}_{\psi_e}
    \mathcal{R}_{j,e}
    \mathcal{W}_{\psi_e}(\mathbf{v}),
  \label{eq:wavelet_expert_compact}
\end{equation}
where $\mathcal{R}_{j,e}$ applies learned spectral convolution to the approximation, horizontal, vertical and diagonal wavelet subbands. We use one wavelet-decomposition level. Within each subband, Fourier coefficients are multiplied by learned complex-valued weights and transformed back before wavelet reconstruction; the two learned Fourier blocks use the dimensions specified below. Complex coefficients are counted as two real parameters in all parameter totals. The eight bases provide alternative localization and regularity biases, but the expert index is not assigned a fixed physical interpretation. The transform uses symmetric boundary extension, implemented with \texttt{pytorch\_wavelets} 1.3.0.

Given per-channel routing weights $\lambda_{j,e}^{(\ell)}$, block $j$ computes
\begin{equation}
  \mathbf{v}_{j+1}^{(\ell)}
  = \operatorname{Mish}\!\left[
      W_j^{(0)}\mathbf{v}_{j}^{(\ell)}
      + \sum_{e=1}^{E}
      \lambda_{j,e}^{(\ell)}\odot
      \mathcal{K}_{j,e}\!\left(\mathbf{v}_{j}^{(\ell)}\right)
    \right],
  \label{eq:ewib}
\end{equation}
where $W_j^{(0)}$ is a $1\times1$ convolutional skip path and $\odot$ broadcasts each expert weight over the spatial dimensions of its assigned feature channel. Two pointwise projections, $48\rightarrow128$ and $128\rightarrow c_{\max}$ with $c_{\max}=3$, transform the final latent field into $2{,}304$ source features for decoding.

For a $48\times48$ input, the single-level approximation and three detail subbands of db$N$ have side length $m_N=\lfloor(48+2N-1)/2\rfloor$, giving $24,25,\ldots,31$ for db1--db8. Each subband is transformed with a two-dimensional real fast Fourier transform (FFT). Two complex weight tensors per subband have shape $W\times W\times k_N\times k_N$, where $k_N=\lfloor m_N/2\rfloor+1$, giving $13,13,14,14,15,15,16,16$. The output spectrum is initialized to zero; learned products are assigned first to the first $k_N$ rows and then to the last $k_N$ rows, in both cases using the first $k_N$ columns. At these dimensions the row blocks overlap, the second assignment overwrites the overlap, and their union covers the stored real-FFT spectrum. There is no unchanged-coefficient pass-through. These dimensions describe the expert kernels; the gate has its own wavelet encoder.

\subsubsection{Task-conditioned routing}
\label{sec:methods_routing}

Each EWIB has a gate that depends on the current latent field and the explicit task identifier. The task identifier is one-hot encoded and linearly embedded in 16 dimensions. A wavelet encoder with spatial downsampling compresses the current latent field, after which the context vector and task embedding are concatenated and passed through a multilayer perceptron with hidden dimensions $512$, $256$, $128$, $64$ and $32$, and Mish activations. The output logits are reshaped to $W\times E$ and normalized over the expert dimension,
\begin{equation}
  \lambda_{j,w,e}^{(\ell)}
  = \frac{\exp g_{j,w,e}(\mathbf{v}_j^{(\ell)},\ell)}
  {\sum_{e'=1}^{E}\exp g_{j,w,e'}(\mathbf{v}_j^{(\ell)},\ell)}.
  \label{eq:gate_softmax}
\end{equation}
The gate is therefore both sample-conditioned and task-conditioned, and routing is dense: every expert is evaluated before its output is weighted. With $E=8$ and $L=3$, a single feature channel has $E^L=512$ possible dominant expert sequences. This number describes routing choices per channel and should not be read as 512 independent whole-model pathways.

Joint pretraining augments each task minibatch update with routing regularization and a learned task-conditioned logit bias. Let $\bar{\boldsymbol\lambda}_j^{(\ell)}$ be the current task's gate probabilities averaged over minibatch samples and channels, and let $\mathbf{q}_j^{(t)}$ be the detached exponential moving average (EMA) for task $t$. If $A_\ell$ denotes the other tasks whose EMAs have been initialized, the implemented objective for that update is
\begin{align}
 \mathcal L_{\mathrm{joint}}^{(\ell)}
 &=\mathcal L_{\mathrm{data}}^{(\ell)}
   +0.05\,\mathcal L_{\mathrm{sep}}^{(\ell)}
   +0.02\,\mathcal L_{\mathrm{ent}}^{(\ell)}
   +10^{-4}\mathcal L_z^{(\ell)},\label{eq:joint_objective}\\
 \mathcal L_{\mathrm{sep}}^{(\ell)}
 &=-\frac{1}{|A_\ell|}\sum_{t\in A_\ell}
    \sum_{j=1}^{L}\frac{w_j}{\sum_k w_k}
    \operatorname{JSD}(\bar{\boldsymbol\lambda}_j^{(\ell)},\mathbf q_j^{(t)}),\\
 \mathcal L_{\mathrm{ent}}^{(\ell)}
 &=\frac1L\sum_{j=1}^{L}w_j\,
       \mathbb E_{n,c}[H(\boldsymbol\lambda_{j,n,c}^{(\ell)})].
\end{align}
Here $\operatorname{JSD}$ denotes the Jensen--Shannon divergence and $w=(1,4,4)$, so the effective layer separation coefficients are $(0.00556,0.02222,0.02222)$ before averaging over reference tasks. The separation term is zero when $A_\ell$ is empty. The EMA update uses decay $0.95$; reference probabilities are detached, clamped at $10^{-8}$ and normalized before pairwise JSD evaluation. Entropy is averaged after computing each sample/channel distribution's entropy, rather than computed from the averaged distribution. The $z$-loss is the mean over layers and sample/channel entries of the squared log-sum-exp of gate logits. Separation and entropy begin at zero-based epoch index 5, after five initial epochs; the $z$-loss is active throughout. Ablations remove the task identifier or separation term to examine their effects on routing and prediction.

\subsubsection{Geometry-adaptive decoder}
\label{sec:methods_decoder}

The decoder maps the $48\times48$ latent source array to a task's native output coordinates without rasterizing the target field onto the latent grid. For task $\ell$, the latent sites are assigned source coordinates $\{\mathbf{z}^{(\ell)}_j\}_{j=1}^{2{,}304}$ spanning the physical bounding box used by that task. For query $\mathbf{q}^{(\ell)}_i$ and neighboring source $\mathbf{z}^{(\ell)}_j$, the decoder constructs
\begin{equation}
  \boldsymbol{\phi}^{(\ell)}_{ij}
  = \left[
      \mathbf{f}^{(\ell)}_j,
      \mathbf{z}^{(\ell)}_j,
      \mathbf{q}^{(\ell)}_i,
      \sin(\pi q_{i,1}),\cos(\pi q_{i,1}),
      \sin(\pi q_{i,2}),\cos(\pi q_{i,2}),s_\ell(\mathbf q_i)
    \right]\in\mathbb{R}^{12},
  \label{eq:decoder_features_methods}
\end{equation}
where $\mathbf{f}^{(\ell)}_j\in\mathbb{R}^{3}$ is the projected latent feature and $s_\ell$ is the signed distance function (SDF) feature. A kernel multilayer perceptron with hidden widths $64$, $128$ and $64$ maps this vector to $\mathbf{k}^{(\ell)}_{ij}\in\mathbb{R}^{c_\ell}$. The SDF-enabled paper configuration includes $s_\ell(\mathbf q_i)$ as the twelfth feature. In the current geometry builder it is the minimum distance to the four sides of the task's bounding rectangle, positive inside; it does not encode internal holes or internal boundaries such as the HeatExchanger plate. The clean width-48 checkpoint reported in this paper was trained with this feature active; results still attributed to the historical checkpoint are identified as such where they appear.

For neighborhood $\mathcal{N}_{r_\ell}(\mathbf{q}_i)$, predictions are computed by normalized inverse-distance aggregation,
\begin{equation}
  \widehat{\mathbf{u}}^{(\ell)}(\mathbf{q}_i)
  = \frac{\displaystyle\sum_{j\in\mathcal{N}_{r_\ell}(\mathbf{q}_i)}
      w_{ij}\mathbf{k}^{(\ell)}_{ij}}
    {\displaystyle\sum_{j\in\mathcal{N}_{r_\ell}(\mathbf{q}_i)}w_{ij}+\varepsilon},
  \qquad
  w_{ij}=\frac{1}{\lVert\mathbf{q}_i-\mathbf{z}_j\rVert_2+\varepsilon}.
  \label{eq:idw_methods}
\end{equation}
The sparse reduction uses \texttt{segment\_csr} from \texttt{torch-scatter} and has complexity $\mathcal O(|\mathcal E_\ell|)$ in the number of source--query edges. Both the weight and normalization use $\varepsilon=10^{-6}$. The initial radius is the mean distance to the twentieth-nearest non-self latent source, estimated over up to 500 randomly sampled source points. If any query has fewer than three neighbors, the search is repeated once with radius multiplied by $1.5$. An empty neighborhood yields zero output: kernel biases occur within the per-edge MLP and are not added after aggregation. Changing output queries rebuilds the neighborhood graph without retraining the weights; the initial radius depends on the latent sources, not the query density. CosmicDose uses Euclidean distances in the latitude--longitude chart, in degrees, without longitude wrapping or polar correction; these are not geodesic distances.

\subsection{Joint pretraining}
\label{sec:methods_pretraining}

The three pretraining tasks share one model. Within each epoch, the trainer traverses the complete minibatch loader of each task in turn, using batches of up to 16 samples and a separate Adam update for every batch. Thus task sample counts affect the number of optimizer updates; this is not an equal-task combined minibatch update. Inputs and outputs are standardized separately for each task using only its training split, and predictions are transformed back to physical units before the reported metrics are computed. For a batch from task $\ell$, the implemented data term is the sum of per-sample relative $L^2$ errors in normalized output space,
\begin{equation}
  \mathcal{L}_{\mathrm{data}}^{(\ell)}
  =\sum_{n\in\mathcal{B}_\ell}
    \frac{\lVert\widehat{\widetilde{\mathbf{u}}}_n^{(\ell)}-\widetilde{\mathbf{u}}_n^{(\ell)}\rVert_2}
         {\lVert\widetilde{\mathbf{u}}_n^{(\ell)}\rVert_2},
  \label{eq:data_loss_methods}
\end{equation}
The model is trained for 120 epochs with Adam, an initial learning rate of $10^{-3}$, weight decay $10^{-5}$ and a multiplicative learning-rate factor of $0.7$ every 20 epochs. The paper checkpoint was trained on one NVIDIA A40 with gradient checkpointing (120 epochs, 26.4 GPU-hours). The acquisition, transfer, audit and path-ablation runs derived from it use A40 GPUs, the two full-parameter controls A100 GPUs, and the full-training ablations and single-task GEODE specialists GH200 GPUs. The normalized fields are denoted by tildes, and the loss uses no additive denominator constant ($\varepsilon_{\mathrm{loss}}=0$). Unless stated otherwise, the same epoch budget, optimizer, batch size and learning-rate schedule are used for the matched single-task models and architectural ablations.

Checkpoint selection must use a validation split that is disjoint from the final test set. The LDC and CosmicDose results use the source-dataset partitions. The historical PlasticDeform checkpoint was selected with access to the nominal test set; all GEODE accuracies reported in the comparison tables therefore come from the clean rerun described here, whose checkpoint (epoch 118 of 120) was selected on validation data only.

The train/validation/test counts are $3{,}949/493/495$ for LDC, $4{,}011/4{,}012/359$ for CosmicDose and $13{,}410/1{,}490/100$ for clean PlasticDeform. For PlasticDeform, the final 10\% of the 14,900-row source training array, in stored order, is reserved for validation and excluded from both fitting and normalization statistics. Checkpoint selection minimizes the equally weighted mean of the three task-wise mean validation relative $L^2$ losses in normalized space. This criterion differs from the physical-space median reported at test time and from the validation mean-squared error (MSE) used for external baselines.

\subsection{Sequential task acquisition and parameter isolation}
\label{sec:methods_acquisition}

HeatExchanger and Subchannel are introduced after the three-task checkpoint has been fixed. At each acquisition step, the expert library and every parameter, preprocessing statistic and geometry state used by an earlier task are frozen. The new task receives a private interface comprising its input encoder, output projections, skip paths, geometry decoder and gate. This interface contains $3.9$ million trainable parameters, or $2.1\%$ of the $190.6$-million-parameter pretrained model. No earlier training examples are replayed in the principal sequential protocol.

For an earlier task $t$, write its predictor as
\begin{equation}
  f_t(\mathbf{a},\mathbf{q})
  =D_{\psi_t}\!\left(B_\Theta(E_{\phi_t}(\mathbf{a}),R_{\rho_t}),\mathbf{q}\right).
\end{equation}
If acquisition changes only a disjoint new-task parameter vector $\boldsymbol{\eta}$, then $\partial f_t/\partial\boldsymbol{\eta}=0$. Earlier functions are consequently unchanged in exact arithmetic. We also evaluate them after each acquisition. This construction guarantees non-interference, but it does not guarantee useful transfer from the frozen library; that question is tested separately against randomized and dense controls.

The matched HeatExchanger data-efficiency study uses $1\%$, $10\%$, $50\%$ and $100\%$ of its training data. Every fraction and comparison arm is trained for 100 epochs with an initial learning rate of $10^{-3}$. The sequential chain uses its originally specified acquisition recipe and is reported separately from this matched audit; results from these two protocols are not interchanged. The chain's Subchannel step exists in two declared variants that differ only in the objective: the profile-aware variant additionally includes a unit-weight relative $L^2$ term on each output channel after subtracting that sample's spatial mean, and the untreated variant uses the standard loss. This term prevents the large absolute temperature level from overwhelming the smaller wall-to-core spatial variation.

HeatExchanger contains $1{,}200/173/173$ and Subchannel $3{,}200/800/1{,}000$ train/validation/test instances. The interface-acquisition, data-fraction and library-audit runs share one optimization protocol; the chain's Subchannel step differs from it only in the objective variant stated above. The private interface (input encoder with a 1024-wide hidden layer, output projection, private skip paths, private decoder and a fresh gate) is trained from the paper checkpoint for 100 epochs with Adam at learning rate $10^{-3}$, decayed by $0.6$ every 60 epochs, batch size 16 and seed 0. The chain's Subchannel step uses 200 epochs and is reported in both objective variants, and the Subchannel library audit uses 200 epochs with the profile term. Gate-only arms train the shared gate network and task embedding under the same schedule; same-architecture specialists start from random initialization under the same schedule, for 100 epochs in the fraction sweep and 200 epochs for the chain comparison. Data fractions take a contiguous window of the training split, the first $\lceil f N\rceil$ instances with a floor of eight, so that every arm at a fraction sees the same instances; the window can be moved by an offset, and the reported fractions use offset zero.

\subsection{Datasets and preprocessing}
\label{sec:methods_datasets}

\paragraph{Lid-driven cavity flow.}
LDC represents boundary-driven incompressible flow in a square cavity, the canonical setting for the confined recirculating flows found in reactor plena, coolant cavities and mixing volumes. Each sample is conditioned on a lid-velocity history and predicts horizontal velocity, vertical velocity and pressure on a $65\times65$ Cartesian grid. The physical coordinate ranges are $x\in[-0.064,0]$ and $y\in[0,0.064]$. We use 3,949 training instances and 495 test instances from the source partition, with one complete, distinct boundary-condition realization as the splitting unit.

\paragraph{Global cosmic-ray dose reconstruction.}
CosmicDose reconstructs a global sea-level reference dose-rate field
(EXPACS/PARMA) from sparse neutron-monitor observations, the problem underlying
radiation protection for aviation crews, high-altitude and space operations and
ground-level monitoring networks. The output is represented on a
$181\times361$ latitude--longitude grid with latitude in $[-90,90]$ degrees and
longitude in $[-180,180]$ degrees. We use $4{,}011$ training and $359$ test instances, each corresponding to a distinct day in the neutron-monitor and solar-modulation record.

The intrinsic structure of this task must be stated explicitly, because it governs
how its reported error should be interpreted. The reference generator evaluates a
forward transport model in which the field at a given location depends on the solar
modulation state, the local vertical cutoff rigidity, the atmospheric depth and the
local water fraction. In the configuration used here the last two are held fixed and
cutoff rigidity is a function of position rather than of date, so within this
configuration the time-varying reference fields constitute an essentially
one-parameter family indexed by the solar modulation state. The neutron-monitoring
stations supplied to the model as input are among those from which that state is
conventionally evaluated, so the modulation state is in principle recoverable from
the inputs. Low relative error on this task is therefore an expected property of any
model that recovers a single scalar and decodes it accurately, and neither the
reported error nor the skill score against a per-node mean-field predictor should be
read as evidence of high-dimensional reconstruction, of inference under
underdetermination, or of physical discovery. We include CosmicDose because of its
interface rather than its difficulty: it supplies scattered station readings on a
spherical coordinate chart with an output discretization an order of magnitude larger
than the other tasks, and it is the task in this study whose native representation a
common field-tensor substrate would have to eliminate. No claim in this paper depends
on CosmicDose being a difficult inverse problem.

Because adjacent days can be correlated, the exact date indices are retained and
released with the code. As a diagnostic of split proximity, the median
test-to-training nearest-neighbor distance divided by the median
training-to-training nearest-neighbor distance in standardized input space is
$3.25$, so test inputs are on average farther from the training set than training
inputs are from one another; the analogous ratios are $1.59$ for LDC and $0.99$ for
PlasticDeform.

\paragraph{Elastoplastic deformation.}
PlasticDeform predicts the von Mises stress field of a dog-bone tensile specimen under a sign-reversing displacement history, the path-dependent response that enters fatigue and integrity assessment of load-bearing components in energy plants. Each sample has a distinct loading history, and the target is evaluated at 3,060 nodes of an irregular finite-element discretization spanning approximately $x\in[-63,47]$ and $y\in[4,34]$. The source provides 14,900 training instances and 100 nominal test instances; the clean protocol uses 13,410 for fitting and 1,490 for validation. This paper reports only the clean validation-selected rerun described in \Cref{sec:methods_pretraining}.

\paragraph{Heat-exchanger flow.}
HeatExchanger is a two-dimensional transverse cross-section of a three-dimensional steady, incompressible, forced-convection turbulent-flow simulation of a heat-exchanger tube with a wavy tape insert (22\,mm diameter, 0.8\,m long), computed with ANSYS Fluent using the pressure-based solver, the $k$--$\varepsilon$ model with enhanced wall treatment, constant fluid properties and no gravity; the section is taken at the amplitude peak of the insert. Its native unstructured computational fluid dynamics (CFD) mesh contains 3,977 nodes, a scalloped external boundary and a zero-thickness internal plate. The source campaign, generated by our group for a companion study~\citep{roy2026adversarial}, comprises 1,546 cases with inlet velocity sampled uniformly on $[4,5]$\,m\,s$^{-1}$, inlet temperature on $[263,323]$\,K and an imposed wall heat-flux profile with peak values up to about 22\,kW\,m$^{-2}$; the split used here is 1,200 training, 173 validation and 173 test cases. The model predicts pressure and the two in-plane velocity components from a 102-dimensional condition vector; the axial velocity channel of the source data is not used. The native CFD General Notation System (CGNS) connectivity is used for visualization; no image rasterization is used for model training or evaluation.

\paragraph{Pressurized-water-reactor subchannel.}
Subchannel represents the transverse center plane of a square subchannel extracted from a pressurized-water-reactor fuel-rod array, 800\,mm long, simulated with ANSYS Fluent using the renormalization-group (RNG) $k$--$\varepsilon$ model with enhanced wall treatment~\citep{hossain2025virtualsensing}. Four quarter-circle fuel rods bound a cross-shaped coolant region discretized at 1{,}733 nodes. The source campaign comprises 5,000 cases with inlet velocity on $[4.05,4.95]$\,m\,s$^{-1}$, inlet temperature on $[536.4,655.6]$\,K and a sinusoidal axial heat-flux profile with peak values on $[540,660]$\,kW\,m$^{-2}$; the split used here is 3,200 training, 800 validation and 1,000 test cases. The 102-dimensional input comprises inlet temperature, inlet velocity and a 100-point axial heat-flux profile. Outputs are axial velocity, coolant temperature and turbulent kinetic energy. The released dataset provides coordinates without connectivity; triangulation is used only for visualization, with triangles removed whose centroids lie within the analytically specified rods. The model itself consumes query coordinates and does not use this reconstructed connectivity.

Across all tasks, normalization statistics are fitted on the training data only. Output channels are normalized independently, predictions are inverse-transformed before physical-space evaluation, and physical coordinates retain the units used by each dataset. No governing-equation residual, conservation penalty, incompressibility constraint or constitutive constraint is included in training.

\subsection{Baselines and ablation protocols}
\label{sec:methods_baselines}

We organize comparisons according to the question they answer. A per-node mean field computed from the training targets provides a no-learning control. Proper orthogonal decomposition followed by ridge regression provides a linear reduced-order baseline. The historical implementation sweeps ranks $\{8,16,32,64,128\}$ and ridge coefficients $\{10^{-3},10^{-1},10^{1},10^{3}\}$, minimizing median relative error on an internal holdout comprising the first 20\% of training rows. However, its POD basis and normalization were fitted before removing that holdout; its selection was therefore not independent for the complete pipeline. The verified clean CPU rerun fits both the basis and preprocessing on the fitting partition only, selects on the supplied validation partitions (or the clean PlasticDeform validation split), and opens test data only after selection. Its complete protocol and results are given in \Cref{sec:cx_clean_pod,tab:cx_clean_pod}; the principal comparison uses these clean results. Single-task GEODE models use the same backbone, preprocessing and optimization budget as the joint model but see only one task, thereby isolating the cost of heterogeneous sharing. The dense ablation replaces the routed wavelet mixture with a dense backbone of 22 million parameters, while the no-label ablation removes the task embedding from the gates. External neural operators and scientific foundation models are evaluated only where their native input and output interfaces apply. In this paper, a baseline ``adapter'' denotes an input--output compatibility module added by us: a learned lift from the task's native conditioning vector into the model's latent representation and a readout at the supplied output coordinates. It is distinct from GEODE's private continual-acquisition interface, which also includes task-specific skip paths, a geometry decoder and a fresh gate. If a task-specific adapter, rasterization or remeshing is required, that change is identified explicitly and the result is labeled as adapted rather than native; an incompatible model is reported as not applicable rather than assigned a proxy score.

Routing interventions replace the live soft gate with a one-hot dominant route. Each pretrained task is evaluated on its own route and on the dominant route associated with every other task, so that the penalty from hardening the gate is separated from the penalty from route substitution. Effective expert count is defined as $\exp[H(\overline{\boldsymbol{\lambda}})]$, ranging from one for a one-hot distribution to eight for a uniform distribution.

The acquisition audit contains five arms. The gate-only arm updates the new task embedding and router while holding every other parameter fixed; router capacity is varied from the task embedding alone through rank-16 adapters to a private full gate containing 2.3 million parameters. The complete-interface arm updates the new input and output interface with frozen experts. Unrestricted fine-tuning updates all parameters using only the new task. Rehearsal updates all parameters while sampling earlier and new datasets within the same training budget. A from-scratch specialist supplies the task-specific reference. The trained-library control is additionally compared with a norm-matched random library constructed by drawing each complex expert-weight tensor from a complex Gaussian using \texttt{torch.randn} and rescaling it to preserve the original tensor's Frobenius norm (rescaling denominator clamped at $10^{-30}$). Architecture, trainable interface, optimization and data are otherwise held fixed. A dense frozen backbone provides a separate control without wavelet experts.

To test query-set flexibility, we uniformly remove $50\%$ of the native output nodes at evaluation, rebuild the source--query neighborhood graph and evaluate the unchanged checkpoint on the retained queries. Decoder sensitivity is tested by multiplying the calibrated neighborhood radius by $0.75$, $1.5$ and $2.0$. Robustness experiments add independent Gaussian noise with standard deviation $\delta\in\{0.01,0.05,0.1\}$ to normalized inputs and measure both prediction error and changes in the mean gate distribution. These tests establish behavior on modified discretizations and perturbed inputs; they do not establish generalization to a new physical boundary.

The specialist DeepONet uses branch and coordinate-trunk MLPs with three 256-unit Gaussian error linear unit (GELU) hidden layers and a linear output, with 128 basis functions per output channel. It is trained for 200 epochs with Adam at $10^{-3}$, batch size 32 and cosine annealing. The adapted GINO uses \texttt{neuraloperator} 2.0.0, a learned vector lift, a $32\times32$ latent grid, width 32, 16 Fourier modes and three FNO layers. Its 30-epoch Adam/cosine schedule starts at $10^{-3}$, with batch size 32 (16 for CosmicDose) and up to 4,000 sampled training queries (6,000 for CosmicDose); validation and test use all native queries. Both select by normalized validation MSE. The additional operator and pretrained-model adapters, seeds, exact sizes and resource accounting are given in \Cref{sec:cx_wave_protocol,sec:cx_methods_repro,tab:cx_allocations}.

\subsection{Evaluation metrics and statistical analysis}
\label{sec:methods_metrics}

The principal metric is the per-sample relative $L^2$ error in physical units,
\begin{equation}
  e_n^{(\ell)}
  =100\times
    \frac{\lVert\widehat{\mathbf{u}}_n^{(\ell)}-\mathbf{u}_n^{(\ell)}\rVert_2}
         {\lVert\mathbf{u}_n^{(\ell)}\rVert_2},
  \label{eq:relative_l2_methods}
\end{equation}
reported as a percentage. For multichannel outputs, the norm is taken over all reported nodes and channels after inverse normalization. We report the median, interquartile range, mean and 95th percentile because the PlasticDeform distribution is heavy-tailed. We also report root-mean-square error divided by the standard deviation of the target
field, which is insensitive to constant offsets. The two metrics are related exactly.
For one output channel of sample $n$ evaluated at $N$ nodes, let $\bar u_n$ be its
spatial mean and $\sigma_n^2=N^{-1}\lVert\mathbf u_n-\bar u_n\mathbf 1\rVert_2^2$ its
spatial variance. Because $\mathbf u_n-\bar u_n\mathbf 1$ is orthogonal to
$\mathbf 1$, $\lVert\mathbf u_n\rVert_2^2
=\lVert\mathbf u_n-\bar u_n\mathbf 1\rVert_2^2+N\bar u_n^2$, and therefore
\begin{equation}
  \frac{\lVert\widehat{\mathbf u}_n-\mathbf u_n\rVert_2}{\lVert\mathbf u_n\rVert_2}
  =\frac{1}{\kappa_n}\,\frac{\mathrm{RMSE}_n}{\sigma_n},
  \qquad
  \kappa_n=\frac{\lVert\mathbf u_n\rVert_2}{\lVert\mathbf u_n-\bar u_n\mathbf 1\rVert_2}
  =\sqrt{1+\bar u_n^{2}/\sigma_n^{2}}\;\geq 1 .
  \label{eq:scale_factor}
\end{equation}
Full-field relative error therefore understates error relative to spatial variation
by the factor $\kappa_n$, which is governed by the ratio of the field's mean to its
spatial standard deviation and is independent of the model. For multichannel outputs
the decomposition holds with spatial means taken channel by channel and norms taken
over all channels. The range of approximately $2$ to $121$ quoted in the main text is
the range, across tasks and channels, of the median of $\kappa_n$ over the test
samples of each task and channel.

To determine whether a small relative error merely reflects a nearly constant target, we compare against the per-node training mean $\overline{\mathbf{u}}^{(\ell)}$ and compute the squared-error skill score
\begin{equation}
  S^{(\ell)}
  =1-
  \frac{\sum_n\lVert\widehat{\mathbf{u}}_n^{(\ell)}-\mathbf{u}_n^{(\ell)}\rVert_2^2}
       {\sum_n\lVert\overline{\mathbf{u}}^{(\ell)}-\mathbf{u}_n^{(\ell)}\rVert_2^2}.
  \label{eq:skill_methods}
\end{equation}
Here $S=0$ corresponds to the mean-field predictor and $S=1$ to exact reconstruction. Headline values come from one trained checkpoint per arm unless a seed count is stated explicitly; percentiles describe variation across test instances and do not quantify optimization uncertainty across training seeds. Any claim about training stability will therefore require independent reruns rather than test-case percentiles. For the joint model, the heat-exchanger interface, and the subchannel interface over the trained and the norm-matched random library, two such reruns from seeds 1 and 2 with the same recipe are reported in Table~\ref{tab:seeds}; the interface reruns are evaluated with the seed-0 query subset of the adaptive-radius decoder so that their errors are comparable.

For Subchannel, profile error is also evaluated after subtracting each channel's spatial mean, allowing absolute field level and spatial variation to be examined separately. Task-specific physical diagnostics, including discrete velocity divergence for LDC, non-negativity of von Mises stress and hemispheric asymmetry for CosmicDose, use the same numerical operation on prediction and reference and are reported as empirical checks rather than enforced constraints.

\subsection{Computational implementation}
\label{sec:methods_implementation}

Models are implemented in PyTorch. Sparse decoder aggregation uses \texttt{torch-scatter}. The jointly pretrained model contains 190.6 million real-valued parameters when each complex spectral coefficient is counted as two real values. Of these, 187.1 million parameters are in the three expert layers, 3.2 million are in the three gates, approximately 235,000 are in the initial projection, approximately 7,000 are in skip connections, approximately 6,500 are in the output projections and approximately 17,500 are in the geometry decoder. Because routing is dense, its inference complexity remains linear in the number of experts; the present work does not evaluate sparse top-$k$ routing.

Inference timing is measured on one NVIDIA A40 GPU using float32, batch size one, 15 warm-up calls and 50 timed calls with explicit device synchronization. We report the median wall-clock latency. Hardware-specific timings are not combined within a comparison, and reference-solver runtimes are treated as indicative data-generation costs because the solver software and hardware differ among datasets.

The recorded Delta A40 environment uses Python 3.11.13, PyTorch 2.8.0 with CUDA 12.8, \texttt{torch-scatter} 2.1.2, \texttt{pytorch\_wavelets} 1.3.0 and NumPy 2.2.6 on Red Hat Enterprise Linux 9.6. The GH200 runs use the corresponding DeltaAI module environment. The canonical trainer sets PyTorch and NumPy seeds to 0. Additional baseline initialization seeds and numerical settings are specified separately in \Cref{sec:cx_methods_repro}. Recorded GPU allocations for the 28 additional baseline arms are listed in \Cref{tab:cx_allocations}; reconstructed historical costs are not scheduler measurements.

%% file: sections/supp_results.tex
\section{Supplementary Results}
\label{sec:supplementary_results}

This Supplementary Information provides the detailed numerical results, diagnostic visualizations and intervention studies referred to in the main text. Unless stated otherwise, errors are per-sample relative $L^2$ errors evaluated in physical space and are reported as percentages. All GEODE values are those of the clean, validation-selected checkpoint and of the runs regenerated from it, except where a table or paragraph is explicitly marked as historical; $^{\mathrm{h}}$ marks such entries in tables. LDC and CosmicDose use separate validation and test arrays. Medians are arithmetic medians; quartiles and 95th percentiles in the updated headline table use linear interpolation. Legacy intervention summaries retain their original recorded precision. Results from distinct protocol audits are labeled explicitly and should not be combined numerically with the chained continual-acquisition experiment.

\subsection{Task representations and model architecture}
\label{sec:supp_tasks_architecture}

\begin{figure}[htbp]
  \centering
  \includegraphics[width=\textwidth]{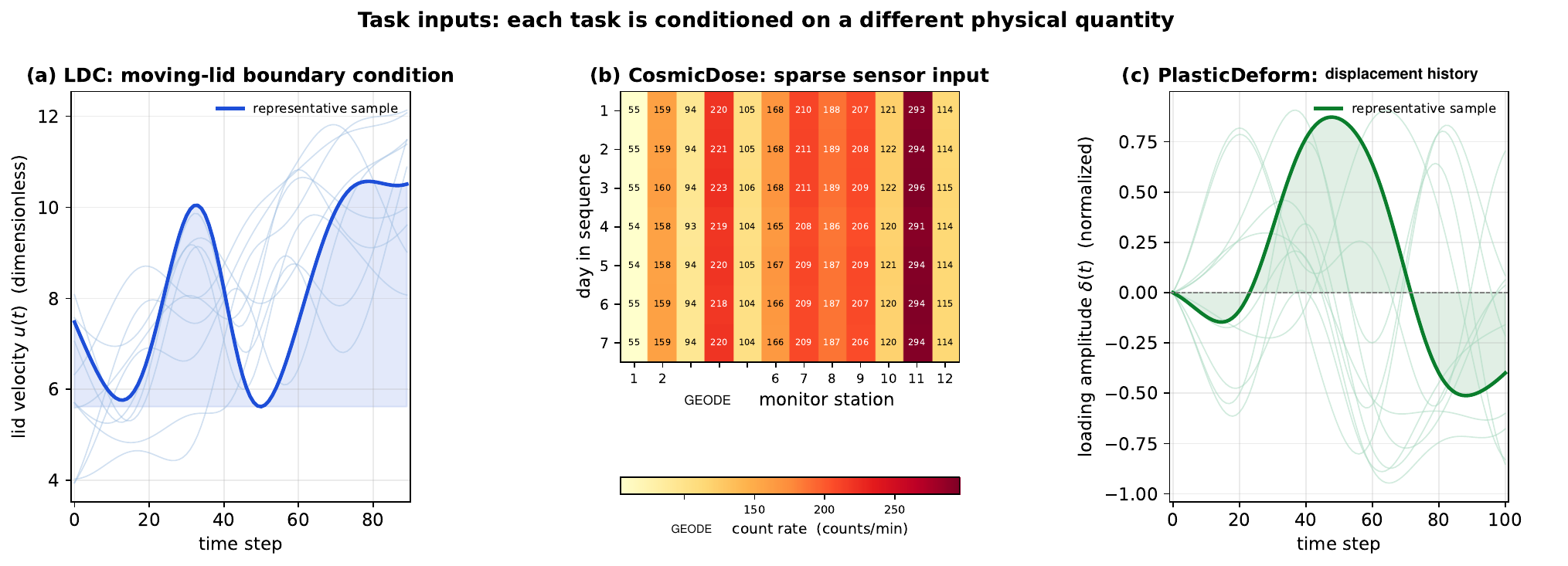}
  \vspace{0.6em}
  \includegraphics[width=\textwidth]{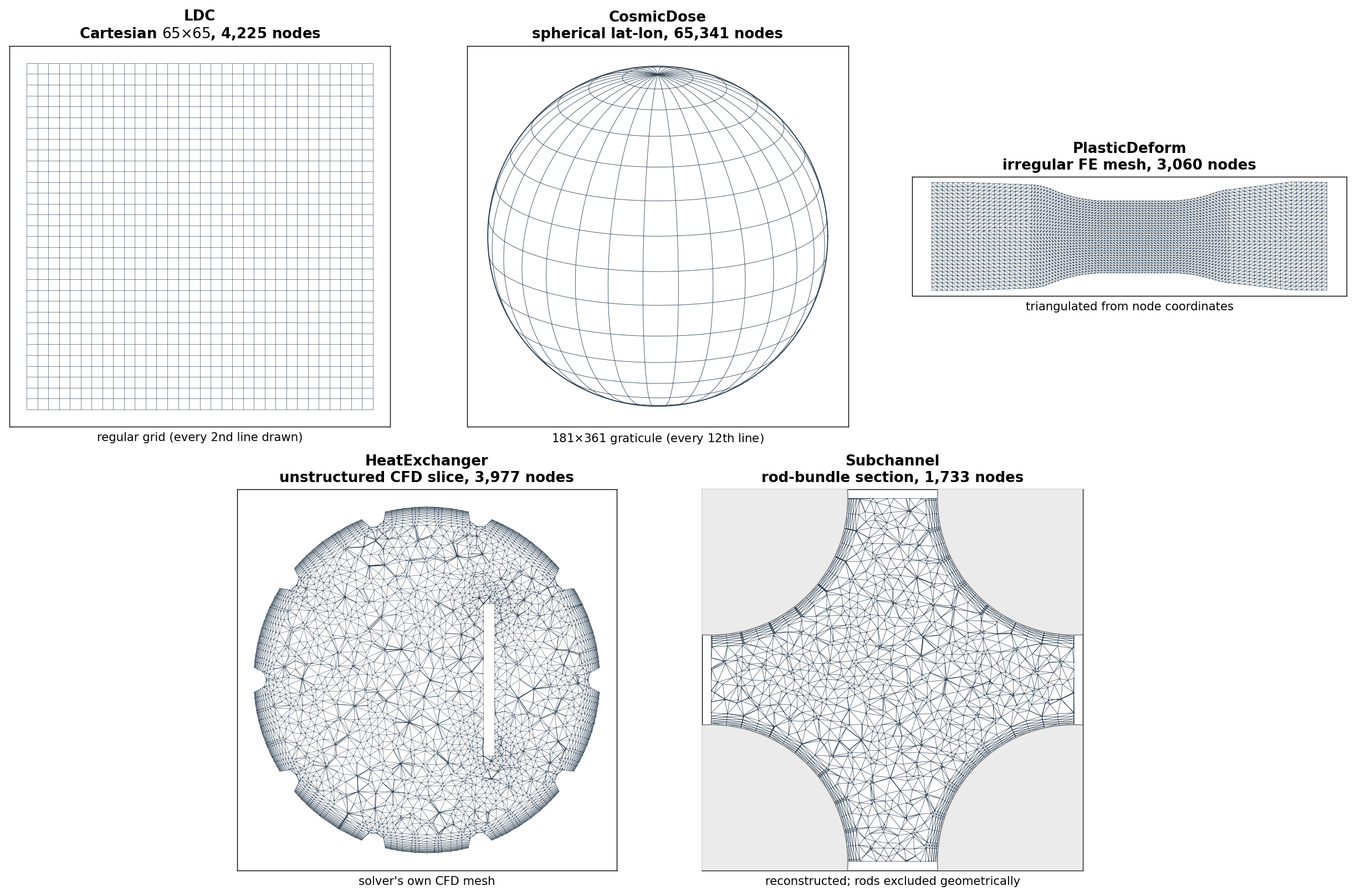}
  \caption{\textbf{Heterogeneous inputs and output domains used to evaluate GEODE.} Top, representative conditioning signals for the three jointly pretrained tasks. LDC is conditioned on a time-dependent lid-velocity profile, CosmicDose receives seven daily measurements from twelve neutron-monitoring stations, and PlasticDeform is conditioned on a sign-reversing displacement history. Bottom, the five output domains: a Cartesian cavity-flow grid, a structured spherical dose grid, an irregular finite-element specimen, an unstructured heat-exchanger domain containing an internal boundary, and a nuclear-reactor subchannel bounded by four fuel rods.}
  \label{fig:supp_task_overview}
\end{figure}

The five tasks differ in input semantics, output variables, coordinate systems and discretizations (\Cref{fig:supp_task_overview}). LDC, CosmicDose and PlasticDeform are jointly pretrained. HeatExchanger and Subchannel are withheld from pretraining and introduced sequentially. Conditioning vectors range from 84 to 102 dimensions, while the output discretizations contain between $1{,}733$ and $65{,}341$ nodes.

\begin{figure}[htbp]
  \centering
  \includegraphics[width=\textwidth]{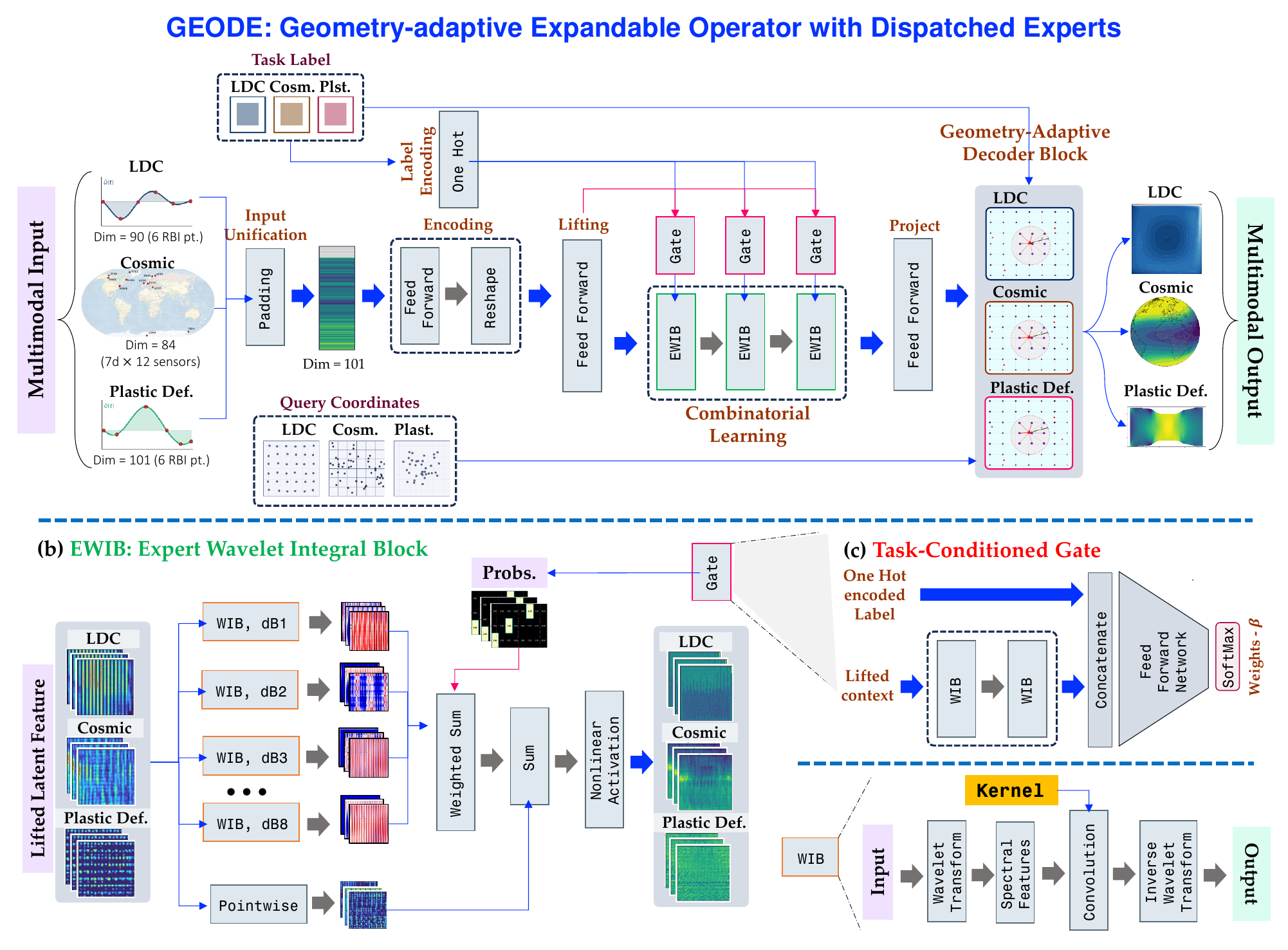}
  \caption{\textbf{Detailed architecture of GEODE.} Task-specific inputs are mapped to a common latent field, processed by a shared task-conditioned wavelet-expert library, and decoded on the native physical output domain. The decoder feature vector is specified in Methods. During acquisition, a new task interface and routing variables are optimized while the shared library and all previous interfaces remain fixed.}
  \label{fig:supp_architecture}
\label{fig:architecture}
\label{fig:latent_pipeline}
\end{figure}

GEODE maps each task-specific input to an abstract $48\times48$ latent grid. Three wavelet-integral layers, each containing eight Daubechies experts, operate in this shared latent space. A task-conditioned gate assigns expert weights independently at each layer and feature channel. Source and query coordinates and harmonic positional features are supplied to the geometry-adaptive decoder. The decoder feature vector includes the signed-distance term specified in Methods. For a new system, the expert library of the $190.6$-million-parameter model and all earlier interfaces remain fixed, while a $3.9$-million-parameter input--output interface and gate are trained (\Cref{fig:supp_architecture}).

\subsection{Per-task accuracy of joint pretraining}
\label{sec:supp_joint_accuracy}

\Cref{tab:supp_joint_results} reports the complete error distribution for the jointly trained model. PlasticDeform has a heavy-tailed relative-error distribution: its mean is $4.80\%$ and its 95th percentile is $34.7\%$, despite a median of $0.399\%$. The large relative errors occur predominantly for low-magnitude target fields. Normalizing root-mean-square error (RMSE) by the standard deviation of the complete target field gives $0.074\,\sigma$ for PlasticDeform.

\begin{table}[htbp]
  \centering
  \caption{\textbf{Test performance of the jointly trained GEODE model.} Results use 495 LDC, 359 CosmicDose and 100 PlasticDeform test instances. The mean-field control predicts the per-node mean of the corresponding training outputs. All values are from one training run; the GEODE values are those of the clean, validation-selected checkpoint.}
  \label{tab:supp_joint_results}
\label{tab:results}
  \small
  \setlength{\tabcolsep}{5pt}
  \renewcommand{\arraystretch}{1.15}
  \begin{tabular}{@{}lccc@{}}
    \toprule
    \textbf{Metric} & \textbf{LDC} & \textbf{CosmicDose} & \textbf{PlasticDeform} \\
    \midrule
    Median relative $L^2$ error (\%) & 0.755 & 0.031 & 0.399 \\
    Interquartile range (\%) & 0.54--1.34 & 0.019--0.053 & 0.25--2.52 \\
    Mean relative $L^2$ error (\%) & 1.076 & 0.039 & 4.798 \\
    95th percentile (\%) & 2.679 & 0.089 & 34.67 \\
    \addlinespace
    Mean-field error, median (\%) & 59.80 & 1.79 & 20.13 \\
    Mean-field skill score & 0.9993 & 0.9995 & 0.9935 \\
    RMSE divided by field $\sigma$ & 0.020 & 0.0054 & 0.074 \\
    \bottomrule
  \end{tabular}
\end{table}

\begin{figure}[htbp]
  \centering
  \includegraphics[width=\textwidth]{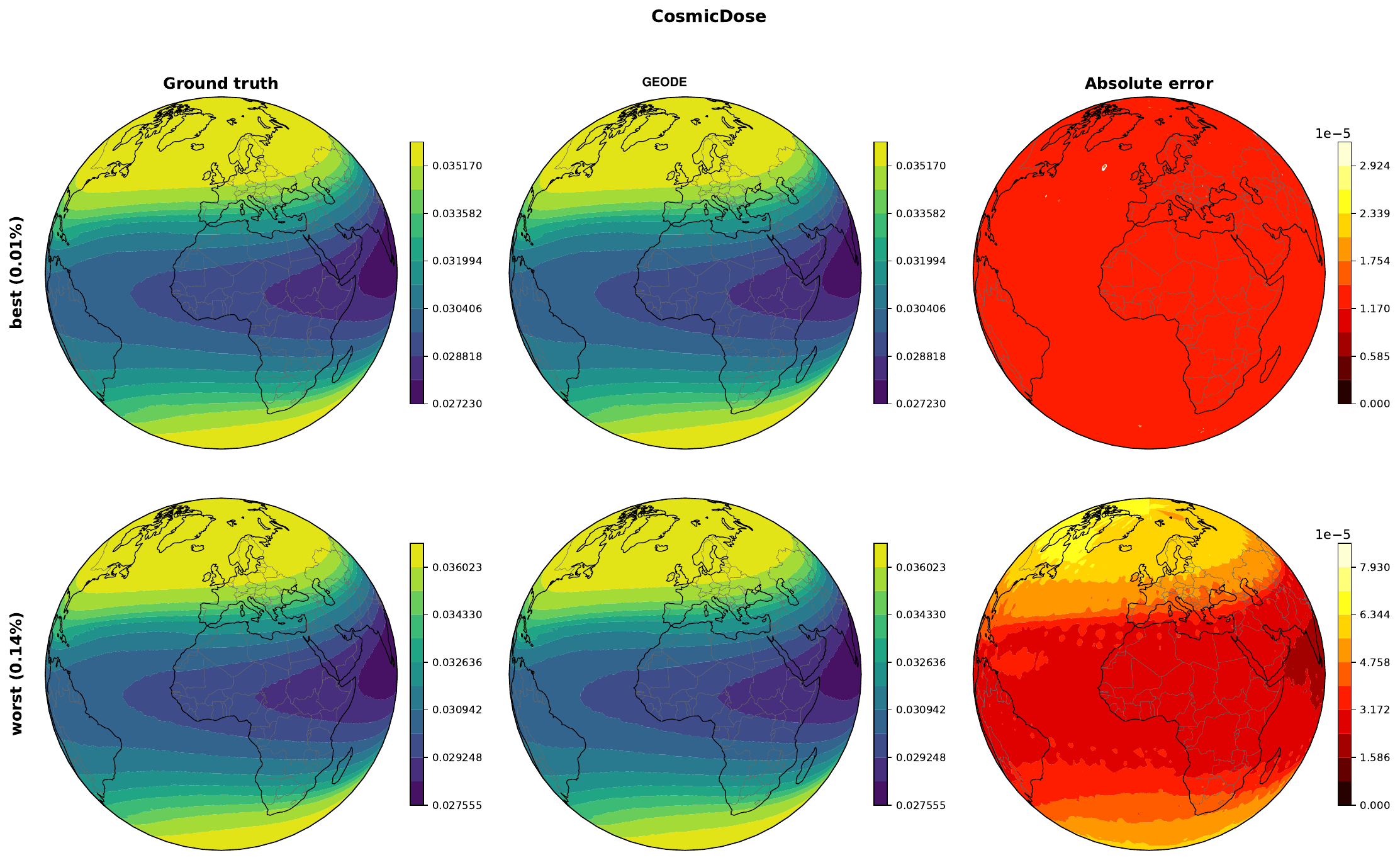}
  \caption{\textbf{CosmicDose predictions for the lowest- and highest-error test instances.} Columns show the reference field, GEODE prediction and absolute pointwise error. The largest residuals occur in regions of steep spatial variation.}
  \label{fig:supp_cosmic_fields}
\end{figure}

\begin{figure}[htbp]
  \centering
  \includegraphics[width=\textwidth]{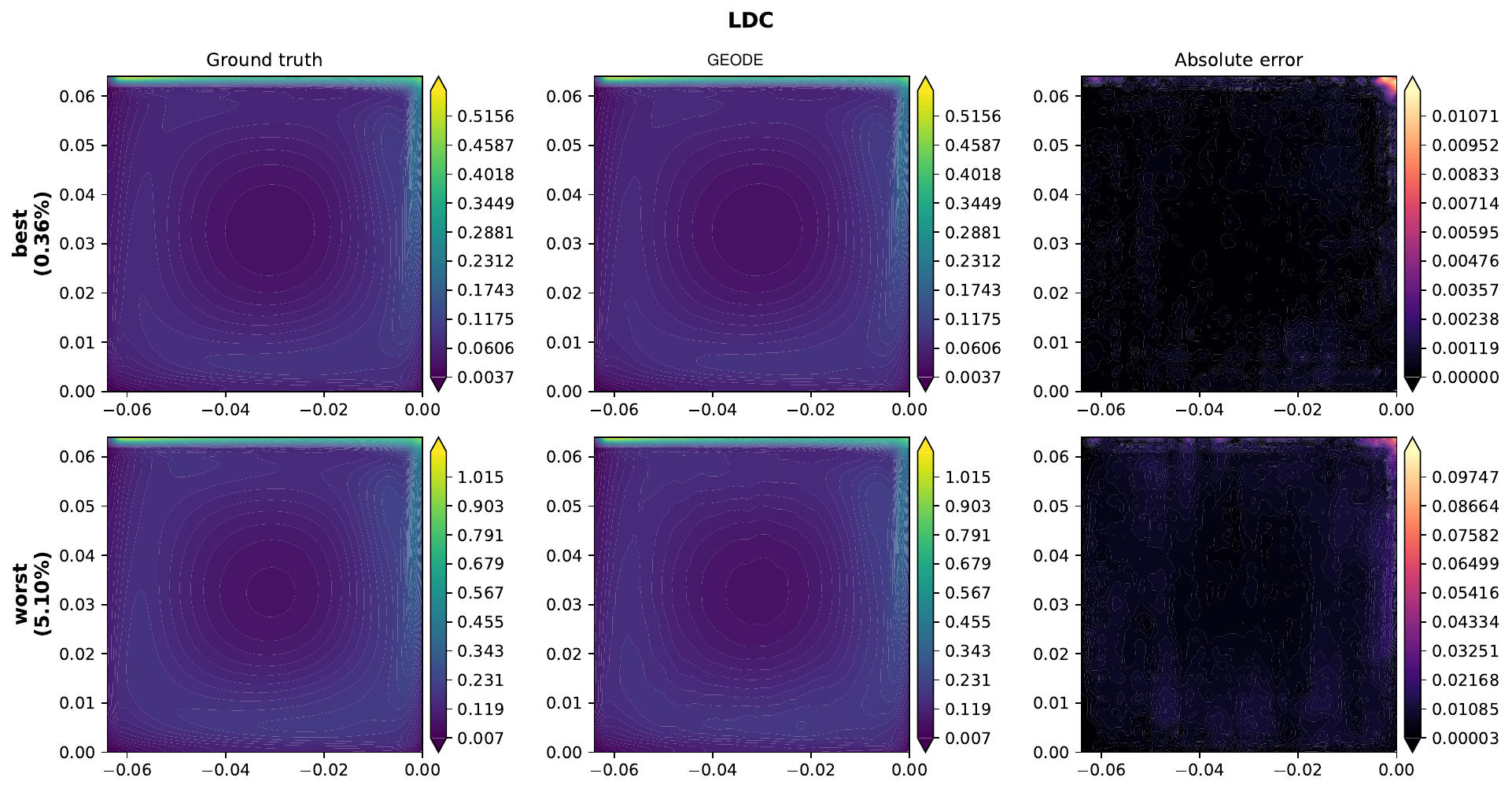}
  \caption{\textbf{LDC velocity predictions for the lowest- and highest-error test instances.} Columns show the reference $v$-velocity field, GEODE prediction and absolute pointwise error.}
  \label{fig:supp_ldc_fields}
\label{fig:ldc}
\end{figure}

\begin{figure}[htbp]
  \centering
  \includegraphics[width=\textwidth]{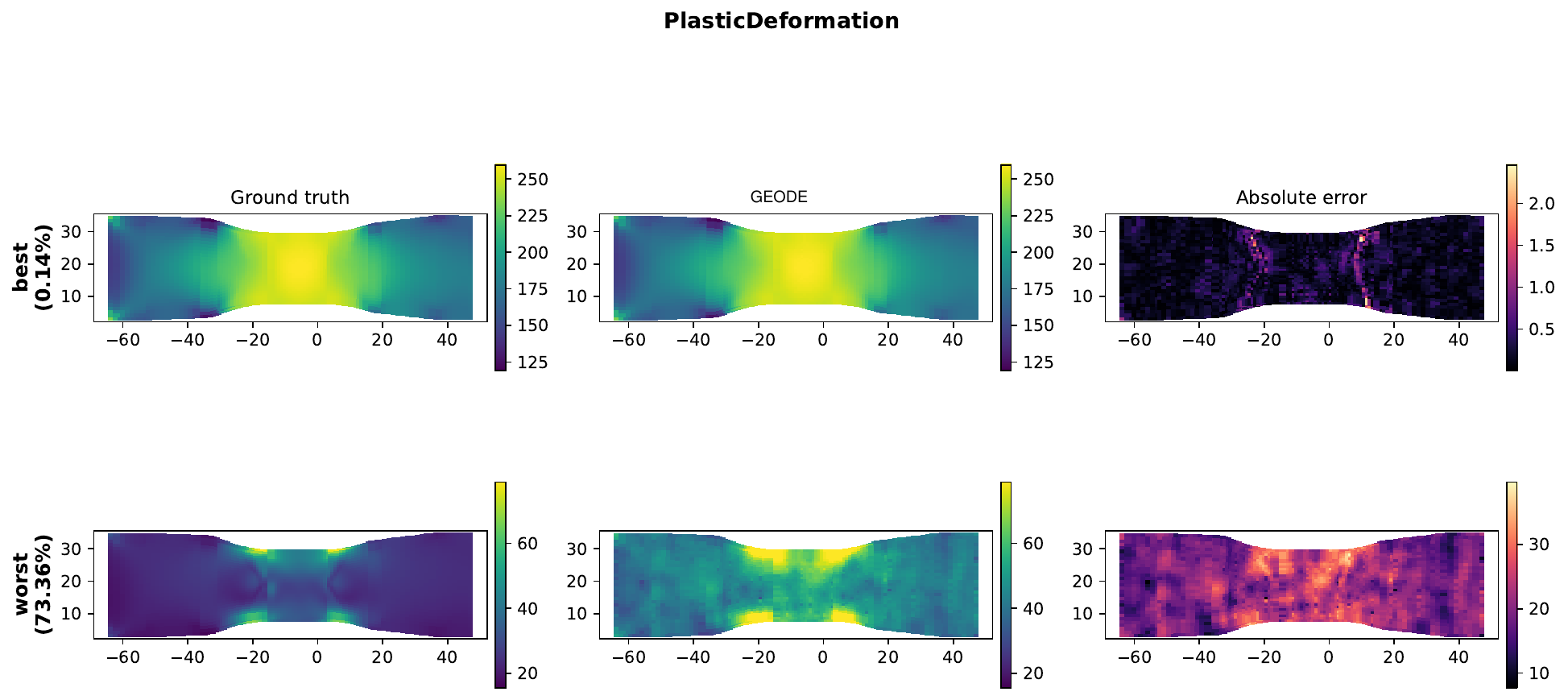}
  \caption{\textbf{PlasticDeform predictions for the lowest- and highest-error test instances.} Columns show reference von Mises stress, GEODE prediction and absolute pointwise error. The relative error of the worst case is amplified by the small target-field magnitude.}
  \label{fig:supp_plastic_fields}
\end{figure}

The lowest- and highest-error examples are shown separately for the three tasks in \Cref{fig:supp_cosmic_fields,fig:supp_ldc_fields,fig:supp_plastic_fields}. These complete panels complement the representative cases retained in the composite main-text figure.

\begin{figure}[htbp]
  \centering
  \includegraphics[width=\textwidth]{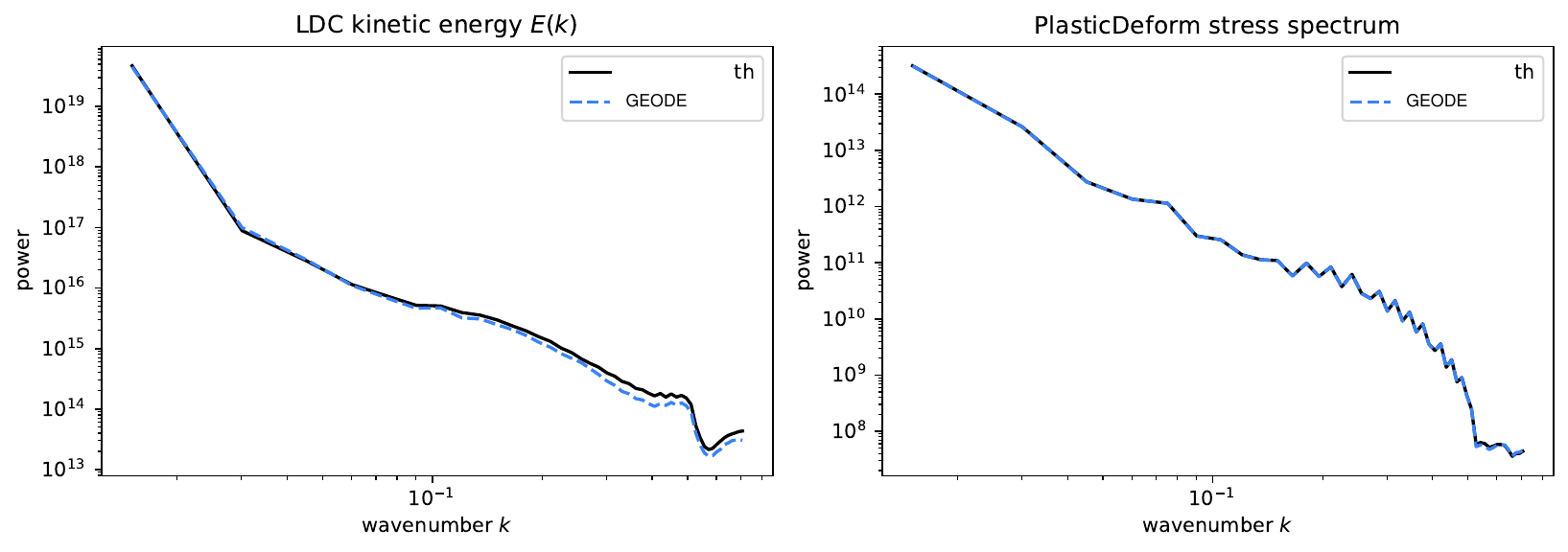}
  \caption{\textbf{Predicted and reference field content across spatial scales.} Radially averaged spectra are computed over ten test instances. The LDC prediction modestly underestimates the highest-wavenumber content.}
  \label{fig:supp_spectrum}
\end{figure}

The predicted PlasticDeform stress spectrum follows the reference across the resolved wavenumber range. For LDC, the kinetic-energy spectrum agrees at large and intermediate scales but shows modest attenuation at the highest wavenumbers (\Cref{fig:supp_spectrum}).

\begin{table}[htbp]
  \centering
  \caption{\textbf{Selected physical diagnostics not included in the training objective.} These empirical diagnostics do not imply exact satisfaction of conservation laws or constitutive constraints.}
  \label{tab:supp_physical_diagnostics}
  \small
  \setlength{\tabcolsep}{5pt}
  \renewcommand{\arraystretch}{1.15}
  \begin{tabular}{@{}lll@{}}
    \toprule
    \textbf{Task} & \textbf{Diagnostic} & \textbf{Result} \\
    \midrule
    LDC & Predicted/reference mean absolute divergence & 0.995 \\
    CosmicDose & Predicted/reference north--south asymmetry & 0.0159/0.0159 \\
    PlasticDeform & Fraction of nodes with negative von Mises stress & $0$ \\
    PlasticDeform & Minimum predicted von Mises stress & $2.49$ \\
    \bottomrule
  \end{tabular}
\end{table}

\subsubsection{Representational compatibility and external comparisons}

The source review covers MORPH, PDEformer-2 and PDE-FM~\citep{rautela2025morph,ye2025pdeformer2,soares2025pdefm}. The audit records what has actually been executed and distinguishes it from published input contracts. A flexible output geometry does not by itself supply a valid conditioning input. Conversely, a station vector or loading history does not establish that an alternative architecture is incapable of solving the task. The addition of a learned adapter must be disclosed for every method, including GEODE.

\begin{table}[htbp]
\centering\scriptsize
\caption{\textbf{External-model evidence and evaluation status.} Source review is not an executed benchmark; no numerical score is inferred from compatibility.}
\label{tab:supp_compatibility}
\begin{tabular}{@{}p{2.1cm}p{4.9cm}p{6.8cm}@{}}
\toprule Method & Interface evidence & Status for the present datasets \\
\midrule
DeepONet & Parameter/history branch and coordinate trunk & Clean evaluation on all five tasks; Plastic validation held out from training. \\
GINO & Official operator with our learned vector-to-field lift & Corrected-coordinate, clean-split evaluation completed on all three base tasks (30 epochs). \\
FNO, WNO, GNOT, Transolver & Task-specific operator families & Clean bounded evaluations completed with documented vector-input adapters (\Cref{tab:cx_wave_details}). \\
MPP, Poseidon & Published field-history interfaces & Present-dataset execution and conditioning contracts not yet verified. \\
DPOT & Public Fourier-attention operator transformer and checkpoints & Adapted execution verified on the tasks reported in \Cref{tab:cx_dpot}: learned vector lift and native-query readout. Native field-history input is not claimed; see \Cref{tab:cx_dpot}. \\
MORPH & Public multidimensional field-tensor model and checkpoints & Adapted execution verified on the tasks reported in \Cref{tab:cx_morph}: learned vector lift and native-query readout. Native field-history input is not claimed; see \Cref{tab:cx_morph}. \\
PDEformer-2 & PDE/condition graph, coordinate queries, public checkpoints and inverse examples & Source reviewed. Requires a sufficiently specified governing problem and observation/loading contract; no present-dataset runtime accuracy claim. \\
PDE-FM & Spatial--spectral state-space model with dataset adapters & Paper reviewed. Runnable checkpoint availability and present-dataset input contract remain unverified. \\
GEODE & Shared lift for the three base tasks; private interfaces for acquired tasks & Executed on all five. Fixed geometry within each task; the clean single-task PlasticDeform run did not converge (see \Cref{tab:supp_operator_comparison}). \\
\bottomrule
\end{tabular}
\end{table}

\Cref{tab:supp_operator_comparison} records completed outputs and explicitly unexecuted rows. Provisional and diagnostic entries are retained for traceability and must not support a final method ranking.

\begin{table}[htbp]
  \centering
  \caption{Median physical relative $L^2$ error (\%) from completed runs. ST denotes separate single-task training; pretr. FT denotes full fine-tuning of pretrained weights; POD denotes proper orthogonal decomposition. $^{\dagger}$ The clean single-task PlasticDeform run did not converge under the shared recipe (its validation loss sat at $0.90$ for the first sixty epochs) and is reported as obtained; the historical specialist reached $0.34\%$. The joint GEODE row is the clean signed-distance-enabled checkpoint, selected on validation data only. DeepONet and GINO use a training-only PlasticDeform validation split; GINO uses corrected coordinates and a learned vector-to-field lift. NE means not yet evaluated, not incompatible. Separate DeepONet specialists used 200 epochs and GINO 30. The additional clean baselines use up to 100 epochs, validation-based stopping and a 20-minute training/validation cap; their exact budgets and adapters are detailed in the supplement. Differing schedules and capacities prevent a matched-compute or converged-model ranking.}
  \label{tab:supp_operator_comparison}
  \scriptsize
  \setlength{\tabcolsep}{3pt}
  \renewcommand{\arraystretch}{1.15}
  \begin{tabular}{@{}llccc@{}}
    \toprule
    \textbf{Method} & \textbf{Protocol} & \textbf{LDC} & \textbf{CosmicDose} & \textbf{PlasticDeform} \\
    \midrule
    Per-node training mean & Control & 59.80 & 1.79 & 20.13 \\
    POD plus ridge & ST, clean & 10.9836 & 0.3378 & 20.1056 \\
    DeepONet & ST, clean & 0.236 & 0.0471 & 0.502 \\
    GINO + vector lift & ST, clean & 2.195 & 0.0654 & 1.083 \\
    FNO + adapters & ST, clean & 0.2708 & 0.0952 & 0.5034 \\
    WNO + adapters & ST, clean & 1.1889 & 0.0752 & 2.2866 \\
    GNOT + vector input & ST, clean & 0.4016 & 0.0432 & 1.5610 \\
    Transolver + vector input & ST, clean & 0.2189 & 0.0791 & 1.6689 \\
    Shared DeepONet & Joint, clean & 0.7425 & 0.0529 & 0.7253 \\
    MORPH-Ti + adapters & ST, pretr. FT & 1.1531 & 0.0787 & 1.3073 \\
    MORPH-Ti + adapters & ST, random init. & 1.0889 & 0.0632 & 0.8914 \\
    DPOT-Ti + adapters & ST, pretr. FT & 0.1548 & 0.0467 & 0.8485 \\
    DPOT-Ti + adapters & ST, random init. & 0.1991 & 0.0521 & 0.8354 \\
    NCWNO + adapters & Joint, clean & 11.3864 & 0.9450 & 5.9175 \\
    GEODE architecture & ST, clean & 0.385 & 0.038 & $11.8^{\dagger}$ \\
    GEODE & Joint, clean & \textbf{0.755} & \textbf{0.0307} & \textbf{0.399} \\
    MPP / Poseidon & Pretrained, not evaluated & NE & NE & NE \\
    PDEformer-2 & Pretrained, not evaluated & NE & NE & NE \\
    PDE-FM & Availability unverified & NE & NE & NE \\
    \bottomrule
  \end{tabular}
\end{table}

The matched GEODE specialists attain $0.385\%$ and $0.038\%$ on LDC and CosmicDose (the clean PlasticDeform specialist did not converge under the shared recipe, \Cref{tab:supp_operator_comparison}; the historical specialist reached $0.34\%$). The joint model is worse by a factor of $2.0$ on LDC and better by a factor of $1.2$ on CosmicDose.

\subsection{Routing through the expert library}
\label{sec:supp_routing}

\begin{figure}[htbp]
  \centering
  \includegraphics[width=\textwidth]{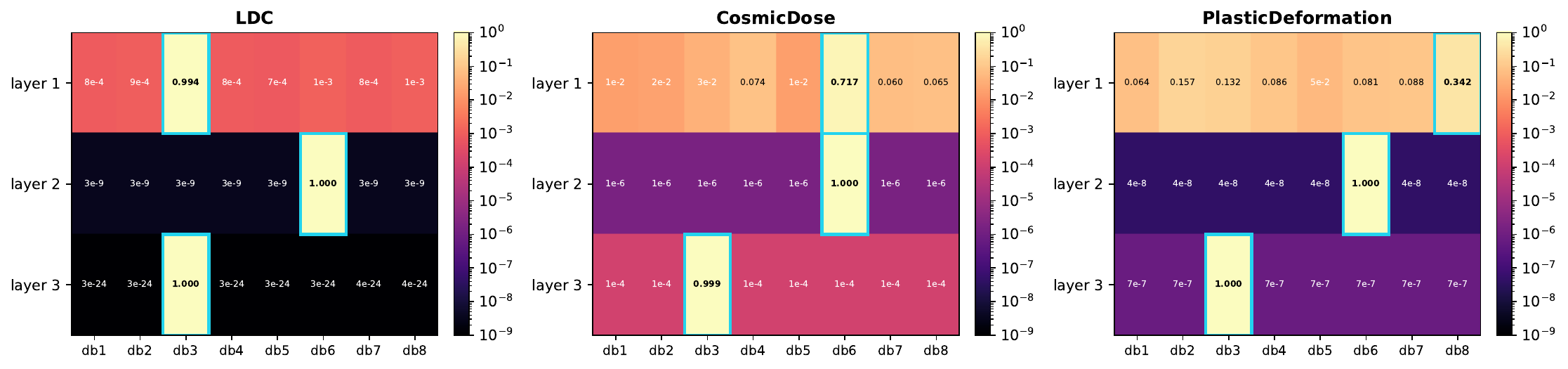}
  \caption{\textbf{Task-dependent routing through the shared wavelet-expert library.} Mean gate probability is shown for each expert and layer on a logarithmic color scale. The dominant expert is outlined.}
  \label{fig:supp_gate_activity}
\label{fig:routing}
\end{figure}

The gate distributions and effective expert counts reveal task-specific routing in the first layer and, for the pretrained tasks, a common dominant route in the deeper layers (\Cref{fig:supp_gate_activity,tab:supp_routes}). The effective expert count is the exponential of the entropy of the mean gate distribution; one indicates concentration on a single expert and eight indicates a uniform mixture.

\begin{table}[htbp]
  \centering
  \caption{\textbf{Dominant expert and effective expert count.} The value in parentheses is the effective number of experts. The first three rows are jointly pretrained tasks; HeatExchanger and Subchannel are acquired tasks evaluated with their trained interfaces.}
  \label{tab:supp_routes}
\label{tab:routing}
  \small
  \setlength{\tabcolsep}{5pt}
  \renewcommand{\arraystretch}{1.15}
  \begin{tabular}{@{}lccc@{}}
    \toprule
    \textbf{Task} & \textbf{Layer 1} & \textbf{Layer 2} & \textbf{Layer 3} \\
    \midrule
    LDC & db3 (1.1) & db6 (1.0) & db3 (1.0) \\
    CosmicDose & db6 (3.0) & db6 (1.0) & db3 (1.0) \\
    PlasticDeform & db8 (6.5) & db6 (1.0) & db3 (1.0) \\
    HeatExchanger & db5 (7.1) & db1 (2.0) & db5 (1.9) \\
    Subchannel$^{\mathrm{h}}$ & db8 (3.9) & db5 (2.0) & db6 (1.0) \\
    \bottomrule
  \end{tabular}
\end{table}

Hard-route interventions test whether the assignments are load-bearing (\Cref{tab:supp_route_intervention}). Each hard route forces the evaluated task through the dominant expert sequence of one pretrained task. Because the pretrained routes share their second- and third-layer experts, the cross-task intervention among them primarily changes the first-layer assignment; the acquired HeatExchanger route differs at every layer.

\begin{table}[htbp]
  \centering
  \caption{\textbf{Effect of routing interventions.} Entries are median relative $L^2$ errors (\%). Live gate uses the learned soft probabilities. The HeatExchanger route is that of its private interface after acquisition; the pretrained tasks' values are identical on the three-task checkpoint.}
  \label{tab:supp_route_intervention}
  \small
  \setlength{\tabcolsep}{4pt}
  \renewcommand{\arraystretch}{1.15}
  \begin{tabular}{@{}lccccc@{}}
    \toprule
    \textbf{Evaluated task} & \textbf{Live gate} & \textbf{LDC route} & \textbf{Cosmic route} & \textbf{Plastic route} & \textbf{Heat exch. route} \\
    \midrule
    LDC & 0.755 & 0.722 & 88.3 & 71.3 & 58.3 \\
    CosmicDose & 0.031 & 1.41 & 0.441 & 1.54 & 1.72 \\
    PlasticDeform & 0.399 & 16.1 & 21.8 & 1.44 & 46.7 \\
    HeatExchanger & 0.719 & 19.8 & 14.3 & 9.49 & 3.64 \\
    \bottomrule
  \end{tabular}
\end{table}

Cross-task substitution increases error by factors of approximately $13$ to $117$ relative to the live gate ($3$ to $122$ relative to the task's own hard route; the lower end arises because the CosmicDose one-hot route is itself far from its live gate). The live soft gate is more accurate than the task's own one-hot route by factors of $14$ and $3.6$ for CosmicDose and PlasticDeform, whose first-layer gates are diffuse, indicating that the lower-weight experts make a measurable contribution; for LDC the two are comparable ($0.96$). On the acquired HeatExchanger interface the live gate is $5.1$ times better than its own hard route and $13$ to $27$ times better than the pretrained routes. Applied to the norm-matched random library, all routes remain uniformly inaccurate and differ by at most $22\%$ (LDC $50$--$64\%$, CosmicDose $1.7$--$1.8\%$, PlasticDeform $44$--$48\%$), without a task-aligned pattern. Routing therefore carries no interpretable task information when the experts carry no learned structure.

\begin{figure}[htbp]
  \centering
  \includegraphics[width=\textwidth]{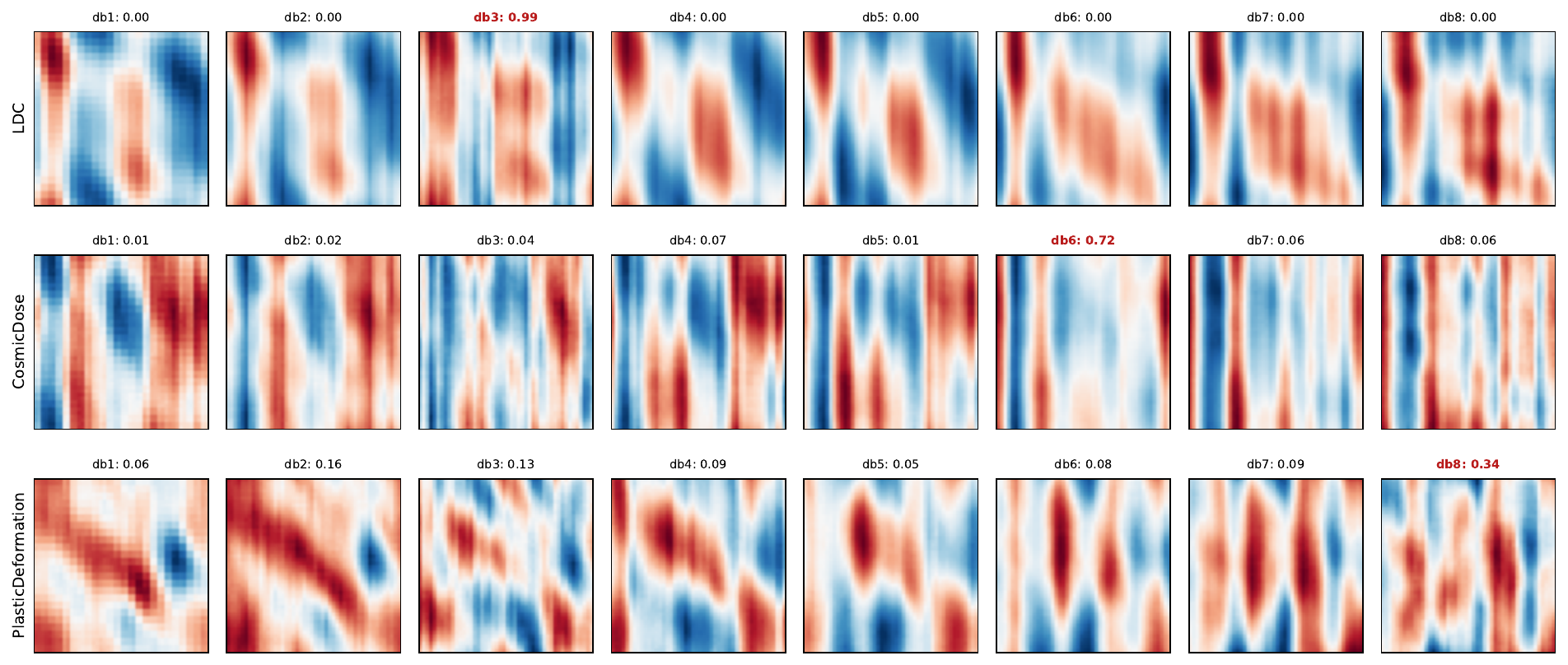}
  \caption{\textbf{Feature maps generated by the first-layer wavelet experts.} One latent channel is shown for a representative instance from each task. Each panel is annotated with its gate weight, and the dominant expert is highlighted. The panels visualize differentiated latent computations but do not by themselves establish causality.}
  \label{fig:supp_expert_contours}
\end{figure}

\begin{table}[htbp]
  \centering
  \caption{\textbf{Matched routing ablations.} Entries are median relative $L^2$ errors (\%). The dense backbone replaces the eight-expert mixture with one expert ($22$ million parameters). Without the label, all three tasks converge on one route (db7, db7, db1).}
  \label{tab:supp_routing_ablations}
  \small
  \setlength{\tabcolsep}{5pt}
  \renewcommand{\arraystretch}{1.15}
  \begin{tabular}{@{}lccc@{}}
    \toprule
    \textbf{Model} & \textbf{LDC} & \textbf{CosmicDose} & \textbf{PlasticDeform} \\
    \midrule
    Dense backbone & 4.75 & 0.102 & 0.379 \\
    Expert mixture without task label & 4.33 & 0.104 & 0.331 \\
    GEODE & 0.755 & 0.031 & 0.399 \\
    \bottomrule
  \end{tabular}
\end{table}

Re-presenting LDC through a newly initialized interface demonstrates that expert indices are not a unique taxonomy. On the paper checkpoint, whose LDC route is db3, db6, db3, a re-presented LDC with task separation enabled converges to db8, db5, db5 with effective counts $2.1$, $1.2$ and $2.8$ and reaches $0.99\%$. With task separation disabled, three nominally different seeds converge to db8 in the first block, db4, db4 or db3 in the second and db5 in the third, reaching $1.04\%$, $0.91\%$ and $0.92\%$, within $1.2$ to $1.4\times$ of the original interface; a defect found after these runs showed that the wavelet module reset the random seed at import, so the three share one initialization and their spread measures run-to-run noise, not seed dependence. The model therefore rediscovers an effective but non-identical allocation.

\subsection{Sequential acquisition and forgetting controls}
\label{sec:supp_sequential}

Routing alone does not acquire HeatExchanger. Updating only the task embedding and the $2.3$-million-parameter gate network leaves the error at $10.15\%$ and, because that network is shared, degrades the pretrained tasks. Updating the complete task interface with frozen experts reduces the error to $0.719\%$ under the same full-data protocol (\Cref{tab:supp_component_audit}).

\begin{table}[htbp]
  \centering
  \caption{\textbf{Component audit for heat-exchanger acquisition.} Errors are median relative $L^2$ errors (\%). The final three columns report earlier-task errors after adaptation. NA indicates that the specialist does not contain those tasks. All rows use the same 100-epoch, full-data protocol from the paper checkpoint; the full fine-tuning row uses batch size 8 on an A100.}
  \label{tab:supp_component_audit}
  \scriptsize
  \setlength{\tabcolsep}{3pt}
  \renewcommand{\arraystretch}{1.15}
  \begin{tabular}{@{}lccccc@{}}
    \toprule
    \textbf{Training strategy} & \textbf{Parameters} & \textbf{Heat exch.} & \textbf{LDC} & \textbf{Cosmic} & \textbf{Plastic} \\
    \midrule
    Specialist from scratch & all & 0.502 & NA & NA & NA \\
    Embedding and shared gate & 2.3M & 10.15 & 22.6 & 0.306 & 6.0 \\
    Task interface and gate & 3.9M & 0.719 & 0.755 & 0.031 & 0.399 \\
    Full fine-tuning & 190.6M & 9.38 & 16.4 & 0.905 & 5.65 \\
    Rehearsal (all earlier data) & 190.6M & 0.661 & 2.69 & 0.053 & 1.12 \\
    \bottomrule
  \end{tabular}
\end{table}

Full fine-tuning neither learns the target as accurately nor preserves the earlier tasks. Joint rehearsal reaches $0.661\%$ on HeatExchanger within the same budget, but requires all earlier data and leaves LDC at $2.69\%$, CosmicDose at $0.053\%$ and PlasticDeform at $1.12\%$. By contrast, the interface method changes no parameter or state used by an earlier task.

\begin{table}[htbp]
  \centering
  \caption{\textbf{Performance after two sequential acquisitions.} Errors are measured after Subchannel has been added. No earlier data are replayed and no shared expert parameter is updated. The HeatExchanger step is the full-data interface run of the component audit and is unchanged by the second step because its interface is frozen; specialists use 200 epochs. The Subchannel entry is the standard-loss variant of the step; the profile-aware variant reaches $0.183\%$ (\Cref{tab:supp_profile_loss}).}
  \label{tab:supp_sequential_acquisition}
  \small
  \setlength{\tabcolsep}{5pt}
  \renewcommand{\arraystretch}{1.15}
  \begin{tabular}{@{}llccc@{}}
    \toprule
    \textbf{Task} & \textbf{Stage} & \textbf{Specialist} & \textbf{GEODE} & \textbf{Ratio} \\
    \midrule
    LDC & Pretrained & NA & 0.755 & NA \\
    CosmicDose & Pretrained & NA & 0.031 & NA \\
    PlasticDeform & Pretrained & NA & 0.399 & NA \\
    HeatExchanger & Sequential step 1 & 0.507 & 0.719 & $1.42\times$ \\
    Subchannel & Sequential step 2 & 0.034 & 0.083 & $2.4\times$ \\
    \bottomrule
  \end{tabular}
\end{table}

\begin{figure}[htbp]
  \centering
  \includegraphics[width=\textwidth]{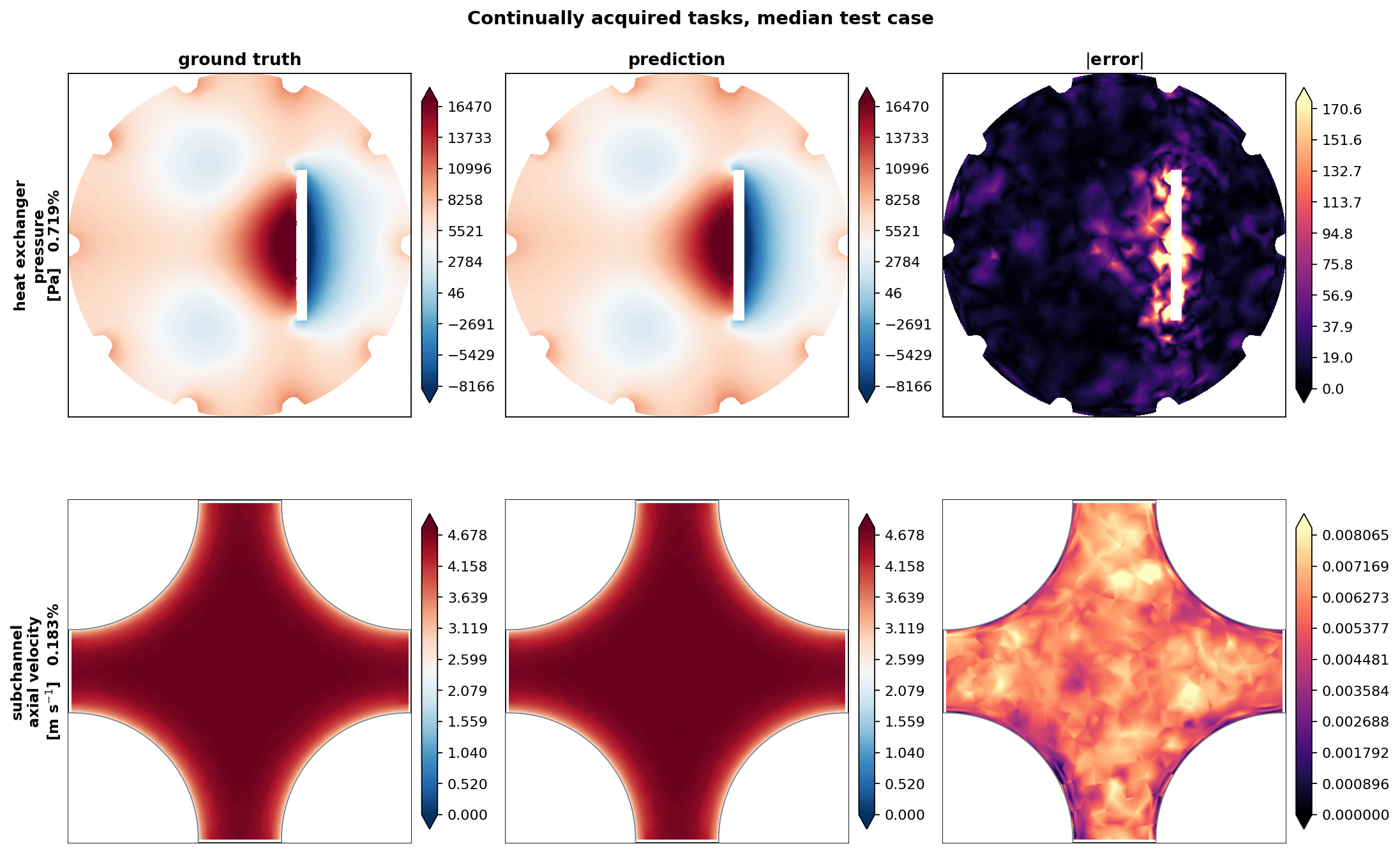}
  \caption{\textbf{Predictions for the sequentially acquired tasks.} Columns show the reference field, GEODE prediction and absolute pointwise error for a median-error test case after both acquisition steps in the clean profile-aware chain (HeatExchanger $0.719\%$, Subchannel $0.183\%$). HeatExchanger pressure is shown on its native unstructured CFD mesh; Subchannel axial velocity is shown on the cross-shaped coolant domain.}
  \label{fig:supp_continual_fields}
\label{fig:cl_fields}
\end{figure}

\subsection{Transfer by data fraction and field component}
\label{sec:supp_transfer}

The data-fraction sweep shows a crossover between the frozen-library interface and a specialist (\Cref{tab:supp_data_efficiency}). At $1\%$ of the HeatExchanger data, the trained library improves over both a specialist and the norm-matched random library. At $10\%$, the two libraries are within $4\%$ of each other and both trail the specialist. At $50\%$ and $100\%$, the specialist is more accurate.

\begin{table}[htbp]
  \centering
  \caption{\textbf{Heat-exchanger adaptation as a function of training-data availability.} Entries are median relative $L^2$ errors (\%) under one matched protocol (100 epochs from the paper checkpoint). NE denotes an experiment not performed. Gate-only adaptation also degrades the pretrained tasks once the fraction exceeds $1\%$ (LDC $0.76$, $3.4$, $57.9$ and $22.6\%$ at $1$, $10$, $50$ and $100\%$).}
  \label{tab:supp_data_efficiency}
  \small
  \setlength{\tabcolsep}{5pt}
  \renewcommand{\arraystretch}{1.15}
  \begin{tabular}{@{}lcccc@{}}
    \toprule
    \textbf{Adaptation strategy} & \textbf{1\%} & \textbf{10\%} & \textbf{50\%} & \textbf{100\%} \\
    \midrule
    Specialist from scratch & 8.92 & 0.707 & 0.494 & 0.502 \\
    Task embedding and gate only & 10.27 & 9.67 & 9.49 & 10.15 \\
    Interface, trained library & 5.88 & 1.32 & 0.877 & 0.719 \\
    Interface, random library & 7.50 & 1.37 & NE & 0.790 \\
    \bottomrule
  \end{tabular}
\end{table}

Norm-matched randomization preserves the architecture and weight scale but removes the functions learned during pretraining. The held-out tasks in \Cref{tab:supp_library_audit} are adapted independently from the same three-task checkpoint, rather than sequentially. Randomizing the shared library destroys earlier tasks and is therefore a diagnostic intervention, not a deployable continual-learning strategy.

\begin{table}[htbp]
  \centering
  \caption{\textbf{Contribution of learned expert content.} Entries are median physical-space relative $L^2$ errors (\%). The dense backbone is adapted from its own three-task checkpoint. All entries are single runs on the paper checkpoint.}
  \label{tab:supp_library_audit}
  \scriptsize
  \setlength{\tabcolsep}{3pt}
  \renewcommand{\arraystretch}{1.15}
  \begin{tabular}{@{}lccccc@{}}
    \toprule
    \textbf{Frozen backbone} & \textbf{LDC} & \textbf{Cosmic} & \textbf{Plastic} & \textbf{Heat exch.} & \textbf{Subchannel} \\
    \midrule
    Trained wavelet experts & 0.755 & 0.031 & 0.399 & 0.719 & 0.121 \\
    Norm-matched random experts & 63.7 & 1.75 & 45.8 & 0.790 & 4.524 \\
    Dense backbone without experts & 4.749 & 0.102 & 0.379 & 0.811 & NA \\
    \bottomrule
  \end{tabular}
\end{table}

At full data, learned expert content is worth about $10\%$ for HeatExchanger: the trained library gives $0.719\%$, compared with $0.790\%$ for the random library and $0.811\%$ for the dense backbone. Subchannel gives the opposite result, with $0.121\%$ for the trained library and $4.52\%$ after randomization, the latter at the constant-level solution ($5.12\%$ for the training-mean field).

\begin{table}[htbp]
  \centering
  \caption{\textbf{Subchannel spatial-profile audit.} Each channel is decomposed into its spatial mean and zero-mean profile. Values are profile errors (\%), with the temperature-level error in kelvin in the last column. Paper checkpoint, cold-start protocol of the historical run, single runs.}
  \label{tab:supp_profile_audit}
  \small
  \setlength{\tabcolsep}{6pt}
  \renewcommand{\arraystretch}{1.15}
  \begin{tabular}{@{}lcccc@{}}
    \toprule
    \textbf{Frozen library} & \textbf{Channel 1} & \textbf{Channel 2} & \textbf{Channel 3} & \textbf{Level error (K)} \\
    \midrule
    Trained experts & 0.75 & 7.62 & 0.27 & 0.47 \\
    Norm-matched random experts & 0.76 & 13.6 & 0.56 & 25.6 \\
    \bottomrule
  \end{tabular}
\end{table}

The profile differences in \Cref{tab:supp_profile_audit} are below a factor of two, whereas the complete-field errors differ by a factor of $37$. What separates the two libraries is whether the field level is learned at all: predicting the training-mean field gives $5.12\%$ on the test split, which is where the randomized library sits. This is important for the Subchannel temperature, whose mean is approximately $574$~K while its spatial variation spans only about $21$~K.

\begin{table}[htbp]
  \centering
  \caption{\textbf{Effect of a spatial-profile loss.} The first two rows compare two Subchannel specialists trained identically for 200 epochs except for the profile term; the chain row compares the two variants of the regenerated chain step on the paper checkpoint. On the earlier checkpoint the chain row read $0.161$ and $0.063$, that is, the profile term helped there and hurts here; single runs in every cell.}
  \label{tab:supp_profile_loss}
  \small
  \setlength{\tabcolsep}{5pt}
  \renewcommand{\arraystretch}{1.15}
  \begin{tabular}{@{}lccc@{}}
    \toprule
    \textbf{Protocol} & \textbf{Metric} & \textbf{Standard loss} & \textbf{Profile-aware loss} \\
    \midrule
    Subchannel specialists & Temperature profile error (\%) & 3.12 & 1.74 \\
    Subchannel specialists & Headline error (\%) & 0.034 & 0.034 \\
    Continual-acquisition chain & Subchannel error (\%) & 0.083 & 0.183 \\
    \bottomrule
  \end{tabular}
\end{table}

For HeatExchanger, the effective expert counts are $1.0$, $1.5$ and $3.4$ with learned experts and $2.2$, $1.3$ and $2.0$ after norm-matched randomization. The gate therefore responds to expert content. The acquired Subchannel uses effective counts of $2.1$, $2.0$ and $5.7$ with the trained library and $1.3$, $2.4$ and $2.5$ after randomization, showing that dense recombination is not required uniformly across target systems or layers.

\subsection{Geometry-adaptive decoding and computational cost}
\label{sec:supp_geometry_cost}

To alter the query discretization without retraining, we randomly selected half of the nodes from each original output mesh and rebuilt the decoder graph. Because the coordinates are drawn from the original mesh, this experiment tests stability to a changed query set, not generalization to a new geometry.

\begin{table}[htbp]
  \centering
  \caption{\textbf{Evaluation after reducing the output query set.} Entries are median relative $L^2$ errors (\%).}
  \label{tab:supp_reduced_query}
  \small
  \setlength{\tabcolsep}{6pt}
  \renewcommand{\arraystretch}{1.15}
  \begin{tabular}{@{}lccc@{}}
    \toprule
    \textbf{Query set} & \textbf{LDC} & \textbf{CosmicDose} & \textbf{PlasticDeform} \\
    \midrule
    Complete mesh, matched $n=32$ & 0.862 & 0.0745 & 0.338 \\
    Random $50\%$, same $n=32$ & 0.878 & 0.0774 & 0.345 \\
    \bottomrule
  \end{tabular}
\end{table}

\begin{figure}[htbp]
  \centering
  \includegraphics[width=\textwidth]{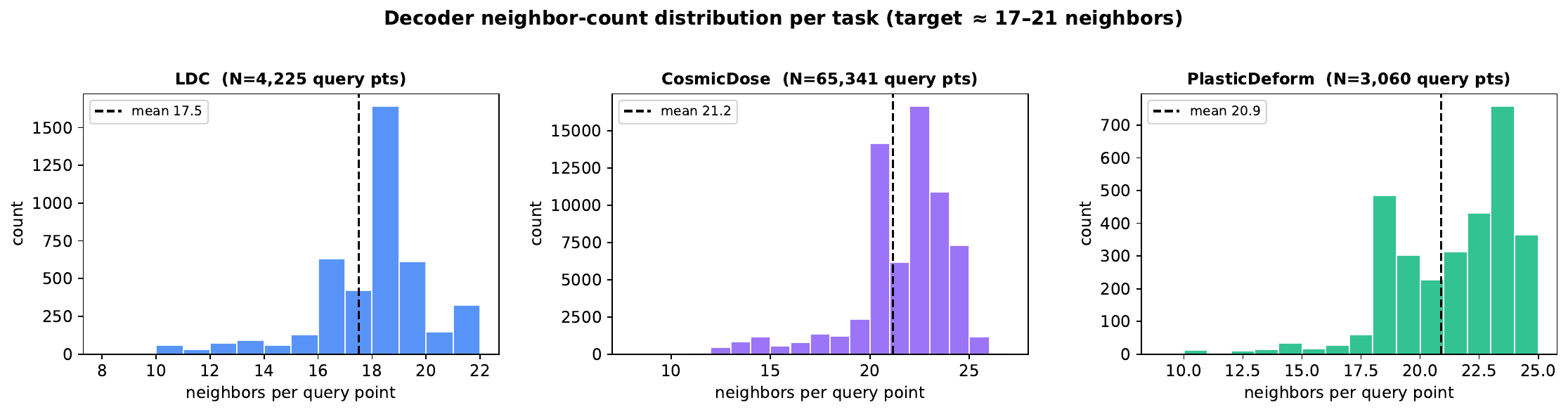}
  \caption{\textbf{Distribution of decoder neighborhood sizes.} The nominal task-specific radii give mean neighbor counts of $17.5$, $21.2$ and $20.9$ for LDC, CosmicDose and PlasticDeform.}
  \label{fig:supp_neighbours}
\end{figure}

\begin{table}[htbp]
  \centering
  \caption{\textbf{Sensitivity to the decoder neighborhood radius.} Entries are median relative $L^2$ errors (\%) on the same 32 examples per task; all learned parameters are fixed.}
  \label{tab:supp_radius}
  \small
  \setlength{\tabcolsep}{6pt}
  \renewcommand{\arraystretch}{1.15}
  \begin{tabular}{@{}lccc@{}}
    \toprule
    \textbf{Radius multiplier} & \textbf{LDC} & \textbf{CosmicDose} & \textbf{PlasticDeform} \\
    \midrule
    $0.75\times$ & 1.12 & 0.0742 & 0.327 \\
    $1.00\times$ & 0.862 & 0.0745 & 0.338 \\
    $1.50\times$ & 1.58 & 0.0770 & 0.418 \\
    $2.00\times$ & 2.02 & 0.0793 & 0.499 \\
    \bottomrule
  \end{tabular}
\end{table}

\begin{figure}[htbp]
  \centering
  \includegraphics[width=\textwidth]{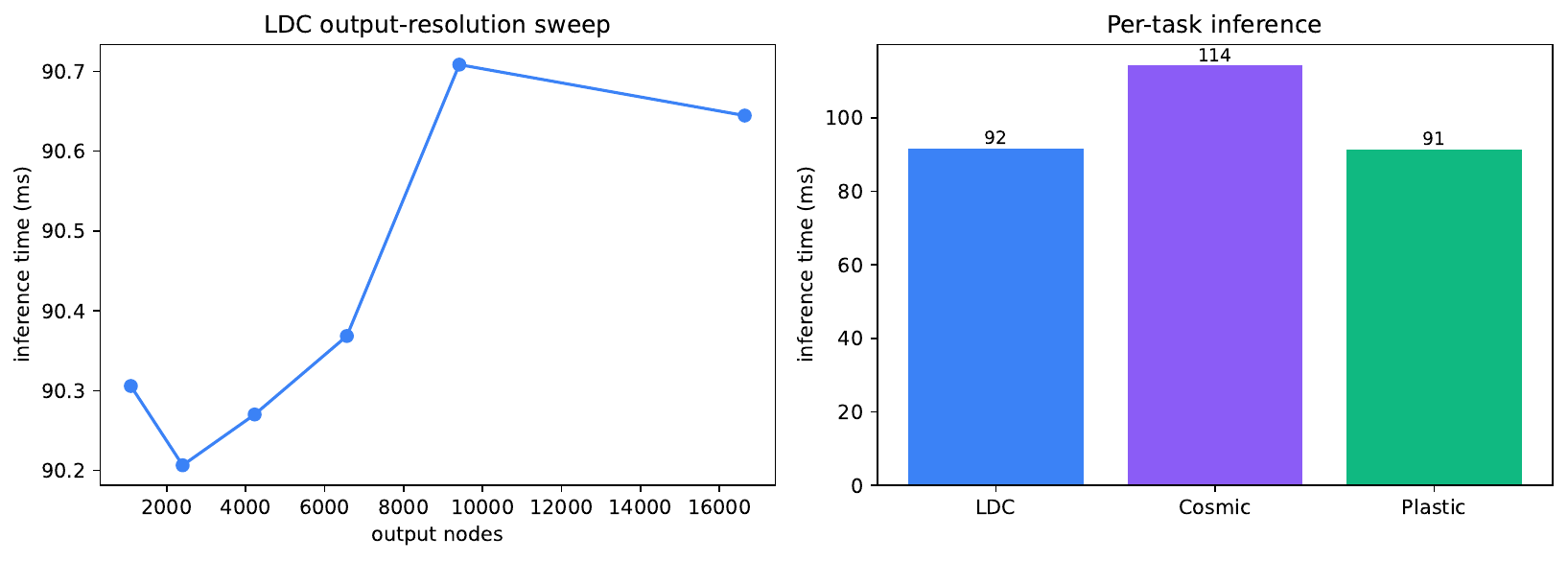}
  \caption{\textbf{Inference cost as the output discretization grows.} Measurements use batch size one in float32 on one NVIDIA A40. PlasticDeform requires $91$~ms for $3{,}060$ nodes, LDC requires $91$~ms for $4{,}225$ nodes, and CosmicDose requires $114$~ms for $65{,}341$ nodes.}
  \label{fig:supp_scalability}
\end{figure}

\begin{table}[htbp]
  \centering
  \caption{\textbf{System-level cost of matched GEODE specialists and the shared model.} Parameter counts are totals needed to serve the indicated task set. The Artifacts column counts separately deployed parameter sets. GPU-hours are measured per-epoch times of the clean runs on one NVIDIA A40 multiplied by the epochs run ($120$ for the base and its specialists, $100$ for the HeatExchanger interface, $200$ for the HeatExchanger and Subchannel specialists and the Subchannel interface); the three base-task specialists are priced from a timing probe of the same configurations on the same A40. The marginal rows refer specifically to HeatExchanger, not to a universal cost per new task.}
  \label{tab:supp_cost}
  \scriptsize
  \setlength{\tabcolsep}{3pt}
  \renewcommand{\arraystretch}{1.15}
  \begin{tabular}{@{}llccc@{}}
    \toprule
    \textbf{Scope} & \textbf{Configuration} & \textbf{Parameters} & \textbf{GPU-hours} & \textbf{Artifacts} \\
    \midrule
    Three tasks & Three specialists & 571.8M & 26.9 & 3 \\
    Three tasks & GEODE & 190.6M & 26.4 & 1 \\
    Add HeatExchanger & New specialist & $+190.6$M & $+1.5$ & $+1$ \\
    Add HeatExchanger & New interface & $+3.9$M & $+4.1$ & 0 \\
    Five tasks & Five specialists & 953.0M & 31.9 & 5 \\
    Five tasks & GEODE plus two interfaces & 198.5M & 40.4 & 1 \\
    \bottomrule
  \end{tabular}
\end{table}

Joint pretraining is approximately compute neutral relative to the three matched specialists, but adaptation through a frozen backbone is not compute neutral. The HeatExchanger interface adds about 49 times fewer trainable parameters than its matched specialist but takes $4.1$ GPU-hours rather than $1.5$ GPU-hours because forward computation still traverses the frozen library; the Subchannel interface takes $9.9$ GPU-hours against $3.5$.

\begin{table}[htbp]
  \centering
  \caption{\textbf{Indicative simulation and surrogate evaluation times.} Reference-solver times are wall-clock times per sample on NCSA Delta CPU nodes (dual AMD EPYC 7763, 128 cores per node); surrogate times are the A40 latencies of \Cref{fig:supp_scalability}. The two use different software and hardware and are not a controlled benchmark. NA indicates that an exact matched inference latency was not reported for that task.}
  \label{tab:supp_solver_timing}
  \small
  \setlength{\tabcolsep}{6pt}
  \renewcommand{\arraystretch}{1.15}
  \begin{tabular}{@{}lccc@{}}
    \toprule
    \textbf{Task} & \textbf{Reference solver} & \textbf{GEODE inference} & \textbf{Indicative speed-up} \\
    \midrule
    LDC & approximately 25 min & 91 ms & approximately $16{,}500\times$ \\
    CosmicDose & approximately 61 s & 114 ms & approximately $535\times$ \\
    PlasticDeform & approximately 5 min & 91 ms & approximately $3{,}300\times$ \\
    HeatExchanger & approximately 18 min & NA & NA \\
    Subchannel & approximately 12 min & NA & NA \\
    \bottomrule
  \end{tabular}
\end{table}

\subsection{Robustness to input perturbations and task-label interventions}
\label{sec:supp_robustness}

We additionally perturbed the normalized input vectors and measured the resulting change in prediction error. The results in \Cref{tab:supp_input_perturbation} come from the robustness protocol and should not be substituted for the frozen headline values. Small perturbations leave the gate nearly unchanged: the maximum movement in the gate distribution is $6\times10^{-4}$ in $L^1$ distance, and the dominant experts do not change.

\begin{table}[htbp]
  \centering
  \caption{\textbf{Error inflation under additive input perturbations.} Entries are the ratio of perturbed to unperturbed median error for perturbation magnitude $\epsilon$ in normalized input space.}
  \label{tab:supp_input_perturbation}
  \small
  \setlength{\tabcolsep}{7pt}
  \renewcommand{\arraystretch}{1.15}
  \begin{tabular}{@{}lccc@{}}
    \toprule
    \textbf{Task} & $\boldsymbol{\epsilon=0.01}$ & $\boldsymbol{\epsilon=0.05}$ & $\boldsymbol{\epsilon=0.10}$ \\
    \midrule
    LDC & 0.99 & 1.28 & 2.46 \\
    CosmicDose & 1.06 & 1.33 & 1.80 \\
    PlasticDeform & 1.03 & 1.11 & 1.27 \\
    \bottomrule
  \end{tabular}
\end{table}

Wrong-label interventions test whether task labels are functionally used rather than simply correlated with the learned representations. They produce large errors across all three tasks (\Cref{tab:supp_wrong_label}). The baseline values are those of the robustness-protocol subset and differ slightly from the full-test-set headline values.

\begin{table}[htbp]
  \centering
  \caption{\textbf{Effect of supplying the wrong task label.} Entries are median relative $L^2$ errors (\%) under the task-label intervention protocol. Each wrong-label column supplies the label of the indicated alternative task while retaining the evaluated task's input.}
  \label{tab:supp_wrong_label}
  \small
  \setlength{\tabcolsep}{5pt}
  \renewcommand{\arraystretch}{1.15}
  \begin{tabular}{@{}lccc@{}}
    \toprule
    \textbf{Evaluated task} & \textbf{Correct label} & \textbf{Wrong label 1} & \textbf{Wrong label 2} \\
    \midrule
    LDC & 0.753 & 82.8 & 64.0 \\
    CosmicDose & 0.074 & 0.561 & 0.396 \\
    PlasticDeform & 0.320 & 16.0 & 13.0 \\
    \bottomrule
  \end{tabular}
\end{table}

A linear probe trained on the padded input vector predicts task identity with $89\%$ accuracy, compared with $33\%$ chance accuracy. Because the tasks have different original input dimensions, the padding pattern itself is informative. This result explains why the model without an explicit task label can recover part of the task identity, while the wrong-label intervention shows that the explicit label remains causally influential when supplied.

\subsection{Specialist baselines and implementation diagnostics}
\label{sec:supp_baseline_status}

\begin{table}[htbp]
\centering\small
\caption{\textbf{Completed DeepONet specialists on all five tasks.} Errors are physical relative $L^2$ percentages. These are separate models, not sequential additions to a shared library. The updated runs use 200 epochs, batch size 32 and validation-selected weights. PlasticDeform uses 13,410 training and 1,490 validation examples; its test set is reserved for evaluation. Different schedules prevent a matched-compute claim.}
\label{tab:supp_deeponet_five}
\begin{tabular}{@{}lrrrr@{}}
\toprule Task & Test $n$ & Median (\%) & IQR (\%) & Parameters \\
\midrule
LDC & 495 & 0.2357 & 0.2002--0.3726 & 418,819 \\
CosmicDose & 359 & 0.04705 & 0.02300--0.06685 & 351,489 \\
PlasticDeform & 100 & 0.5022 & 0.2997--2.2549 & 355,841 \\
HeatExchanger & 173 & 0.5718 & 0.2930--0.8159 & 421,891 \\
Subchannel & 1,000 & 0.1595 & 0.07644--0.2850 & 421,891 \\
\bottomrule
\end{tabular}
\end{table}

\begin{table}[htbp]
\centering\small
\caption{\textbf{Corrected GINO with a learned vector-to-field lift.} Clean validation-selected results after 30 epochs, evaluated at all output nodes. IQR and percentiles describe test instances, not training-seed variability. Parameters count complex weights as two real scalars.}
\label{tab:supp_gino_clean}
\begin{tabular}{@{}lrrrrr@{}}
\toprule Task & Test $n$ & Median (\%) & IQR (\%) & P95 (\%) & Parameters \\
\midrule
LDC & 495 & 2.1951 & 1.6083--2.8930 & 3.7822 & 847,287 \\
CosmicDose & 359 & 0.0654 & 0.0497--0.0914 & 0.1462 & 822,453 \\
PlasticDeform & 100 & 1.0825 & 0.7359--2.8396 & 25.6681 & 892,085 \\
\bottomrule
\end{tabular}
\end{table}

The corrected run uses Adam at initial learning rate $10^{-3}$ with cosine annealing. Batch sizes are 32 for LDC and PlasticDeform and 16 for CosmicDose. Training uses at most 4,000 sampled query nodes for LDC/PlasticDeform and 6,000 for CosmicDose; validation and test evaluation use every node. Validation is checked every two epochs and the lowest validation-MSE state is restored before test evaluation: epoch 30 for LDC and PlasticDeform, and epoch 12 for CosmicDose. PlasticDeform reserves the final 10\% of its original training pool before fitting normalization. These are full training-pool runs with node subsampling, not few-shot pilots or pretrained-FM deployment.

\begin{table}[htbp]
\centering\small
\caption{\textbf{GINO implementation diagnostics, excluded from a final accuracy ranking.} All three runs used 30 epochs and inconsistent query-coordinate normalization. PlasticDeform additionally used test-based checkpoint selection. These numbers are retained as debugging evidence, not an estimate of a correctly implemented GINO benchmark.}
\label{tab:supp_gino_diagnostic}
\begin{tabular}{@{}lrrp{6.5cm}@{}}
\toprule Task & Test $n$ & Median (\%) & Qualification \\
\midrule
LDC & 495 & 59.21 & Coordinate mismatch; most queries lose source neighbors. \\
CosmicDose & 359 & 0.08370 & Coordinate mismatch compresses queries near the latent-grid center. \\
PlasticDeform & 100 & 19.27 & Coordinate mismatch and test-based checkpoint selection. \\
\bottomrule
\end{tabular}
\end{table}

Corrected implementation pilots used 128 training and 32 validation examples, without reading test targets. LDC reached a validation median of $12.69\%$ after a ten-epoch pilot. PlasticDeform reached $6.789\%$ at the selected epoch 20 of a 100-epoch diagnostic; its later validation deterioration indicates overfitting rather than inability to fit the training set. These small-support validation measurements demonstrate that the implementation can learn, but are not comparable to the full-data test columns. Strict checkpoint restoration and prediction round trips passed. The completed three-task GINO diagnostic occupied one GPU for 11 minutes 7 seconds; this is job elapsed time, not a matched training-efficiency benchmark.

The historical canonical GEODE checkpoint is the width-48, three-task model selected at epoch 116. Its stored state has 190.593 million real-scalar-equivalent elements, including buffers. Its decoder kernel input width is 11 (three latent features, two source coordinates, two query coordinates and four harmonic features), confirming that signed distance is disabled. Exact trainable-parameter counts must use the same complex-as-two-real convention for every method. The reproducibility bundle records source hashes, model identity and figure provenance; it does not certify independence of the historical PlasticDeform test set.

\subsection{Training protocol and provenance}
\label{sec:training_details}
The historical canonical checkpoint records width 48, eight experts, three layers, a $48\times48$ latent grid, 120 training epochs, batch size 16 per task, Adam with initial learning rate $10^{-3}$ and weight decay $10^{-5}$, and a step factor of 0.7 every 20 epochs. Its selected checkpoint is epoch 116; the clean paper checkpoint uses the same arguments and is selected at epoch 118 on validation data only. These are stored run arguments; the checkpoint does not embed an executed-source hash. Historical source files and the saved configuration provide its implementation lineage.

The reported training pools contain 3,949 LDC, 4,011 CosmicDose and 14,900 PlasticDeform instances; the nominal test populations contain 495, 359 and 100. LDC and CosmicDose supply separate validation arrays. Historical PlasticDeform instead reused its test arrays for validation; new runs reserve the final 10\% of its original training pool for validation and remove those rows before fitting normalization. Source-dataset partitioning and independence between related trajectories require separate checks and are not established by distinct array paths alone. The historical nearest-neighbor-distance ratios are distribution diagnostics, not proof of absence of temporal leakage.

The updated tables summarize individual trained models. They do not report independent-seed means or standard deviations. Percentiles describe variation across test instances. No final ranking is drawn from the test-selected PlasticDeform results or coordinate-inconsistent GINO runs.

For a test instance, relative error is $100\|\hat u-u\|_2/\|u\|_2$ in physical units, with nodes and output channels included in the norm. The mean-field skill score is $1-\sum_i\|\hat u_i-u_i\|_2^2/\sum_i\|\bar u_{\rm train}-u_i\|_2^2$, where $\bar u_{\rm train}$ is the per-node training mean. Neither a small full-field relative error nor this skill score alone establishes conservation-law satisfaction or accurate low-amplitude spatial profiles.

\subsection{Operator and shared-model comparisons}

\label{sec:cx_wave_protocol}

All runs in \Cref{tab:cx_wave_details} use the same clean splits described above, training-only normalization, Adam at initial learning rate $10^{-3}$, cosine scheduling over at most 100 epochs and gradient clipping at norm 1. Up to 4,096 training queries are sampled per minibatch; validation and test evaluation use all native output nodes. Validation is checked at epoch 1 and every five epochs. Training stops after 30 or more epochs if no improvement of at least 1\% has occurred for 20 epochs, or at the 20-minute training/validation cap. A failed-learning gate after 20 epochs is triggered when both training and best validation have improved by less than 10\%; failed gates yield no accuracy entry. No stopping decision uses test scores.

Single-task transformer batches contain eight examples; grid operators and shared DeepONet use 32. GNOT and Transolver use three layers of width 64 and four attention heads, with two GNOT experts or 32 Transolver slices. FNO uses three layers of width 32 and 16 Fourier modes per axis. WNO uses three layers of width 32, db6 wavelets and three decomposition levels. Both grid operators receive a learned vector-to-four-channel $64\times64$ lift and use bilinear output queries. Adapted NCWNO uses the authors' six dual-tree wavelet experts, width 24, three blocks, two decomposition levels and a $32\times32$ learned lift with bilinear queries. These adapters do not rasterize the target fields; losses are evaluated at their supplied nodes. They are additions to the published input/output interfaces and are counted in model size.

Shared DeepONet pads each input to 101 entries, appends a three-entry task indicator and shares a branch and trunk with 128 basis functions and width-256 hidden layers. Joint training interleaves complete task pools, weighting each minibatch loss inversely by its task's batch count; each minibatch receives a separate Adam update. Selection minimizes mean validation MSE across the three tasks. Consequently, joint selection, single-task selection and independently trained specialists are different protocols. None of these new joint rows establishes sequential retention.

\begin{table}[htbp]

\centering\small

\caption{\textbf{Verified clean table-completion runs.} Median physical relative $L^2$ errors are percentages. Epochs show completed/selected epochs; W denotes the wall-time cap, P validation-plateau stopping and E completion of the epoch budget. Model size includes adapters and counts complex weights as two real scalars. Joint model size is repeated across its task rows, not summed.}

\label{tab:cx_wave_details}

\begin{tabular}{@{}llrrr@{}}

\toprule Method & Task & Median (\%) & Epochs & Parameters (M) \\

\midrule

Shared DeepONet & LDC & 0.7425 & 100/100 E & 0.422 \\

Shared DeepONet & CosmicDose & 0.0529 & 100/100 E & 0.422 \\

Shared DeepONet & PlasticDeform & 0.7253 & 100/100 E & 0.422 \\

FNO + adapters & LDC & 0.2708 & 100/95 E & 2.387 \\

FNO + adapters & CosmicDose & 0.0952 & 30/5 P & 2.289 \\

FNO + adapters & PlasticDeform & 0.5034 & 100/95 E & 2.567 \\

WNO + adapters & LDC & 1.1889 & 85/65 P & 5.050 \\

WNO + adapters & CosmicDose & 0.0752 & 40/20 P & 4.952 \\

WNO + adapters & PlasticDeform & 2.2866 & 40/35 P & 5.230 \\

GNOT + vector & LDC & 0.4016 & 44/40 W & 0.749 \\

GNOT + vector & CosmicDose & 0.0432 & 10/10 W & 0.749 \\

GNOT + vector & PlasticDeform & 1.5610 & 17/17 W & 0.751 \\

Transolver + vector & LDC & 0.2189 & 100/100 E & 0.112 \\

Transolver + vector & CosmicDose & 0.0791 & 28/5 W & 0.112 \\

Transolver + vector & PlasticDeform & 1.6689 & 47/40 W & 0.114 \\

NCWNO + adapters & LDC & 11.3864 & 4/4 W & 12.439 \\

NCWNO + adapters & CosmicDose & 0.9450 & 4/4 W & 12.439 \\

NCWNO + adapters & PlasticDeform & 5.9175 & 4/4 W & 12.439 \\

\bottomrule

\end{tabular}

\end{table}

Checkpoint replay runs on CPU after each GPU run; first-example CPU/GPU predictions must differ by at most $10^{-4}$ in relative norm and 0.002 in maximum normalized absolute difference. All saved error vectors must match the reported sample counts, medians and percentiles. Source hashes, author commits, raw results and replay checks accompany the revision. This verifies saved-output consistency, not statistical significance across random seeds.

\subsection{MORPH adaptation with matched random controls}

\label{sec:cx_morph_protocol}

We use the authors' MORPH-Ti implementation with its 9,864,408-parameter core, patch size 8, width 256, four transformer blocks, four axial-attention heads and 32 cross-attention heads. The public file named \texttt{morph-Ti-FM-max\_ar1\_ep225.pth} contains internal epoch metadata 231. Its SHA256, pinned source commit and repository revision identify the exact checkpoint in the accompanying evidence; the filename is not treated as proof of its training epoch. Loading uses the complete state dictionary with strict key matching.

The conditioning vector is mapped by a learned linear layer to three channels on an abstract $32\times32$ grid. The unchanged MORPH core receives tensor dimensions $(B,1,1,3,1,32,32)$; a learned channel projection and bilinear query readout return outputs at the supplied native coordinates. The lift consumes no target field. This latent tensor is not an observed physical field history, and the adapter is an addition to the published scientific input interface. No target-field rasterization is performed. Total LDC size is 10,143,972 parameters; the frozen-core arms train 279,564 adapter parameters.

Pretrained and randomly initialized cores use identical adapter initialization, clean data splits, training-only normalization and optimization budgets. Frozen cores remain in evaluation mode throughout adaptation. Full-training arms update all weights. Adam learning rates are $10^{-3}$ for adapters and $10^{-4}$ for trainable core weights, with batch size 32, cosine scheduling over at most 100 epochs and norm-1 gradient clipping. Training uses up to 4,096 sampled output nodes per minibatch; validation and test use all native nodes. The 20-minute training/validation cap and validation-based stopping rules match \Cref{sec:cx_wave_protocol}. Each selected checkpoint passes independent CPU replay before publication. Frozen-feature reuse and fully trainable initialization comparisons are kept separate.

\begin{table}[htbp]

\centering\small

\caption{\textbf{Adapted MORPH-Ti controls.} Errors are median physical relative $L^2$ percentages, with linearly interpolated interquartile ranges (IQR) and 95th percentiles. Epochs give completed/selected budgets. These are single-seed results with validation-based stopping, not converged or matched-compute rankings.}

\label{tab:cx_morph}

\begin{tabular}{@{}lllrrrr@{}}

\toprule Task & Core init. & Updates & Median & IQR & p95 & Epochs \\

\midrule

LDC & Pretrained & Adapters & 4.7442 & 3.353--9.859 & 20.172 & 95/95 \\

LDC & Random & Adapters & 35.7849 & 23.347--46.366 & 52.462 & 85/85 \\

LDC & Pretrained & All & 1.1531 & 0.899--2.346 & 9.595 & 60/40 \\

LDC & Random & All & 1.0889 & 0.806--2.263 & 9.820 & 95/75 \\

CosmicDose & Pretrained & All & 0.0787 & 0.048--0.120 & 0.248 & 55/35 \\

CosmicDose & Random & All & 0.0632 & 0.046--0.088 & 0.141 & 30/10 \\

Plastic & Pretrained & All & 1.3073 & 0.928--3.131 & 34.916 & 100/100 \\

Plastic & Random & All & 0.8914 & 0.653--2.975 & 17.280 & 100/100 \\

\bottomrule

\end{tabular}

\end{table}

\subsection{DPOT adaptation with matched random controls}
\label{sec:cx_dpot_protocol}

We use the authors' DPOT implementation with the public DPOT-Ti checkpoint (\texttt{model\_Ti.pth}, Hugging Face repository \texttt{hzk17/DPOT}, revision \texttt{2adec1cf}): a 7,534,643-parameter adaptive-Fourier-mixing transformer with image size 128, patch size 8, embedding width 512, 32 retained modes, four blocks, four input channels, ten input time steps and one output step. The checkpoint's stored \texttt{argparse.Namespace} metadata is the only non-tensor object allowed at load time, and the weights are loaded strictly.

The conditioning vector is mapped by a learned linear layer to 40 channels on an abstract $16\times16$ grid, bilinearly upsampled to $128\times128$ and presented to the unchanged core as ten latent slots with four channels. These learned slots are not an observed physical field history and contain no target-field input. A learned $1\times1$ channel projection and bilinear query readout return outputs at the native coordinates of each task. The adapted models contain 8.41 to 8.58 million parameters depending on the task's input width and output channels. Adapter layers are initialized with a fixed seed shared by the pretrained and random arms; the random arm re-initializes the core with a second fixed seed.

Pretrained and randomly initialized cores use identical adapter initialization, clean data splits, training-only normalization and optimization budgets: Adam with learning rate $10^{-3}$ for the adapters and $10^{-4}$ for the core, at most 100 epochs and 1,200 seconds of training per arm, checkpoint selection on the validation split, and early stopping on a validation plateau. Selected checkpoints are re-evaluated in full float32 precision with TensorFloat-32 (TF32) disabled, because the reduced-precision cuDNN inference used during training produced test errors that differed from an independent CPU replay; the float32 evaluation reproduces the CPU replay to within $10^{-4}$ relative error per sample for every arm. Training and selection were not repeated.

\begin{table}[htbp]
\centering\small
\caption{\textbf{Adapted DPOT-Ti controls.} Errors are median physical relative $L^2$ percentages, with linearly interpolated interquartile ranges (IQR) and 95th percentiles. Epochs give completed/selected budgets; LDC and PlasticDeform arms stopped at the wall-clock budget, CosmicDose arms at a validation plateau. These are single-seed results with validation-selected checkpoints and float32 re-evaluation; they are bounded-budget comparisons, not converged or matched-compute results.}
\label{tab:cx_dpot}
\begin{tabular}{@{}lllrrrr@{}}
\toprule Task & Core init. & Updates & Median & IQR & p95 & Epochs \\
\midrule
LDC & Pretrained & All & 0.1548 & 0.101--0.421 & 1.839 & 85/85 \\
LDC & Random & All & 0.1991 & 0.134--0.375 & 0.833 & 84/84 \\
CosmicDose & Pretrained & All & 0.0467 & 0.025--0.069 & 0.100 & 60/40 \\
CosmicDose & Random & All & 0.0521 & 0.032--0.074 & 0.103 & 45/25 \\
PlasticDeform & Pretrained & All & 0.8485 & 0.548--3.801 & 50.87 & 25/25 \\
PlasticDeform & Random & All & 0.8354 & 0.618--6.948 & 34.61 & 25/20 \\
\bottomrule
\end{tabular}
\end{table}

\subsection{Reproducibility and resource accounting}
\label{sec:cx_methods_repro}

The ordinary operator arms use PyTorch and NumPy training seed 0, Adam with default betas $(0.9,0.999)$, epsilon $10^{-8}$ and zero weight decay. Cosine annealing has $T_{\max}=100$ and zero minimum learning rate. GNOT/Transolver evaluation batches contain one sample; the other ordinary arms use eight. The common selection criterion is mean normalized validation MSE. Validation is also checked at final or budget termination. Joint arms pad outputs to three channels but compute loss only on each task's valid channels. Shared DeepONet has three 256-unit GELU hidden layers in both branch and trunk, followed by linear outputs and 128 basis functions per channel.

MORPH and DPOT use random-core seed 1701, adapter initialization seed 1903 and training seed 0. Identical adapters are used in each pretrained/random pair. Adapter and trainable-core Adam learning rates are $10^{-3}$ and $10^{-4}$, respectively, with zero weight decay, cosine scheduling and norm-1 gradient clipping. MORPH training/evaluation batches are 32/8; DPOT uses 8/4. All these archived trainers disable cuDNN autotuning. Universal bitwise determinism is not claimed. Complete checkpoint identities, source hashes, stopping records and measured allocations accompany the machine-readable evidence; a source default is not substituted for an executed run's arguments.

\begin{table}[htbp]
\centering\scriptsize
\caption{Measured allocations for the 28 additional baseline arms. Sizes count complex weights as two real scalars and include adapters. Epochs are completed/selected; a wall-limited final epoch can be partial. GPU-hours are one-GPU Slurm wall time including startup, validation and final evaluation, excluding CPU gates/replay. They are not pure optimizer time or equal-compute budgets. L/C/P denote LDC/CosmicDose/PlasticDeform; pre/rnd denote pretrained/random cores and frz/full denote frozen/full core adaptation.}
\label{tab:cx_allocations}
\begin{tabular}{@{}lrrrr@{}}
\toprule Arm & Total (M) & Trainable (M) & Epochs & GPU-h\\
\midrule
Shared DeepONet joint & 0.422403 & 0.422403 & 100/100 & 0.072222 \\
fno L & 2.386995 & 2.386995 & 100/95 & 0.058333 \\
fno C & 2.288561 & 2.288561 & 30/5 & 0.024444 \\
fno P & 2.567089 & 2.567089 & 100/95 & 0.182500 \\
wno L & 5.050179 & 5.050179 & 85/65 & 0.119444 \\
wno C & 4.951617 & 4.951617 & 40/20 & 0.074444 \\
wno P & 5.230145 & 5.230145 & 40/35 & 0.186111 \\
gnot L & 0.749449 & 0.749449 & 44/40 & 0.338056 \\
gnot C & 0.748551 & 0.748551 & 10/10 & 0.365000 \\
gnot P & 0.750727 & 0.750727 & 17/17 & 0.338333 \\
transolver L & 0.112431 & 0.112431 & 100/100 & 0.270278 \\
transolver C & 0.111533 & 0.111533 & 28/5 & 0.380000 \\
transolver P & 0.113709 & 0.113709 & 47/40 & 0.337500 \\
NCWNO joint & 12.439133 & 12.439133 & 4/4 & 0.386389 \\
morph pre frz L & 10.143972 & 0.279564 & 95/95 & 0.057500 \\
morph rnd frz L & 10.143972 & 0.279564 & 85/85 & 0.051111 \\
morph pre full L & 10.143972 & 10.143972 & 60/40 & 0.050556 \\
morph rnd full L & 10.143972 & 10.143972 & 95/75 & 0.077500 \\
morph pre full C & 10.125532 & 10.125532 & 55/35 & 0.058889 \\
morph rnd full C & 10.125532 & 10.125532 & 30/10 & 0.034444 \\
morph pre full P & 10.177756 & 10.177756 & 100/100 & 0.263056 \\
morph rnd full P & 10.177756 & 10.177756 & 100/100 & 0.263611 \\
dpot pre full L & 8.466498 & 8.466498 & 85/85 & 0.336667 \\
dpot rnd full L & 8.466498 & 8.466498 & 84/84 & 0.336944 \\
dpot pre full C & 8.405048 & 8.405048 & 60/40 & 0.273889 \\
dpot rnd full C & 8.405048 & 8.405048 & 45/25 & 0.207222 \\
dpot pre full P & 8.579128 & 8.579128 & 25/25 & 0.336944 \\
dpot rnd full P & 8.579128 & 8.579128 & 25/20 & 0.336667 \\
\bottomrule
\end{tabular}
\end{table}

The 14 ordinary arms consumed 3.133056 allocated GPU-hours, eight MORPH arms 0.856667 and two DPOT training jobs 0.673611, totaling 4.663333. The four DPOT CosmicDose/PlasticDeform extensions add 4,157 GPU-seconds (1.154722 GPU-hours), giving 5.818056 GPU-hours for all 28 training arms. Their corrected final inference adds another 13 GPU-seconds. DPOT precision diagnosis and corrected final inference consumed another 23 GPU-seconds; these are not training reruns. This measured accounting is distinct from the reconstructed historical GEODE system-cost estimates.

\subsection{POD-plus-ridge control and acquisition-protocol provenance}
\label{sec:cx_clean_pod}

The clean POD basis is fitted to mean-centered training outputs with randomized singular value decomposition (SVD), seed 0, 32 oversampling vectors and seven power iterations, retaining up to 128 modes. Input means and standard deviations and the output mean use training rows only. We sweep ranks $\{8,16,32,64,128\}$ and ridge penalties $\{10^{-3},10^{-1},10^{1},10^{3}\}$ and select the minimum median physical validation relative $L^2$ error. The supplied LDC/CosmicDose validation partitions and the final 1,490 PlasticDeform training rows are excluded from all fitting. The selected model is saved before test arrays are opened, and is not refitted on validation. Test errors are evaluated on every native node. Direct validation reconstruction agrees with the orthogonal-projection error calculation, and all test error vectors replay from the serialized model using a separate contraction and batch size. The three tasks plus metadata audit completed in 39 seconds on 16 CPU cores, with no GPU allocation.

\begin{table}[htbp]
\centering\small
\caption{Clean POD-plus-ridge results. Errors are physical relative $L^2$ percentages; quantiles use linear interpolation. This is a linear control, using the same clean train/validation/test partitions as the external neural baselines.}
\label{tab:cx_clean_pod}
\begin{tabular}{@{}lrrrrr@{}}
\toprule Task & Rank & Ridge & Median & Mean & P95\\
\midrule
LDC & 64 & 0.001 & 10.9836 & 11.0427 & 21.5412 \\
CosmicDose & 8 & 1000 & 0.3378 & 0.3492 & 0.7192 \\
PlasticDeform & 8 & 0.1 & 20.1056 & 97.2459 & 590.4831 \\
\bottomrule
\end{tabular}
\end{table}

Saved historical checkpoint arguments establish multiple acquisition recipes. The rebuilt chain uses Adam at $10^{-3}$, batch 16 and StepLR with factor 0.6 every 60 epochs: HeatExchanger has a 100-epoch budget (selected zero-based epoch 96), followed by Subchannel with a 200-epoch budget (selected epoch 191) and unit profile-loss weight. The earlier Subchannel chain and its profile variant start from the earlier HeatExchanger parent, use learning rate $3\times10^{-4}$, batch 16, the same step schedule and 200 epochs; their selected epochs are 127 and 198, respectively. The recorded profile variant has profile coefficient 1. These protocols have different parents and must not be treated as one experiment.

The saved full-data trained/random library audit pairs both use seed 0, batch 16, learning rate $10^{-3}$ and StepLR factor 0.6 every 60 epochs. HeatExchanger uses 100 epochs; Subchannel uses 200 epochs and profile coefficient 1. The random control flag is recorded in the saved arguments, and the executed randomization rule is described in Methods. The historical checkpoint identifiers and arguments are archived with the evidence. This audit establishes those saved recipes but does not complete the linkage from every published fraction curve or historical sequential-table cell to its generating checkpoint. The clean chain on the paper checkpoint was rerun in both declared objective variants (\Cref{tab:supp_profile_loss}); historical chain values are not used in the comparison tables.

%% file: sections/supp_extra.tex
\subsection{Implementation details}
\label{app:details}

We use the \texttt{pytorch-wavelets} library for the 2D discrete wavelet transform. Each expert's wavelet convolution decomposes the 2D latent field into approximate ($\mathbb{A}$) and detail ($\mathbb{D} = \mathbb{H}, \mathbb{V}, \mathbb{D}$) coefficients at one decomposition level, and the spectral-domain kernels are then applied in the Fourier domain of these coefficients, combining wavelet-domain localization with spectral-domain efficiency. At the lowest decomposition level, eight separate sets of spectral weights are maintained for the approximate, horizontal-detail, vertical-detail and diagonal-detail subbands (two weights per subband for the positive and negative frequency halves).

The geometry-adaptive decoder uses the \texttt{radius} function from \texttt{torch-cluster} (or equivalently Open3D's \texttt{FixedRadiusSearch}) to build the neighbor graph between latent source points and physical query points. This graph is constructed once per task geometry and cached, since the output mesh is fixed for each task, and the aggregation uses \texttt{segment\_csr} from \texttt{torch-scatter} for efficient batched reduction. We use the Mish activation~\citep{misra2019mish}, defined as $\text{Mish}(x) = x \cdot \tanh(\text{softplus}(x))$, in the EWIBs and the gating MLPs, as it provides smoother gradients than ReLU while avoiding the vanishing-gradient issues of tanh; the decoder's kernel MLP uses GELU. The historical model was trained on a single NVIDIA GH200 GPU at NCSA; the clean rerun uses one NVIDIA A40 with gradient checkpointing.

\subsection{Routing-objective ablation}
\label{app:ablation}

This ablation was run on the historical checkpoint rather than the clean paper checkpoint; it is retained as a mechanism diagnostic and is not combined with the current headline errors. The distinct per-task routing reported in \SecRef{sec:routing}{the routing analysis (Results)} is produced by the task-specialization objective on the gates. \Cref{tab:ablation} isolates its components. With only a task-conditioned bias on the gate logits, the three tasks still share a single first-layer expert. Adding the Jensen--Shannon separation term is what moves the tasks onto distinct first-layer experts, but on its own it costs accuracy on PlasticDeform ($0.37\%$ to $0.45\%$) while leaving LDC and CosmicDose within noise. Adding the entropy-sharpening term retains the distinct routing and recovers that loss, and it is the variant with the lowest error on LDC ($1.09\%$ against $2.24\%$ and $2.13\%$) and PlasticDeform ($0.35\%$). The effect of the gating objective on accuracy is therefore task-dependent rather than negligible: separation alone is not free, and the full objective is what makes distinct routing and accuracy compatible.

\begin{table}[htbp]
  \centering
  \caption{Historical-checkpoint effect of the gating objective on first-layer routing and test accuracy (denormalized median relative $L^2$, \%). ``Distinct'' indicates the three tasks select three different first-layer experts.}
  \label{tab:ablation}
  \small
  \setlength{\tabcolsep}{5pt}
  \renewcommand{\arraystretch}{1.15}
  \begin{tabular}{@{}lcccc@{}}
    \toprule
    \textbf{Gating objective} & \textbf{First-layer routing} & \textbf{LDC} & \textbf{CosmicDose} & \textbf{PlasticDeform} \\
    \midrule
    Task-bias only              & shared      & 2.24 & 0.080 & 0.37 \\
    $+$ JS divergence           & distinct    & 2.13 & 0.086 & 0.45 \\
    $+$ entropy sharpening (GEODE) & distinct & 1.09 & 0.080 & 0.35 \\
    \bottomrule
  \end{tabular}
\end{table}

GEODE concentrates this specialization pressure on the deeper layers~2--3. Even so, the deeper layers do not develop distinct per-task routing and instead converge to a common choice, consistent with the shared-decoder interpretation in \SecRef{sec:routing}{the routing analysis (Results)}: they are driven toward a common latent representation, so routing specialization is naturally confined to the first layer, while the added pressure still benefits overall optimization.

\subsection{Predicted fields for the sequentially acquired tasks}
\label{app:cl_fields}

\begingroup\revon
\Cref{fig:cl_fields} shows one field per task. Both tasks predict three, and the remaining
fields are given here for completeness. The model, the test case and the rendering are identical
to the main-text figure; only the channel differs.
\endgroup

\begin{figure}[htbp]
  \revon
  \centering
  \includegraphics[width=0.92\textwidth]{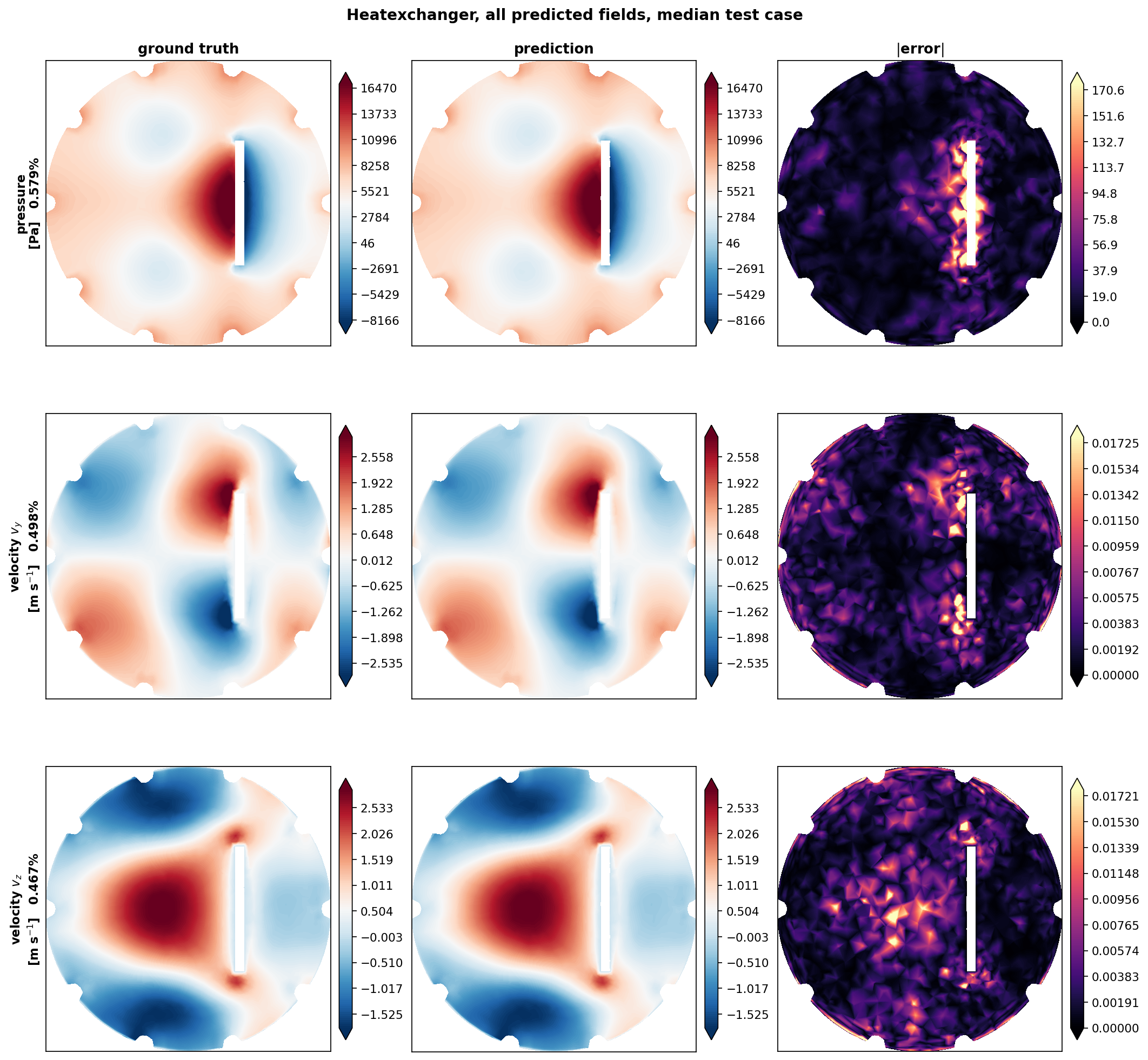}
  \caption{Heat exchanger: pressure and both in-plane velocity components, median test case,
  on the solver's own CFD mesh. The error in every channel concentrates along the internal plate,
  where the field changes most sharply over the shortest distance.}
  \label{fig:cl_fields_hx}
\end{figure}

\begin{figure}[htbp]
  \revon
  \centering
  \includegraphics[width=0.92\textwidth]{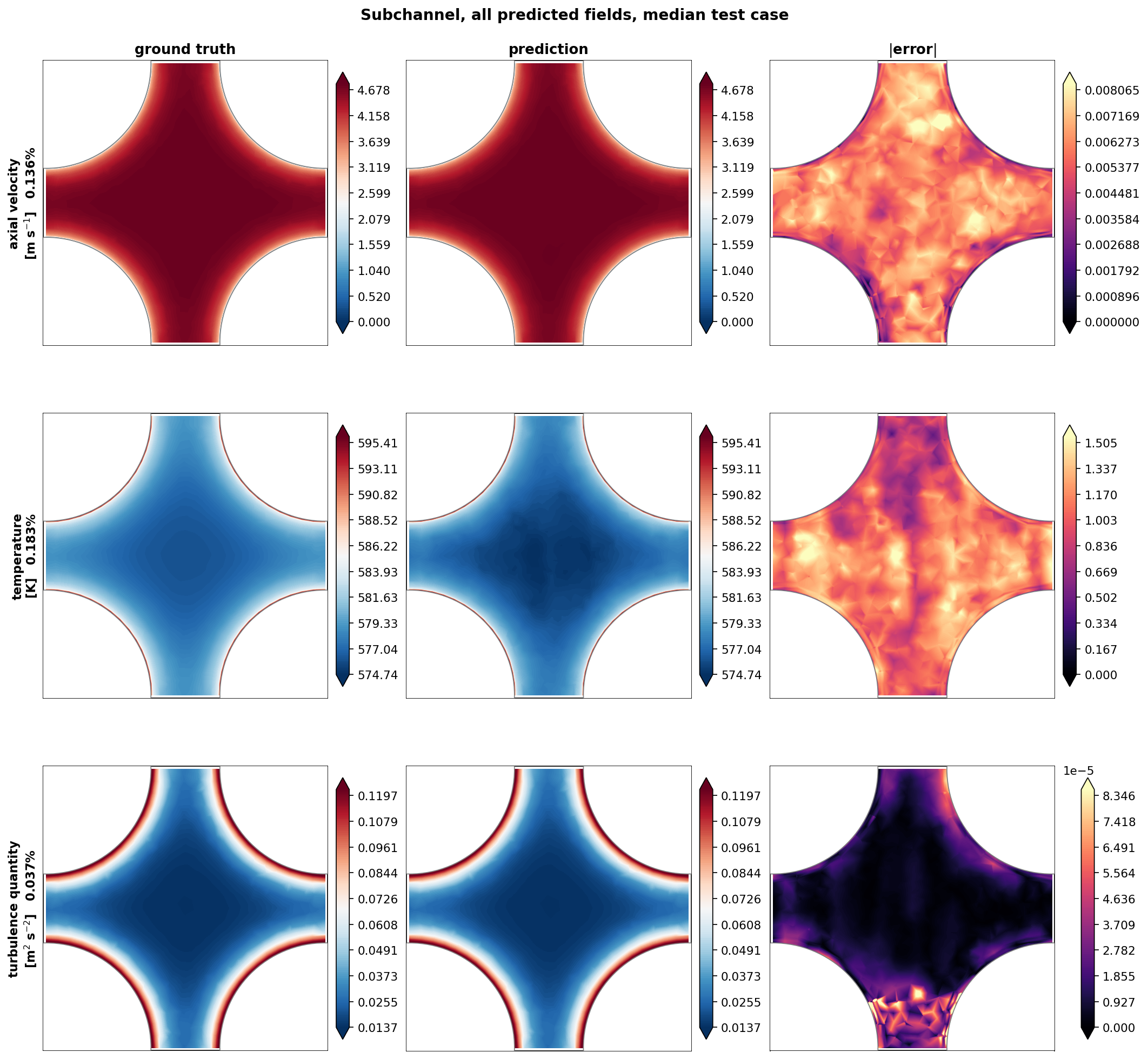}
  \caption{Subchannel: axial velocity, coolant temperature and the turbulence quantity, median
  test case. The temperature panel illustrates the difficulty described in \SecRef{sec:transfer}{the reuse analysis (Results)}:
  against a $574$~K mean, the $21$~K of wall-to-core structure is barely visible on a shared color
  scale, which is why the relative $L^2$ of that channel is a poor guide to whether its shape has
  been learned. The turbulence quantity peaks against the rod walls and that peak is recovered.}
  \label{fig:cl_fields_sub}
\end{figure}

\subsection{Gate distributions across trained models}
\label{app:gates}

\begingroup\revon
\Cref{fig:routing} shows the gates of the three jointly pretrained tasks.
\SecRef{sec:routing}{The routing analysis (Results)} draws on every model trained in this study, and the remaining gate
distributions are collected here so that its claims can be checked rather than taken on trust from the
summary in \Cref{tab:routing}. Presentation follows the main-text figure: log color scale, one
row per expert layer, dominant expert boxed. The number printed at the right of each row is the
effective expert count $\exp(H)$, which is the quantity the analysis argues about and which cannot
be read reliably off a heat map.
\endgroup

\begin{figure}[htbp]
  \revon
  \centering
  \includegraphics[width=0.78\textwidth]{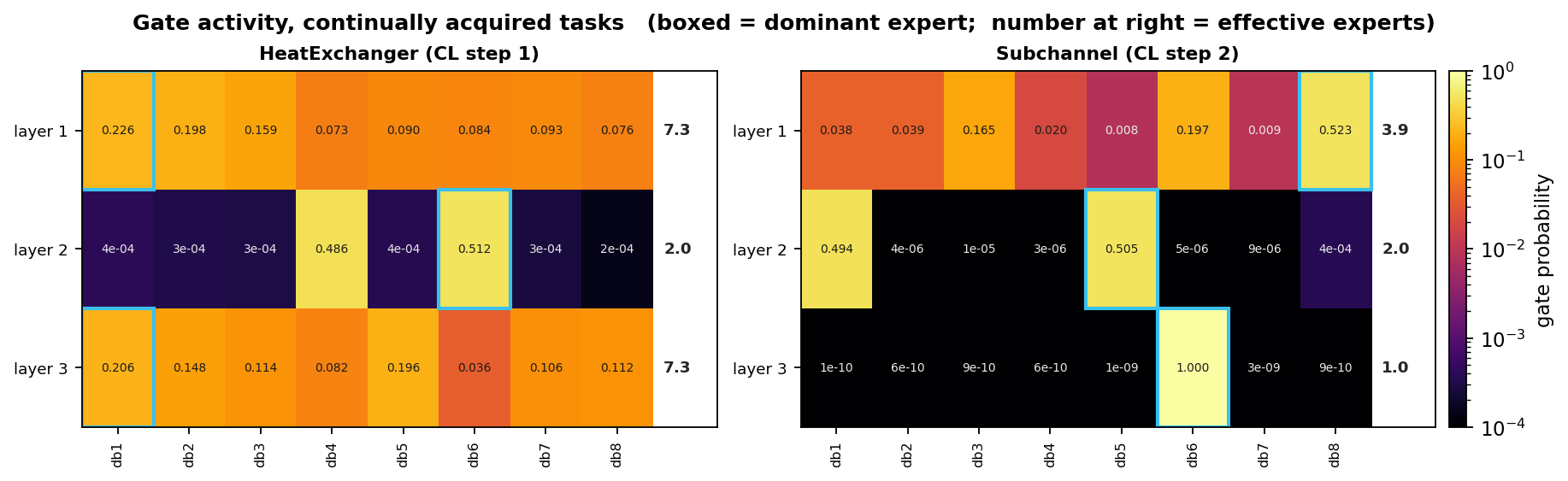}
  \caption{Gates of the two continually acquired tasks, read from the model after both adaptation
  steps. Both mix far more experts than any jointly pretrained task, the heat exchanger reaching
  $7.3$ effective experts in the first and third layers, against $1.1$ to $2.3$ (first layer) and
  $1.0$ (third) for the pretrained tasks.
  \SecRef{sec:routing}{The routing analysis (Results)} attributes this to the frozen library breaking the expert-gate co-adaptation
  that otherwise drives routing to collapse. Two cautions in reading these panels. Where a row is
  close to flat, the boxed maximum is only weakly dominant and should not be read as a selection:
  the heat exchanger's first layer runs from $0.073$ to $0.226$. And the second layer sits at
  exactly $2.0$ effective experts in every continually acquired arm, including the randomized
  control, which indicates that this split is imposed by the deep-layer term in the gating
  objective rather than learned from the data.}
  \label{fig:gates_cl}
\end{figure}

\begin{figure}[htbp]
  \revon
  \centering
  \includegraphics[width=\textwidth]{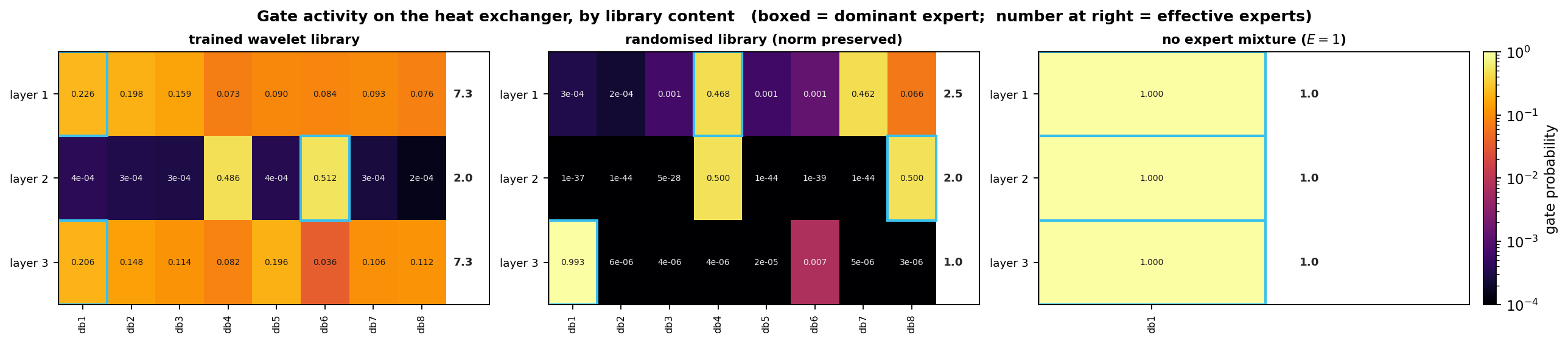}
  \caption{Gates on the same held-out task under three different libraries. Over the trained
  wavelet library the first and third layers are close to uniform; over a library randomized at
  matched Frobenius norm the gate concentrates to $2.5$ effective experts; over a backbone with no
  expert mixture the single available expert carries all the mass, which is a trivial rather than a
  learned collapse and is shown for completeness. Dense mixing therefore appears only when the
  library holds distinct learned functions, so the gate is sensitive to library content and not
  only to the task it serves.}
  \label{fig:gates_controls}
\end{figure}

\begin{figure}[htbp]
  \revon
  \centering
  \includegraphics[width=\textwidth]{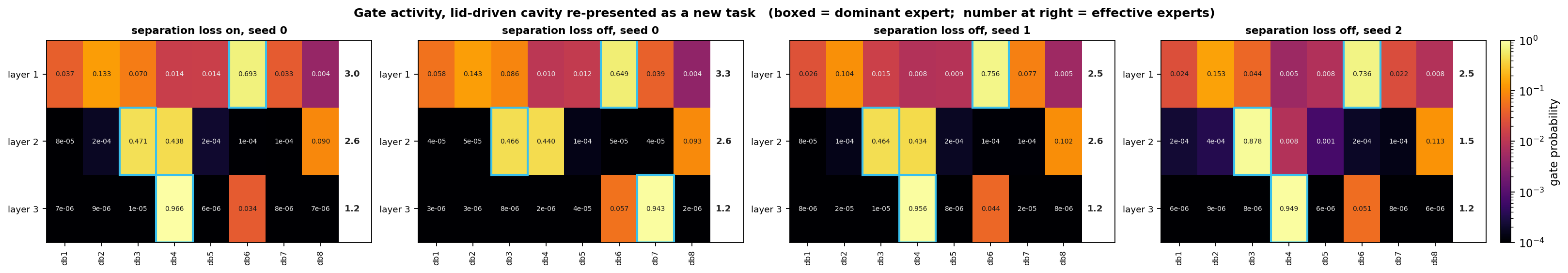}
  \caption{Gates for lid-driven cavity flow re-presented to the trained model as a new task on a
  reshuffled split, across two random seeds and with the routing-separation loss enabled and
  disabled. All four runs converge on db6 followed by db3 in the first two layers, and none
  recovers the assignment db2 followed by db8 that joint training gave this task. The routing is
  therefore reproducible without being unique, and db6 is the expert that joint training assigned to the
  elastoplastic problem.}
  \label{fig:gates_ldcrep}
\end{figure}

\subsection{Seed replicates of the joint model}
The joint three-system model was retrained twice from different random seeds with the recipe, schedule and 120-epoch budget of the reported checkpoint (whose selected epoch is 116), and evaluated with the same pipeline on the same held-out cases. The heat-exchanger interface and the subchannel interface, over the trained library and over the norm-matched random library, were likewise rerun from seeds 1 and 2 and evaluated with the seed-0 query subset. Table~\ref{tab:seeds} lists the medians, with interquartile ranges for the joint model. The specimen error is reproduced within 0.02 percentage points; the cavity and dose errors are roughly three- to fourfold larger for both replicates, so the reported checkpoint is the best of the three on those two systems. All routing, sequential-acquisition and transfer experiments in the main text start from the reported checkpoint. The heat-exchanger interface is seed-stable. For the subchannel, the interface over the trained library learns the field level in all three seeds, whereas over the random library two seeds remain at the constant-level solution and one escapes it, so the trained-versus-random contrast is an escape rate rather than a fixed ratio. The baselines are single-seed.

\begin{table}[htbp]
  \centering
  \caption{\textbf{Training seeds.} Median relative $L^2$ error in percent over the held-out cases (495 cavity, 359 dose, 100 specimen; 100 heat-exchanger and 100 subchannel cases for the interfaces), with interquartile ranges in parentheses for the joint model. Seed 0 is the run reported throughout.}
  \label{tab:seeds}
  \small
  \begin{tabular}{@{}lccc@{}}
    \toprule
    Seed & Cavity & Dose & Specimen \\
    \midrule
    0 (reported) & 0.755 (0.54--1.34) & 0.031 (0.019--0.053) & 0.399 (0.25--2.52) \\
    1 & 2.87 (2.13--3.83) & 0.100 (0.058--0.136) & 0.42 (0.30--1.79) \\
    2 & 1.97 (1.34--3.10) & 0.134 (0.084--0.179) & 0.39 (0.27--2.32) \\
    \midrule
    Interface & Heat exchanger & Subchannel, trained library & Subchannel, random library \\
    \midrule
    0 (reported) & 0.719 & 0.121 & 4.524 \\
    1 & 0.657 & 0.068 & 4.653 \\
    2 & 0.669 & 0.073 & 0.160 \\
    \bottomrule
  \end{tabular}
\end{table}

\subsection{Error distributions}
\label{app:error_dist}
\begin{figure}[htbp]
  \centering
  \includegraphics[width=\textwidth]{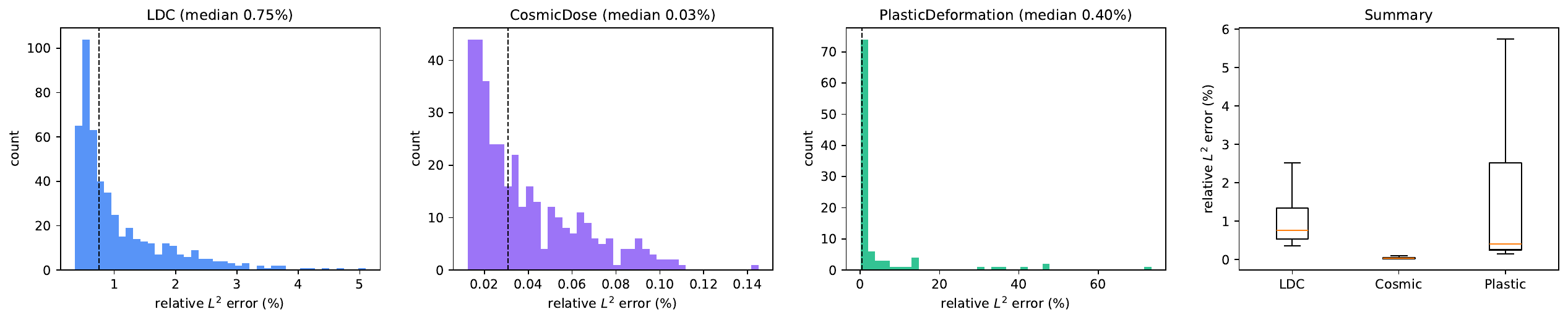}
  \caption{Per-sample relative $L^2$ error distributions across the full test set for each task (LDC $n{=}495$, CosmicDose $n{=}359$, PlasticDeform $n{=}100$). CosmicDose and LDC are tightly concentrated, whereas PlasticDeform is heavy-tailed (median $0.399\%$, mean $4.80\%$, 95th percentile $34.7\%$ using linear interpolation), motivating the use of the median alongside the mean in \Cref{tab:results}.}
  \label{fig:error_dist}
\end{figure}

\subsection{Per-channel predictions for the lid-driven cavity}
\label{app:ldc_channels}
\begin{figure}[!ht]
  \centering
  \includegraphics[width=0.7\textwidth]{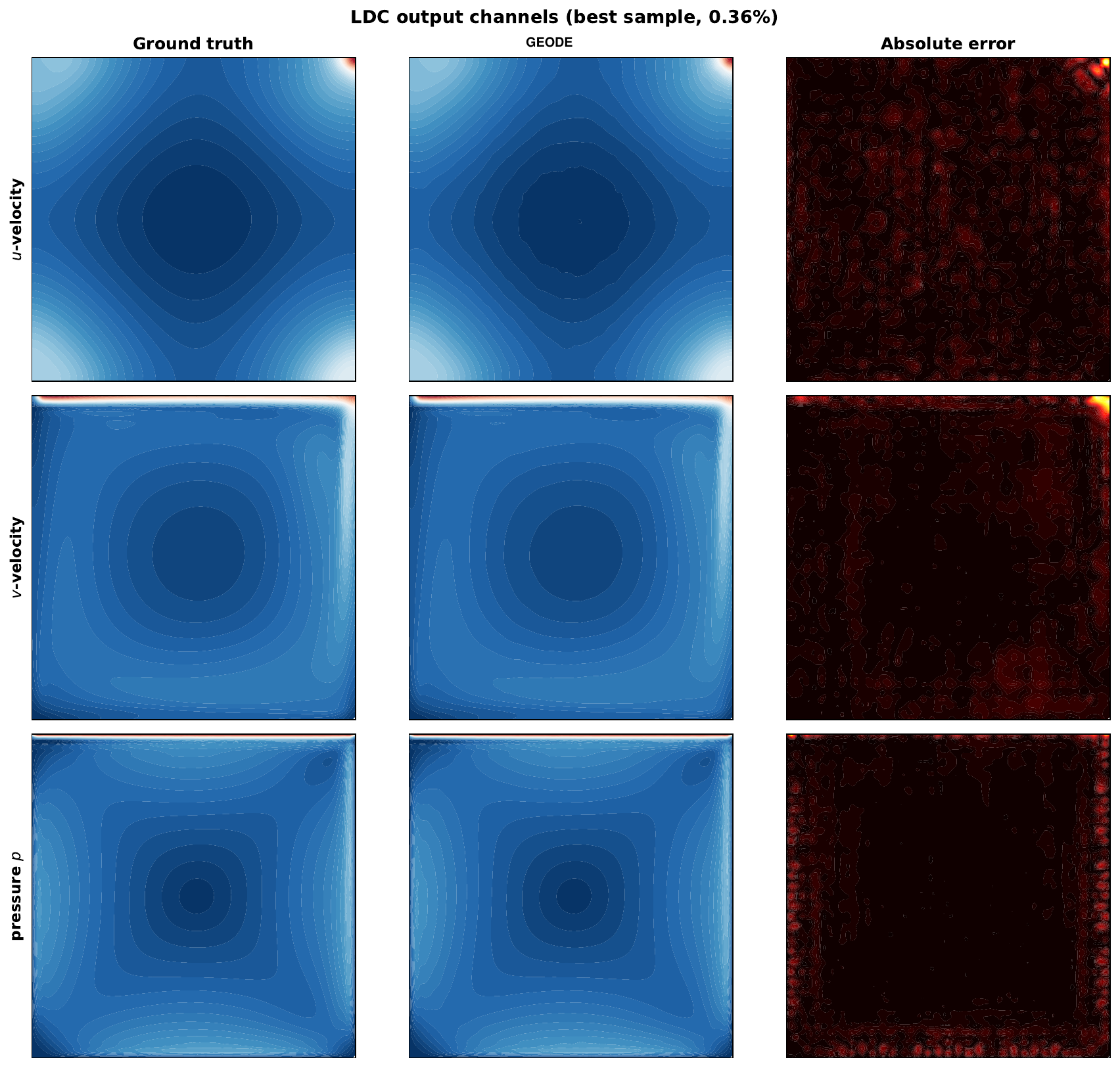}
  \caption{LDC predictions for all three output channels ($u$, $v$, $p$) on a representative test sample: ground truth (\emph{left}), GEODE prediction (\emph{center}) and absolute error (\emph{right}). The model resolves the velocity components and the pressure field jointly within a single set of weights.}
  \label{fig:ldc_channels}
\end{figure}

\Cref{fig:ldc_channels} complements the $v$-velocity comparison of \Cref{fig:ldc} by showing all three LDC output channels for a representative test sample.

%% file: main.bbl
\begin{thebibliography}{29}
\providecommand{\natexlab}[1]{#1}
\providecommand{\url}[1]{\texttt{#1}}
\expandafter\ifx\csname urlstyle\endcsname\relax
  \providecommand{\doi}[1]{doi: #1}\else
  \providecommand{\doi}{doi: \begingroup \urlstyle{rm}\Url}\fi

\bibitem[Lu et~al.(2021)Lu, Jin, Pang, Zhang, and Karniadakis]{lu2021learning}
Lu~Lu, Pengzhan Jin, Guofei Pang, Zhongqiang Zhang, and George~Em Karniadakis.
\newblock Learning nonlinear operators via {DeepONet} based on the universal
  approximation theorem of operators.
\newblock \emph{Nature Machine Intelligence}, 3\penalty0 (3):\penalty0
  218--229, 2021.

\bibitem[Li et~al.(2020)Li, Kovachki, Azizzadenesheli, Liu, Bhattacharya,
  Stuart, and Anandkumar]{li2020fourier}
Zongyi Li, Nikola Kovachki, Kamyar Azizzadenesheli, Burigede Liu, Kaushik
  Bhattacharya, Andrew Stuart, and Anima Anandkumar.
\newblock Fourier neural operator for parametric partial differential
  equations, 2020.

\bibitem[Li et~al.(2021)Li, Kovachki, Azizzadenesheli, Liu, Bhattacharya,
  Stuart, and Anandkumar]{li2021fourier}
Zongyi Li, Nikola Kovachki, Kamyar Azizzadenesheli, Burigede Liu, Kaushik
  Bhattacharya, Andrew Stuart, and Anima Anandkumar.
\newblock Fourier neural operator for parametric partial differential
  equations.
\newblock In \emph{International Conference on Learning Representations}, 2021.

\bibitem[Kovachki et~al.(2023)Kovachki, Li, Liu, Azizzadenesheli, Bhattacharya,
  Stuart, and Anandkumar]{kovachki2023neural}
Nikola Kovachki, Zongyi Li, Burigede Liu, Kamyar Azizzadenesheli, Kaushik
  Bhattacharya, Andrew Stuart, and Anima Anandkumar.
\newblock Neural operator: Learning maps between function spaces with
  applications to {PDE}s.
\newblock \emph{Journal of Machine Learning Research}, 24\penalty0
  (89):\penalty0 1--97, 2023.

\bibitem[Tripura and Chakraborty(2023)]{tripura2023wavelet}
Tapas Tripura and Souvik Chakraborty.
\newblock Wavelet neural operator for solving parametric partial differential
  equations in computational mechanics problems.
\newblock \emph{Computer Methods in Applied Mechanics and Engineering},
  404:\penalty0 115783, 2023.

\bibitem[Li et~al.(2023{\natexlab{a}})Li, Huang, Liu, and
  Anandkumar]{li2022fourier_geometry}
Zongyi Li, Daniel~Zhengyu Huang, Burigede Liu, and Anima Anandkumar.
\newblock Fourier neural operator with learned deformations for {PDE}s on
  general geometries.
\newblock \emph{Journal of Machine Learning Research}, 24\penalty0
  (388):\penalty0 1--26, 2023{\natexlab{a}}.

\bibitem[Navaneeth and Chakraborty(2025)]{navaneeth2025geometry_waveformer}
N.~Navaneeth and Souvik Chakraborty.
\newblock Geometry adaptive waveformer for cardio-vascular modeling.
\newblock \emph{Computers in Biology and Medicine}, 190:\penalty0 110069, 2025.

\bibitem[Li et~al.(2023{\natexlab{b}})Li, Kovachki, Choy, Li, Kossaifi, Otta,
  Nabian, Stadler, Hundt, Azizzadenesheli, and Anandkumar]{li2023geometry}
Zongyi Li, Nikola Kovachki, Christopher Choy, Boyi Li, Jean Kossaifi,
  Shourya~Prakash Otta, Mohammad~Amin Nabian, Maximilian Stadler, Christian
  Hundt, Kamyar Azizzadenesheli, and Anima Anandkumar.
\newblock Geometry-informed neural operator for large-scale {3D} {PDE}s.
\newblock \emph{Advances in Neural Information Processing Systems}, 36,
  2023{\natexlab{b}}.

\bibitem[Sarkar and Chakraborty(2026)]{sarkar2026pigsp2gno}
Subhankar Sarkar and Souvik Chakraborty.
\newblock Physics- and geometry-aware spatio-spectral graph neural operator for
  time-independent and time-dependent {PDE}s.
\newblock \emph{Journal of Computational Physics}, 562:\penalty0 115029, 2026.
\newblock \doi{10.1016/j.jcp.2026.115029}.

\bibitem[Hao et~al.(2023)Hao, Wang, Su, Ying, Dong, Liu, Cheng, Song, and
  Zhu]{hao2023gnot}
Zhongkai Hao, Zhengyi Wang, Hang Su, Chengyang Ying, Yinpeng Dong, Songming
  Liu, Ze~Cheng, Jian Song, and Jun Zhu.
\newblock {GNOT}: A general neural operator transformer for operator learning.
\newblock In \emph{Proceedings of the 40th International Conference on Machine
  Learning}, 2023.

\bibitem[Wu et~al.(2024)Wu, Luo, Wang, Wang, and Long]{wu2024transolver}
Haixu Wu, Huakun Luo, Haowen Wang, Jianmin Wang, and Mingsheng Long.
\newblock Transolver: A fast transformer solver for {PDE}s on general
  geometries.
\newblock In \emph{Proceedings of the 41st International Conference on Machine
  Learning}, 2024.

\bibitem[Yang et~al.(2023)Yang, Liu, Meng, and Osher]{yang2023context}
Liu Yang, Siting Liu, Tingwei Meng, and Stanley~J. Osher.
\newblock In-context operator learning with data prompts for differential
  equation problems.
\newblock \emph{Proceedings of the National Academy of Sciences}, 120\penalty0
  (39):\penalty0 e2310142120, 2023.

\bibitem[Subramanian et~al.(2023)Subramanian, Harrington, Keutzer, Bhimji,
  Morozov, Mahoney, and Gholami]{subramanian2024towards}
Shashank Subramanian, Peter Harrington, Kurt Keutzer, Wahid Bhimji, Dmitriy
  Morozov, Michael~W. Mahoney, and Amir Gholami.
\newblock Towards foundation models for scientific machine learning:
  Characterizing scaling and transfer behavior.
\newblock \emph{Advances in Neural Information Processing Systems}, 36, 2023.

\bibitem[McCabe et~al.(2023)McCabe, R\'{e}galdo-Saint~Blancard, Parker, Ohana,
  Cranmer, Bietti, Eickenberg, Golkar, Krawezik, Lanusse, Pettee, Tesileanu,
  Cho, and Ho]{mccabe2023multiple}
Michael McCabe, Bruno R\'{e}galdo-Saint~Blancard, Liam~Holden Parker, Ruben
  Ohana, Miles Cranmer, Alberto Bietti, Michael Eickenberg, Siavash Golkar,
  Geraud Krawezik, Fran\c{c}ois Lanusse, Mariel Pettee, Tiberiu Tesileanu,
  Kyunghyun Cho, and Shirley Ho.
\newblock Multiple physics pretraining for physical surrogate models, 2023.

\bibitem[Hao et~al.(2024)Hao, Su, Liu, Berner, Ying, Su, Anandkumar, Song, and
  Zhu]{hao2024dpot}
Zhongkai Hao, Chang Su, Songming Liu, Julius Berner, Chengyang Ying, Hang Su,
  Anima Anandkumar, Jian Song, and Jun Zhu.
\newblock {DPOT}: Auto-regressive denoising operator transformer for
  large-scale {PDE} pre-training, 2024.

\bibitem[Herde et~al.(2024)Herde, Raoni\'{c}, Rohner, K\"{a}ppeli, Molinaro,
  de~B\'{e}zenac, and Mishra]{herde2024poseidon}
Maximilian Herde, Bogdan Raoni\'{c}, Tobias Rohner, Roger K\"{a}ppeli, Roberto
  Molinaro, Emmanuel de~B\'{e}zenac, and Siddhartha Mishra.
\newblock Poseidon: Efficient foundation models for {PDE}s, 2024.

\bibitem[Tripura and Chakraborty(2026)]{chakraborty2024ncwno}
Tapas Tripura and Souvik Chakraborty.
\newblock Neural combinatorial wavelet neural operator for catastrophic
  forgetting free in-context operator learning of multiple partial differential
  equations.
\newblock \emph{Computer Physics Communications}, 318:\penalty0 109882, 2026.
\newblock \doi{10.1016/j.cpc.2025.109882}.

\bibitem[McCabe et~al.(2025)McCabe, Mukhopadhyay, Marwah,
  R\'{e}galdo-Saint~Blancard, Rozet, Diaconu, Meyer, Wong, Sotoudeh, Bietti,
  Espejo, Fear, Golkar, Hehir, Hirashima, Krawezik, Lanusse, Morel, Ohana,
  Parker, Pettee, Shen, Cho, Cranmer, and Ho]{mccabe2025walrus}
Michael McCabe, Payel Mukhopadhyay, Tanya Marwah, Bruno
  R\'{e}galdo-Saint~Blancard, Fran\c{c}ois Rozet, Cristiana Diaconu, Lucas
  Meyer, Kaze W.~K. Wong, Hadi Sotoudeh, Alberto Bietti, Irina Espejo, Rio
  Fear, Siavash Golkar, Tom Hehir, Keiya Hirashima, Geraud Krawezik,
  Fran\c{c}ois Lanusse, Rudy Morel, Ruben Ohana, Liam Parker, Mariel Pettee,
  Jeff Shen, Kyunghyun Cho, Miles Cranmer, and Shirley Ho.
\newblock Walrus: A cross-domain foundation model for continuum dynamics, 2025.

\bibitem[Ye et~al.(2025)Ye, Liu, Wu, Jiang, Chen, Zhang, Huang, Meng, Zou, Liu,
  and Dong]{ye2025pdeformer2}
Zhanhong Ye, Zining Liu, Bingyang Wu, Hongjie Jiang, Leheng Chen, Minyan Zhang,
  Xiang Huang, Qinghe Meng, Jingyuan Zou, Hongsheng Liu, and Bin Dong.
\newblock {PDEformer-2}: A versatile foundation model for two-dimensional
  partial differential equations, 2025.

\bibitem[Rautela et~al.(2025)Rautela, Most, Mansingh, Love, Scheinker, Oyen,
  Debardeleben, Lawrence, and Biswas]{rautela2025morph}
Mahindra Rautela, Alexander Most, Xhulja Mansingh, Robert Love, Alexander
  Scheinker, Diane Oyen, Nathan Debardeleben, Earl Lawrence, and Ayan Biswas.
\newblock Morph: Pde foundation models with arbitrary data modality, 2025.

\bibitem[Soares et~al.(2025)Soares, Brazil, Shirasuna, de~Carvalho, and
  Malossi]{soares2025pdefm}
Eduardo Soares, Emilio~Vital Brazil, Victor Shirasuna, Breno W. S.~R.
  de~Carvalho, and Cristiano Malossi.
\newblock Towards a foundation model for partial differential equations across
  physics domains, 2025.

\bibitem[Kobayashi et~al.(2024)Kobayashi, Ahmed, Park, Sarkar, Koric,
  Chakraborty, and Alam]{kobayashi2024mimonet}
Kazuma Kobayashi, Farid Ahmed, Jaewan Park, Subhankar Sarkar, Seid Koric,
  Souvik Chakraborty, and Syed~Bahauddin Alam.
\newblock Virtual sensing to enable real-time monitoring of inaccessible
  locations and unmeasurable parameters, 2024.

\bibitem[Bommasani et~al.(2021)]{bommasani2021opportunities}
Rishi Bommasani et~al.
\newblock On the opportunities and risks of foundation models, 2021.

\bibitem[Roy et~al.(2026)Roy, Kobayashi, Chakraborty, Rizwan-uddin, and
  Alam]{roy2026adversarial}
Samrendra Roy, Kazuma Kobayashi, Souvik Chakraborty, Rizwan-uddin, and
  Syed~Bahauddin Alam.
\newblock Adversarial vulnerabilities in neural operator digital twins:
  Gradient-free attacks on nuclear thermal-hydraulic surrogates, 2026.

\bibitem[Hossain et~al.(2025)Hossain, Ahmed, Kobayashi, Koric, Abueidda, and
  Alam]{hossain2025virtualsensing}
Raisa Hossain, Farid Ahmed, Kazuma Kobayashi, Seid Koric, Diab Abueidda, and
  Syed~Bahauddin Alam.
\newblock Virtual sensing-enabled digital twin framework for real-time
  monitoring of nuclear systems leveraging deep neural operators.
\newblock \emph{npj Materials Degradation}, 9\penalty0 (1):\penalty0 21, 2025.
\newblock \doi{10.1038/s41529-025-00557-y}.

\bibitem[Kobayashi et~al.(2025{\natexlab{a}})Kobayashi, Garg, Ahmed,
  Chakraborty, and Alam]{kobayashi2025conformalized}
Kazuma Kobayashi, Shailesh Garg, Farid Ahmed, Souvik Chakraborty, and
  Syed~Bahauddin Alam.
\newblock Distribution-free uncertainty-aware virtual sensing via conformalized
  neural operators, 2025{\natexlab{a}}.

\bibitem[Kobayashi et~al.(2025{\natexlab{b}})Kobayashi, Roy, Koric, Abueidda,
  and Alam]{kobayashi2025tron}
Kazuma Kobayashi, Samrendra Roy, Seid Koric, Diab Abueidda, and Syed~Bahauddin
  Alam.
\newblock From proxies to fields: Spatiotemporal reconstruction of global
  radiation from sparse sensor sequences, 2025{\natexlab{b}}.

\bibitem[He et~al.(2024)He, Kushwaha, Park, Koric, Abueidda, and
  Jasiuk]{he2024sdeeponet}
Junyan He, Shashank Kushwaha, Jaewan Park, Seid Koric, Diab Abueidda, and Iwona
  Jasiuk.
\newblock Predictions of transient vector solution fields with sequential deep
  operator network.
\newblock \emph{Acta Mechanica}, 235\penalty0 (8):\penalty0 5257--5272, 2024.

\bibitem[Misra(2019)]{misra2019mish}
Diganta Misra.
\newblock Mish: A self regularized non-monotonic activation function, 2019.

\end{thebibliography}
